\documentclass[sn-mathphys]{sn-jnl}

\jyear{2022}%

\theoremstyle{thmstyleone}%
\theoremstyle{thmstyletwo}%

\theoremstyle{thmstylethree}%

\usepackage{graphicx}
\usepackage{subcaption}
\usepackage{physics}
\begin{document}

\title[Encoding For Healthcare Data Democratisation]{Data Encoding For Healthcare Data Democratisation and Information Leakage Prevention}


\author*[1]{\fnm{Anshul} \sur{Thakur}}\email{anshul.thakur@eng.ox.ac.uk}
\author[1]{\fnm{Tingting} \sur{Zhu}}\email{tingting.zhu@eng.ox.ac.uk}
\equalcont{These authors contributed equally to this work.}
\author[2]{\fnm{Vinayak} \sur{Abrol}}\email{abrol@iiid.ac.in}
\equalcont{These authors contributed equally to this work.}

\author[1]{\fnm{Jacob} \sur{Armstrong}}\email{jacob.armstrong@eng.ox.ac.uk}
\author*[1]{\fnm{Yujiang} \sur{Wang}}\email{yujiang.wang@eng.ox.ac.uk}
\author[1]{\fnm{David A.} \sur{Clifton}}\email{david.clifton@eng.ox.ac.uk}

\affil*[1]{\orgdiv{Department of Engineering Science}, \orgname{University of Oxford}, \orgaddress{ \postcode{OX3 7DQ}, \state{Oxfordshire}, \country{United Kingdom}}}

\affil[2]{\orgdiv{Infosys Centre for AI}, \orgname{IIIT Delhi}, \orgaddress{\country{India}}}


\abstract{
The lack of data democratization and information leakage from trained models hinder the development and acceptance of robust deep learning-based healthcare solutions. This paper argues that irreversible data encoding can provide an effective solution to achieve data democratization without violating the privacy constraints imposed on healthcare data and clinical models. An ideal encoding framework transforms the data into a new space where it is imperceptible to a manual or computational inspection. However, encoded data should preserve the semantics of the original data such that deep learning models can be trained effectively. This paper hypothesizes the characteristics of the desired encoding framework and then exploits random projections and random quantum encoding to realize this framework for dense and longitudinal or time-series data. Experimental evaluation highlights that models trained on encoded time-series data effectively uphold the information bottleneck principle and hence, exhibit lesser information leakage from trained models.           
}

\maketitle

\section{Introduction}
\label{sec:sec1}
In recent years, deep learning has demonstrated remarkable success in a wide variety of fields \cite{Goodfellow}, and it is expected to have a significant impact on healthcare as well \cite{hinton}. Many attempts have been made to achieve this breakthrough in healthcare informatics, which often deals with noisy, heterogeneous, and non-standardized electronic health records (EHRs) \cite{7801947}. However, most clinical deep learning tools are either not robust enough or have not been tested in real-world scenarios \cite{challenges,jama_rev}. Deep learning solutions, approved by regulatory bodies, are less common in healthcare informatics, which shows that deep learning hasn't had the same level of success as in other fields such as speech and image processing \cite{aisu2022regulatory}. Along with well-known explainability challenges in deep learning models \cite{sapoval2022current}, the lack of \emph{data democratization} \cite{lewis2020data} and \emph{latent information leakage} (information leakage from trained models) \cite{liu2020privacy,mireshghallah2020privacy} can also be regarded as a major hindrance in the development and acceptance of robust clinical deep learning solutions. In the current context, data democratization and information leakage can be described as:    

\begin{itemize}
\setlength{\itemsep}{0.5\baselineskip}
    \item \textbf{\emph{Data democratization :}} It involves making digital healthcare data available to a wider cohort of the AI researchers. Achieving healthcare data democratization can result in global clinical models that are trained on data sampled from multiple geographical locations instead of being limited to a single site. These models are expected to be robust to population-specific distribution shifts and to exhibit better generalization. The wider access to healthcare data might also facilitate algorithmic contributions tailored for healthcare applications through a broader AI research base. However, healthcare data is ``sensitive'' and is rightly protected by data privacy laws making data democratization difficult \cite{vokinger_stekhoven_krauthammer_2020,thakur2021dynamic}. 
    
    \item \textbf{\emph{Latent Information Leakage:}} Deep learning models are known for their higher complexity and ability to learn the non-targeted latent information about the underlying population \cite{mireshghallah2020privacy}. This latent information often acts as an inductive bias to improve the predictive performance of the model. However, the latent information can be sensitive or help in inferring the information such as age, sex, and chronic or acute medical conditions of the patients. The revelation of this sensitive patient information can be considered a privacy violation.  
\end{itemize}
Hence, data democratization and prevention of latent information leakage are two of the important factors required to develop better clinical deep learning solutions that are secure and widely acceptable.

\vspace{0.1cm}
Data democratization can be equated with the irreversible de-identification of healthcare data so that no patient can be linked to an electronic health record (EHR). A ``truly'' de-identified dataset cannot be considered sensitive or private, so sharing it publicly would not result in a violation of any data privacy laws \cite{el2015anonymising}. However, researchers have not developed a truly irreversible de-identification mechanism, and there is always a risk of re-identification \cite{el2015anonymising,henriksen2016re,vokinger_stekhoven_krauthammer_2020}. It is a common practice to anonymize healthcare data, but the resulting data might not always be considered to be completely de-identified. In general, the notion of anonymity or de-identification is closely related to the amount of computational effort and time required to re-identify a patient from the data. An EHR can be considered non-anonymous (even after the anonymization process) if the efforts to re-identify the patient are considered reasonable. The ``reasonable efforts'' are subjective and should often change with advancements in technology \cite{vokinger_stekhoven_krauthammer_2020}. As a result, simple data anonymization is not enough to achieve ``true'' de-identification and data democratization. Hence, there is a requirement for information processing mechanisms that could mask private information while retaining the data semantics to enable data sharing or democratization.          

\vspace{0.1cm}
Aside from data democratization, trained clinical deep learning models also raise privacy concerns. These models have been shown to learn bio-markers of diabetic retinopathy, anemia, and chronic kidney disease from fundus images \cite{zhang2021deep}. Apart from that, deep learning models can also predict gender, sex, ethnicity, and smoking status from a fundus image \cite{poplin2018prediction}. Hence, it is quite possible that a model trained for predicting diabetic retinopathy from fundus images can learn a feature representation that may reveal non-targeted patient characteristics and sensitive information regarding the ailment of a patient suffering from chronic kidney disease and anemia. In the same way, a model trained for mortality prediction based on the first 48 hours of hospitalization in the intensive care unit (ICU) can provide information on the patient's acute as well as chronic conditions that may or may not be related to the current ICU stay or mortality prediction (see Results). The extensive feature extraction in deep learning models results in better performance for the targeted task and the discovery of new \emph{non-targeted} or \emph{passive} digital bio-markers for various diseases, thereby improving healthcare provision. This disclosure of non-targeted information, however, violates the privacy of the patients and poses an ethical dilemma. 

\begin{figure}[t]
\centering
\includegraphics[scale=0.34]{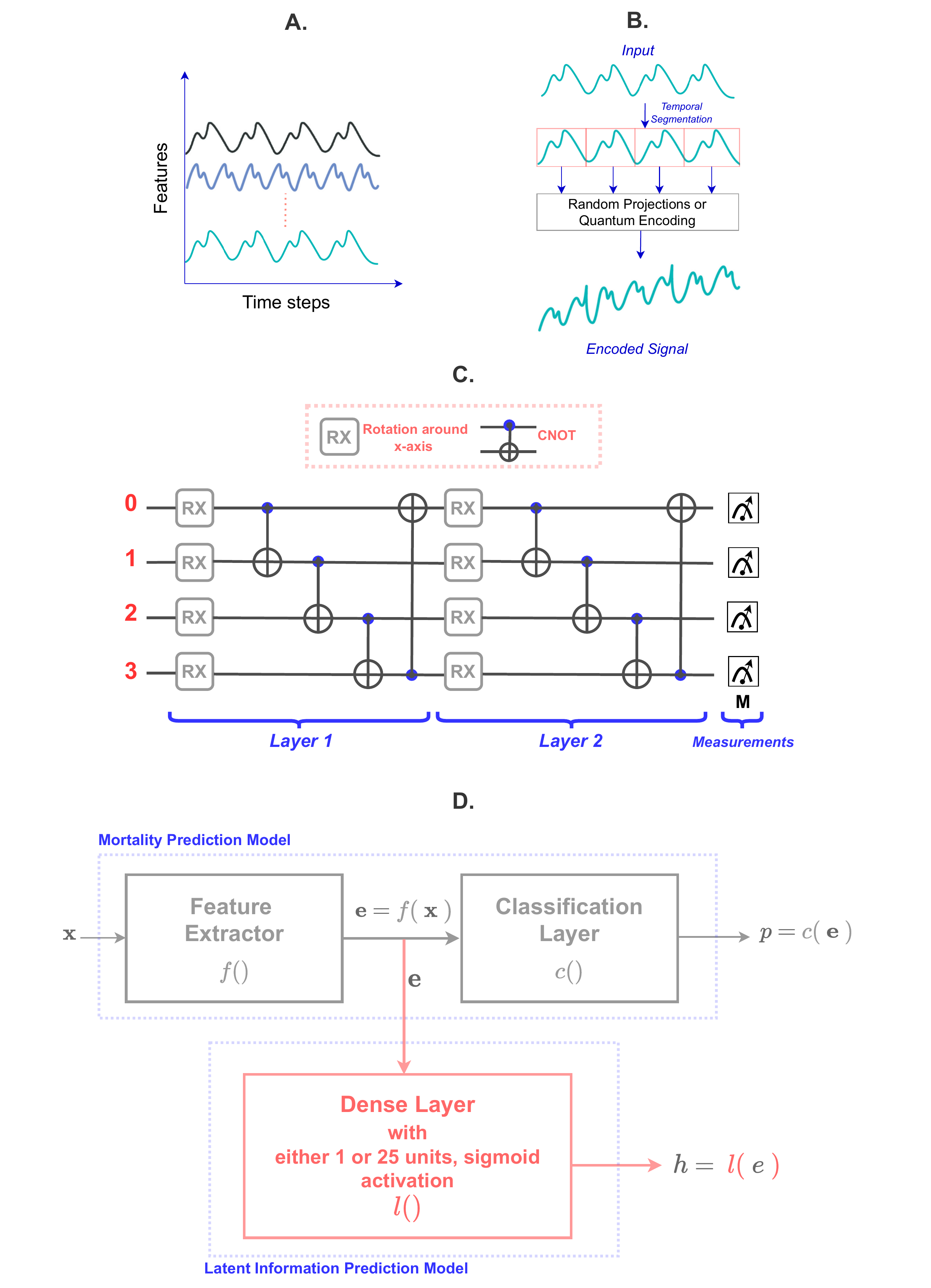}
\caption{\textbf{A.}~Conceptual rendition of a multi-variate time-series as a collection of multiple $1$-d signals.~\textbf{B.} Illustration of the process of encoding one of the 1-d signals within a time-series using the proposed encoding framework.~\textbf{C.}~Illustration of a quantum circuit that is composed of four wires, unitary rotation gates, and controlled-NOT (CNOT) gates. ~\textbf{D.} Illustration of the setup used for evaluating the latent information leakage from the trained mortality prediction models. Penultimate layer embedding from the trained mortality prediction models is given as input to a linear or dense layer dealing with either gender or patient disorders predictions.}
\label{fig:intro}
\end{figure}

\vspace{0.1cm}
Deep learning models can be seen as a combination of feature extraction layers mapping an input example to a compressed, semantic representation or embedding and the last classification layer mapping the embedding to the model output or predictions (Fig.~\ref{fig:intro}D). According to the information bottleneck (IB) principle, an ideal model should minimize mutual information between input and embedding while maximizing it between embedding and the model output \cite{pan2021disentangled,tishby2000information}. In other words, the embedding extracted by the model should only contain task-specific information and must strip spurious or non-task-related information that might be present in the input. To avoid latent information leakage, clinical deep learning models should be designed or trained to follow the IB principle and must only extract the ``relevant'' information from the input patient data. 

\vspace{0.1cm}
This paper argues that encoding healthcare data can achieve both data democratization and latent information leakage prevention simultaneously. To accomplish this, we envision an encoding framework transforming pre-processed and anonymized longitudinal health records or multi-variate time-series data into a new space. This encoding framework should meet the following requirements:

\begin{itemize}
\setlength{\itemsep}{0.5\baselineskip}
    \item \textbf{\emph{One-way transformation:}} The recovery of the original data from its encoded version should either be impossible or extremely computationally challenging. 
    
    \item \textbf{\emph{Imperceptibility of the encoded data:}} The encoded data should be a highly convoluted version of the original data. It should not be possible to infer any information about the original data just by performing a simple manual or computational analysis of its encoded version. Feature scaling or normalization, for example, cannot be considered a viable method of encoding information. 
    
    \item \textbf{\emph{Semantic preservation:}} The encoding framework must preserve the semantic characteristics of the original data to a large extent so that deep learning models can be trained effectively over the encoded data. In theory, the performance of models based on original data and encoded data should be the same.
\end{itemize}
The realization of this envisioned framework will enable the sharing of encoded healthcare data without violating privacy constraints. Ideally, encoded data is imperceptible, and the encoding process is practically irreversible. Therefore, it is very unlikely that any sensitive patient information can be derived from encoded data by either a manual or computational inspection. Nevertheless, there is an obvious trade-off between the imperceptibility and semantic preservation requirements of the envisioned encoding framework. A better semantic preservation results in lesser imperceptibility and vice versa. As a result, the encoded data can be seen as a ``deformed'' version of the original data, and much higher computational effort is required to extract its semantic characteristics. This nature of encoded data results in inherent regularization during model training and indirectly enforces the IB principle (see Results) to prevent latent information leakage.

This paper exploits \emph{random projections} \cite{bingham2001random,vempala2005random} and \emph{random quantum circuits} \cite{yang2021decentralizing,henderson2020quanvolutional} as information processing tools to achieve the desired encoding framework for the multivariate time-series data. Both random quantum circuits and random projections can deform or project the data to a space where it becomes imperceptible. By exploiting random projections or random quantum circuits, the proposed encoding framework performs piece-wise or segment-wise temporal encoding of each feature or each $1$-d signal of a multivariate time series (Figure \ref{fig:intro}B). Since there is no interference among features or signals of the original time series, the resulting encoded time series retains its semantic characteristics. However, random transformations deform each segment of a signal to make them incomprehensible. Due to the fact that the original data, encoding method, transformation matrix (used for random projections), and random quantum circuit will not be made public, it is extremely difficult to reverse the encoding process. Hence, data democratization can be achieved by sharing this encoded data among deep learning researchers. Additionally, higher model complexity is required to extract the relevant semantic information from the deformed or encoded data resulting in regularisation and thus enforcing the IB principle.

\begin{table}[t]
\centering
\caption{Characteristics of MIMIC-III, PhysioNet, and eICU datasets.}
\label{tab:data}
\resizebox{.95\textwidth}{!}{
\begin{tabular}{c||c||c||c||c||c}
\specialrule{.15em}{.12em}{.12em}
\textbf{Dataset} & \textbf{\begin{tabular}[c]{@{}c@{}}\#ICU \\ Stays\end{tabular}} & \textbf{\begin{tabular}[c]{@{}c@{}}\# Positive\\  Cases\end{tabular}} & \textbf{\begin{tabular}[c]{@{}c@{}} Feature \\ Dimensions\end{tabular}} & \textbf{Time-steps} & \textbf{\begin{tabular}[c]{@{}c@{}}\# Train, Validation \\ and Held-out\\ Examples\end{tabular}} \\ \specialrule{.15em}{.12em}{.12em}
MIMIC-III  & 21,156  & 2,799  & 60  & 48  & 14698, 3222, 3236 \\ \specialrule{.1em}{.05em}{.05em}
PhysioNet  & 8,000   & 1,122  & 44  & 48  & 5120, 1280, 1600  \\ \specialrule{.1em}{.05em}{.05em}
eICU    & 122,588    & 57,434 & 284 & 12  & 68648, 17163, 36777 \\ \specialrule{.15em}{.12em}{.12em}
\end{tabular}}
\end{table}

\section{Results}
\label{sec:res}

\subsection{Designed experiments for the performance evaluation}
The proposed encoding framework is evaluated using publicly available datasets: \textbf{1.)} \emph{PhysioNet 2012 challenge} \cite{silva2012predicting}, \textbf{2.)}  \emph{MIMIC-III} \cite{johnson2016mimic,harutyunyan2019multitask} and \textbf{3.)}  \emph{eICU-CRD} \cite{pollard2018eicu, tang2020democratizing}. Both \emph{PhysioNet} and \emph{MIMIC-III} deal with in-hospital mortality prediction based on the first 48 hours of ICU stay. Similarly, \emph{eICU-CRD} is used for the task of acute respiratory failure (ARF) prediction based on the first 12 hours of ICU stay. Each ICU stay is represented by a time series with $48$ and $12$ time steps (separated by $1$ hour) for mortality and ARF prediction, respectively. Each step is represented by a $44$, $60$, and $284$-dimensional feature vector in PhysioNet, MIMIC-III, and eICU datasets, respectively. Table \ref{tab:data} documents the total number of ICU stays or examples available in each dataset. In addition to the clinical features and task labels, meta-data about the patients corresponding to ICU stays are also available. This includes gender information in all datasets as well as chronic, acute, and mixed conditions afflicting the patients in MIMIC-III and information about the ethnicity of the
patients in eICU. More details about the clinical features representing time series in all datasets can be found in the supplementary document. The following experiments are designed to evaluate different aspects of the proposed framework:

\begin{itemize}
\setlength{\itemsep}{0.5\baselineskip}
    \item On the original as well as on the encoded data, we train $5$ different neural networks on each dataset and compare their relative performances. These models include long short-term memory (LSTM) \cite{yu2019review}, temporal $1$-D convolutions \cite{bai2018empirical}, multi-resolution temporal convolutions \cite{martinez2020lipreading}, transformer \cite{vaswani2017attention}, and vision transformer \cite{dosovitskiy2020image}. More details can be found in Section \ref{sec:methods}.
    \item To assess the latent information leakage from the trained models, a single dense layer mapping the penultimate layer embedding to the patient information is used. For the MIMIC-III dataset, gender and $25$ ``latent'' or non-targeted patient disorders (acute, chronic, and mixed) are predicted from the penultimate layer embedding of the trained mortality prediction models. For PhysioNet, we only predict gender as the latent information. 
    
    \begin{figure}[H]
\centering
\begin{subfigure}{.495\textwidth}
\centering
\includegraphics[scale=0.245]{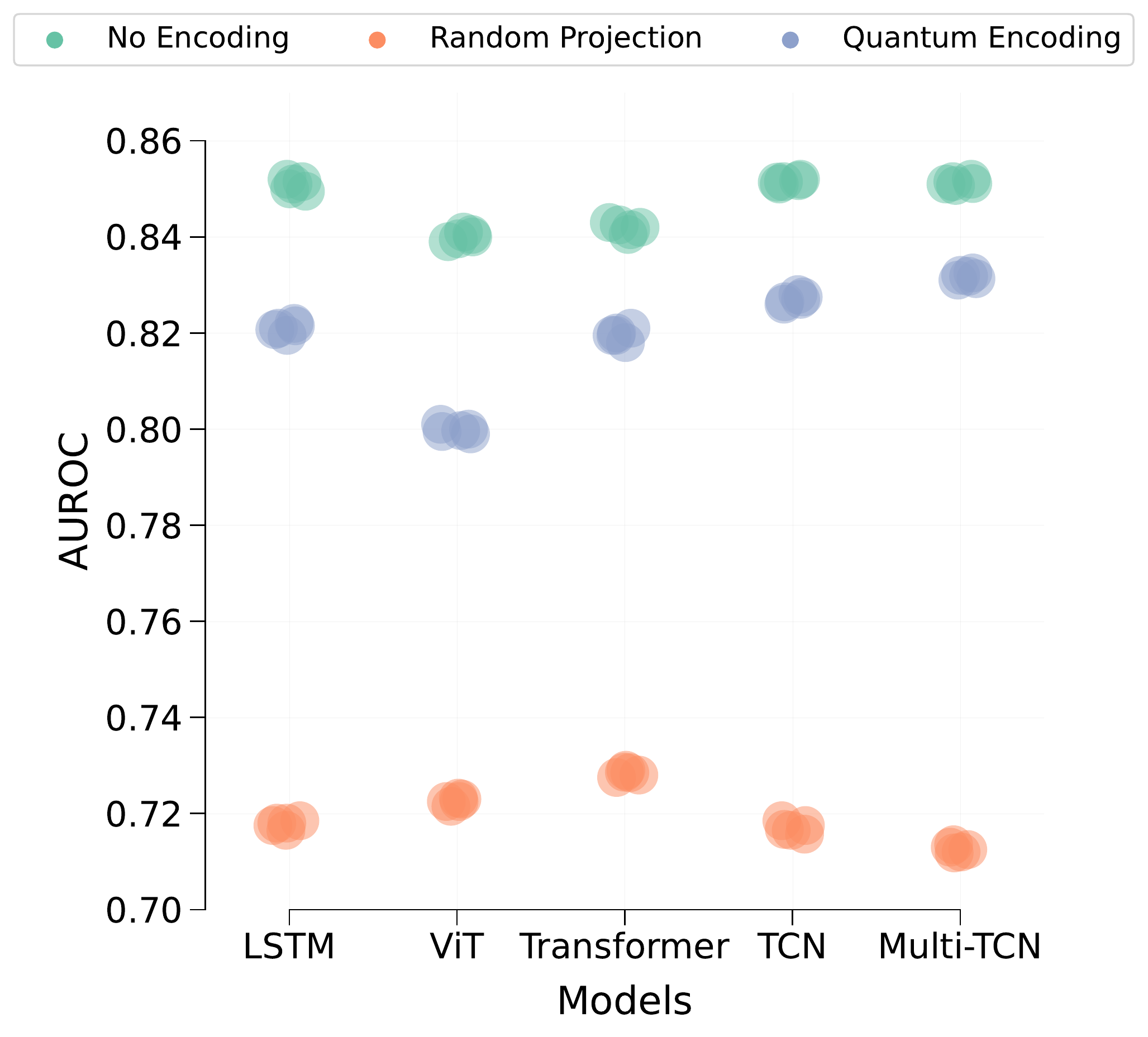}
\caption{Models'~performance on MIMIC-III.}
\end{subfigure}
\begin{subfigure}{.495\textwidth}
\centering
\includegraphics[scale=0.27]{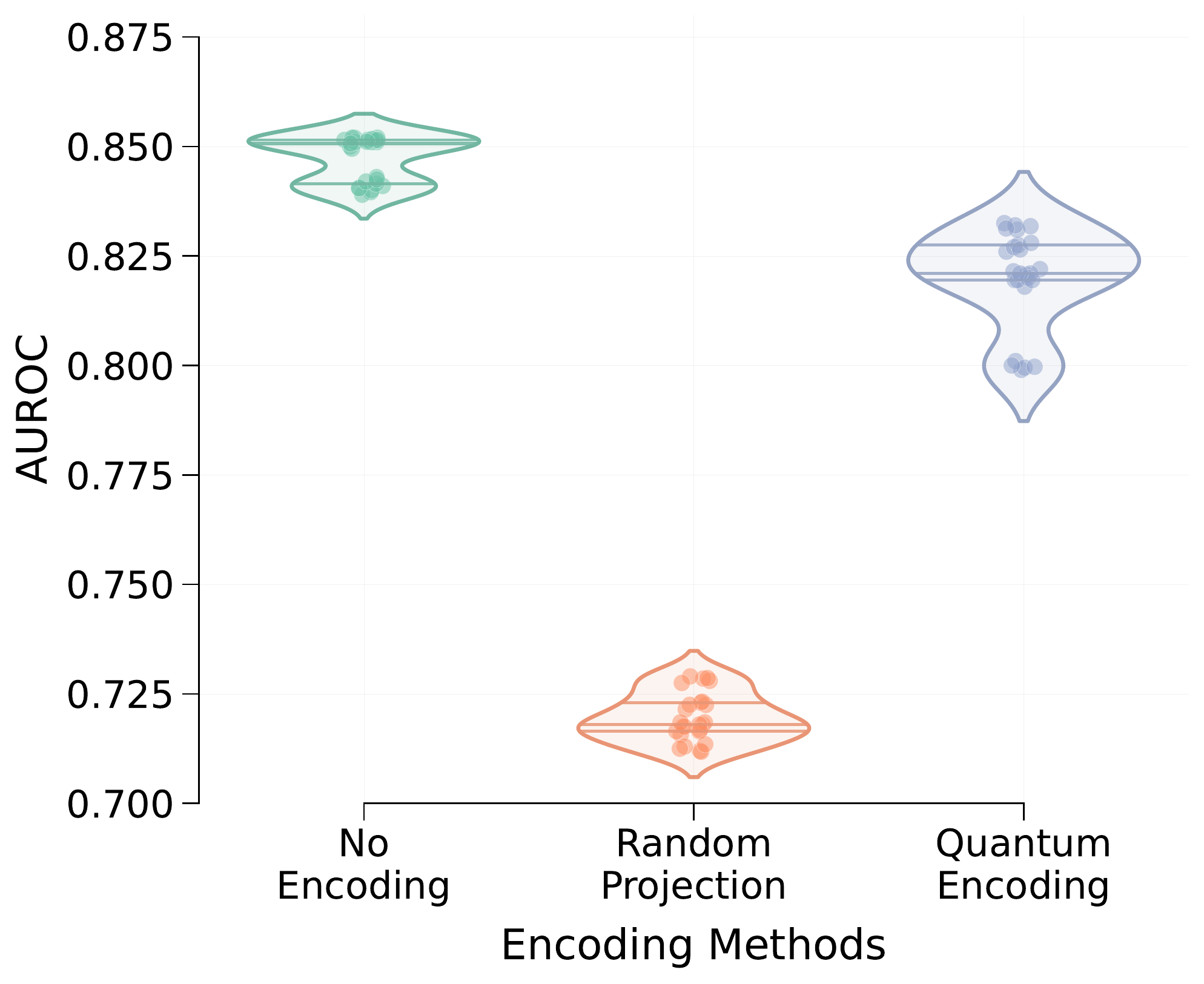}
\vspace{1.75mm}
\caption{Aggregate MIMIC-III performance.}
\end{subfigure}

\begin{subfigure}{.495\textwidth}
\centering
\includegraphics[scale=0.245]{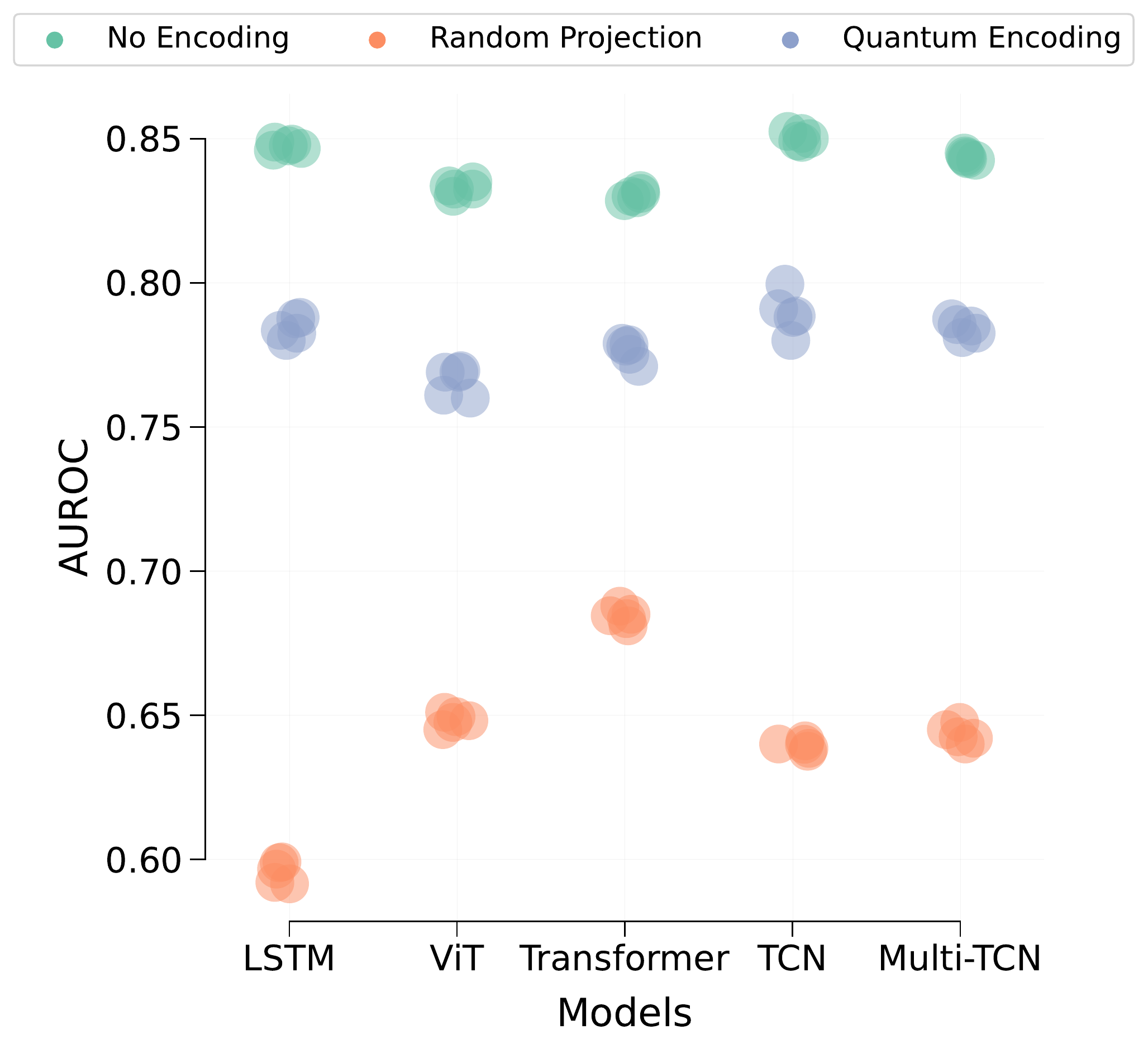}
\caption{Models'~performance on PhysioNet.}
\end{subfigure}
\begin{subfigure}{.495\textwidth}
\centering
\includegraphics[scale=0.27]{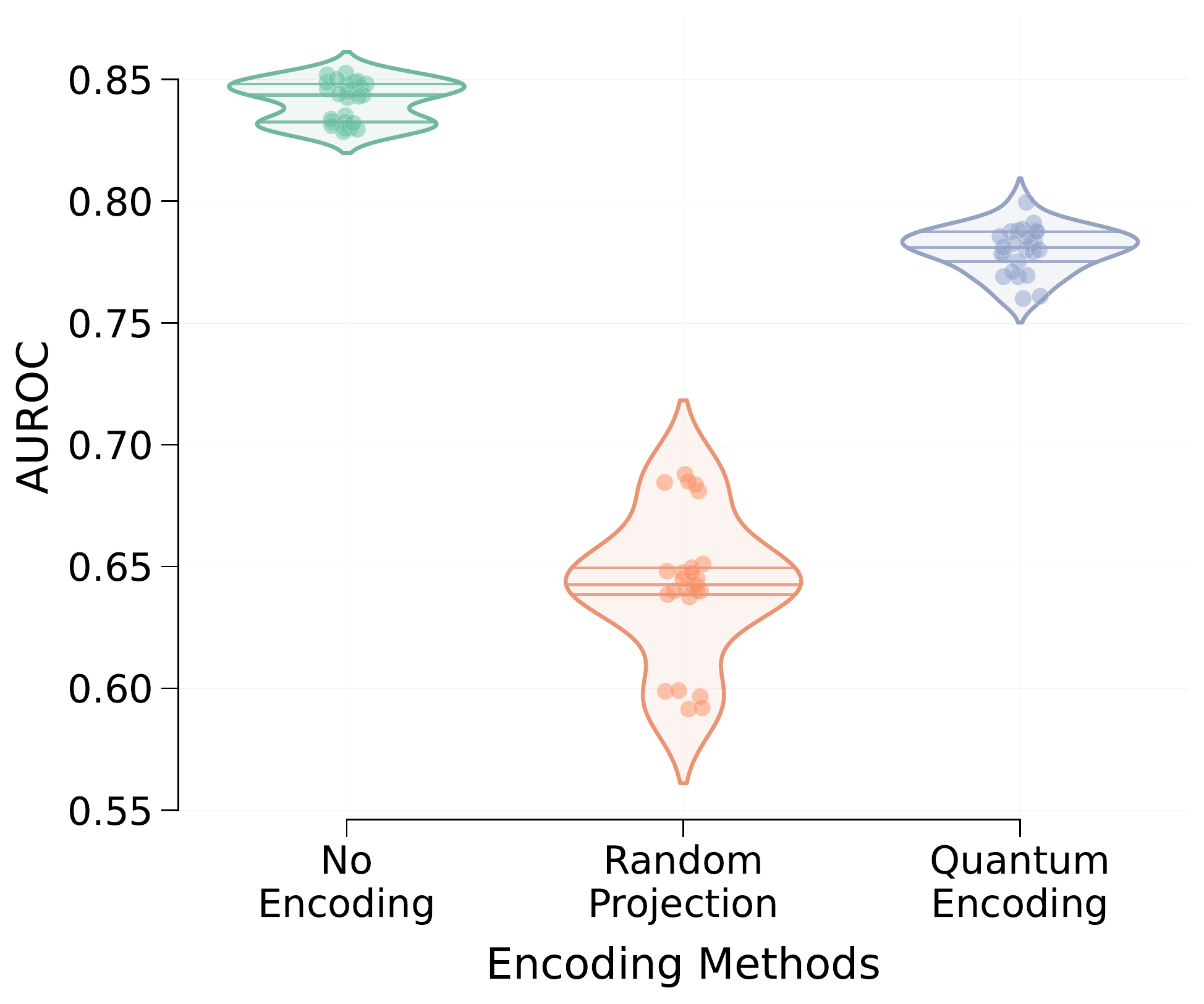}
\vspace{1.75mm}
\caption{Aggregate PhysioNet performance.}
\end{subfigure}

\begin{subfigure}{.495\textwidth}
\centering
\includegraphics[scale=0.245]{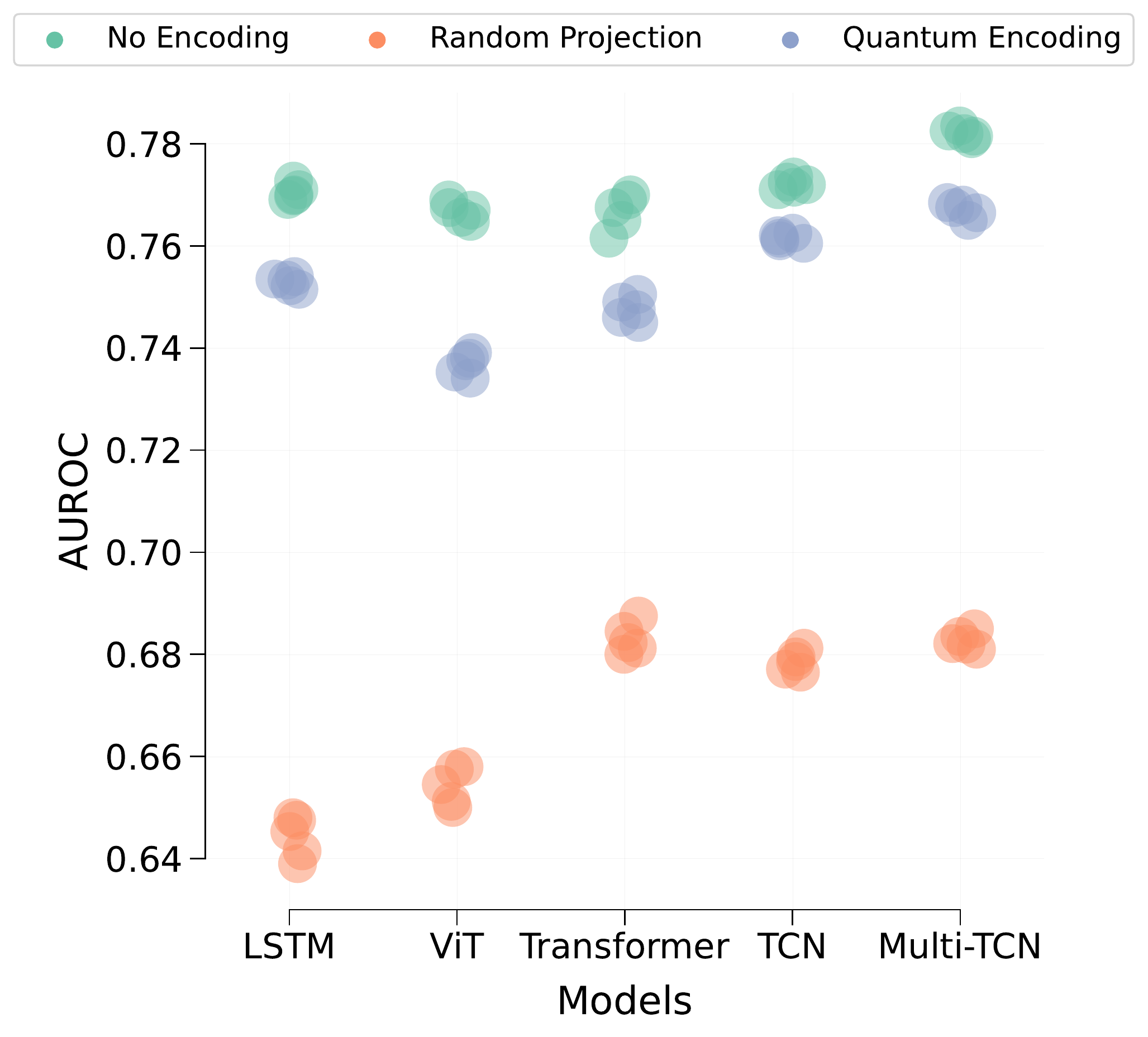}
\caption{Models'~performance on eICU.}
\end{subfigure}
\begin{subfigure}{.495\textwidth}
\centering
\includegraphics[scale=0.27]{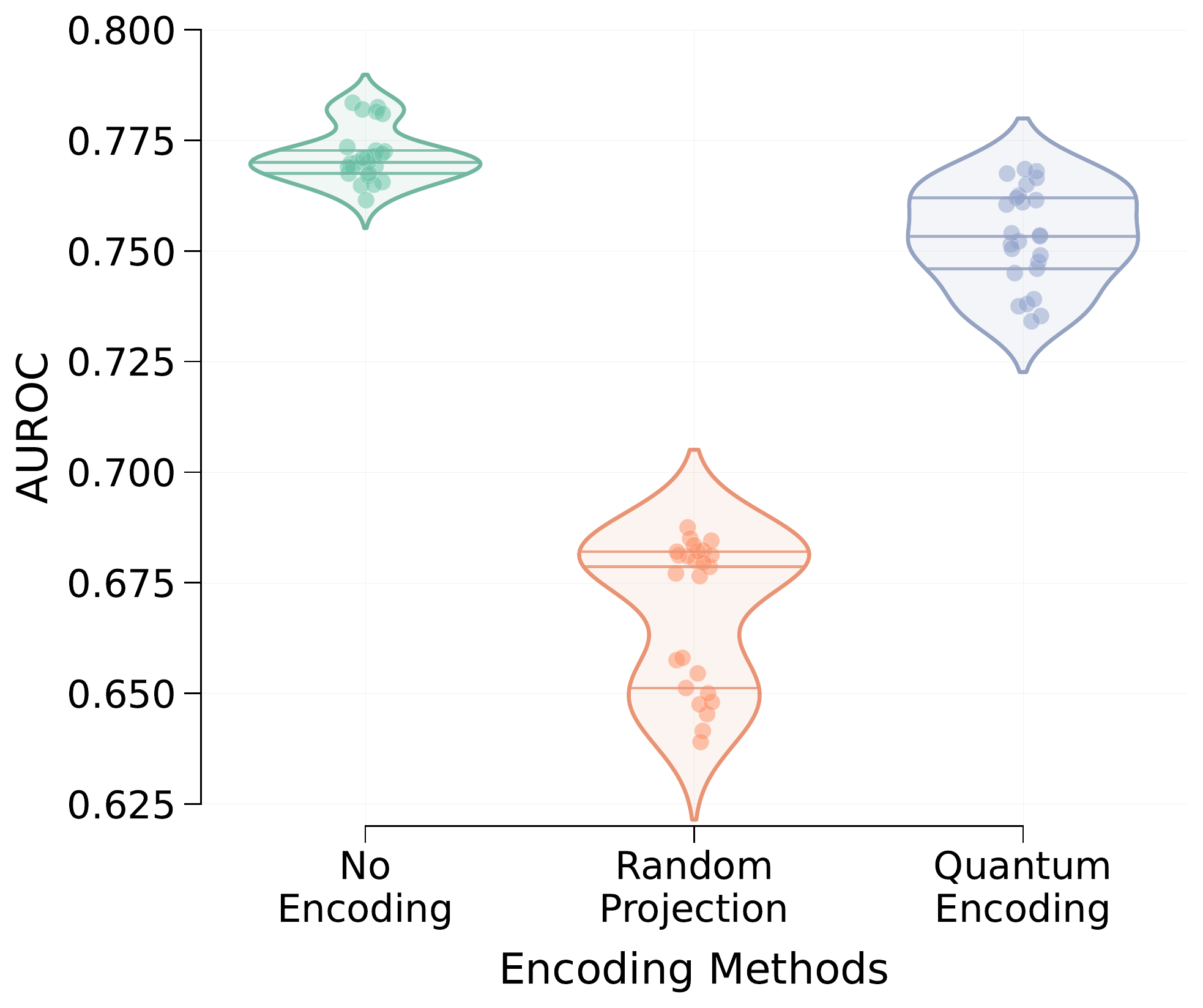}
\vspace{1.75mm}
\caption{Aggregate eICU performance.}
\end{subfigure}
\caption[short]{Performance of LSTM, vision transformer (ViT), transformer, temporal convolutional network (TCN) and multi-branch temporal convolutional network (Multi-TCN) on  \textbf{(a)} MIMIC-III, \textbf{(c)} PhysioNet and \textbf{(e)} eICU, respectively. Performance as a function of encoding methods across different models on \textbf{(b)} MIMIC-III, \textbf{(d)} PhysioNet and \textbf{(f)} eICU, respectively.}
\label{fig:res1}
\end{figure}

    \noindent Similarly, we predict the gender and ethnicity of patients from the trained ARF prediction models.
    
\end{itemize}
Since we are employing only a single linear layer to map embedding to either sex, ethnicity, or patient disorders (Fig.~\ref{fig:intro}D), no further feature transformations are employed. The performance of this latent information prediction depends entirely on the nature of embedding. More details about this experimental setup can be found in Section \ref{sec:methods}.  

\vspace{-0.25cm}
\subsection{Performance on the encoded time-series data}
The performance of different models on the encoded and the original datasets is depicted in Fig.~\ref{fig:res1}. An analysis of this figure highlights the following:

\begin{itemize}
    \itemsep 1em
\item Across all datasets, the performance of models trained and evaluated on the original data is superior to that of the models dealing with the encoded time-series data. Across all models on MIMIC-III, random quantum encoding and random projection-based encoding resulted in an average relative performance drop of $3.52\;(\pm1.25)\%$ and $15.29\;(\pm2.51)\%$, respectively. The average relative performance drop of $5.13\;(\pm1.94)\%$ and $22.44\;(\pm4.75)\%$ are observed in PhysioNet dataset. Similarly, a drop of $2.13\;(\pm1.59)\%$ and $12.45\;(\pm2.29)\%$ was observed for the eICU dataset. This drop is expected as data encoding deforms the time series to preserve patient information. 
    
\item Despite the performance drop exhibited by models trained on the encoded data, these models (especially models trained on the quantum encoded data) seem to be effective in performing the target task. This shows that the encoding framework, either using random projection or random quantum encoding, can retain essential semantic characteristics in the \emph{deformed} encoded data.  

\item The performance of random quantum encoding is significantly better than random projections across all models and datasets. This shows that quantum encoding can preserve the semantic characteristics to a greater extent while deforming the data using random quantum operations.
\end{itemize}

\subsection{Latent information leakage from the trained models}
\subsubsection*{Patients' gender and ethnicity prediction}
\begin{figure}[H]
\centering
\begin{subfigure}{\textwidth}
\centering
\includegraphics[scale=0.26]{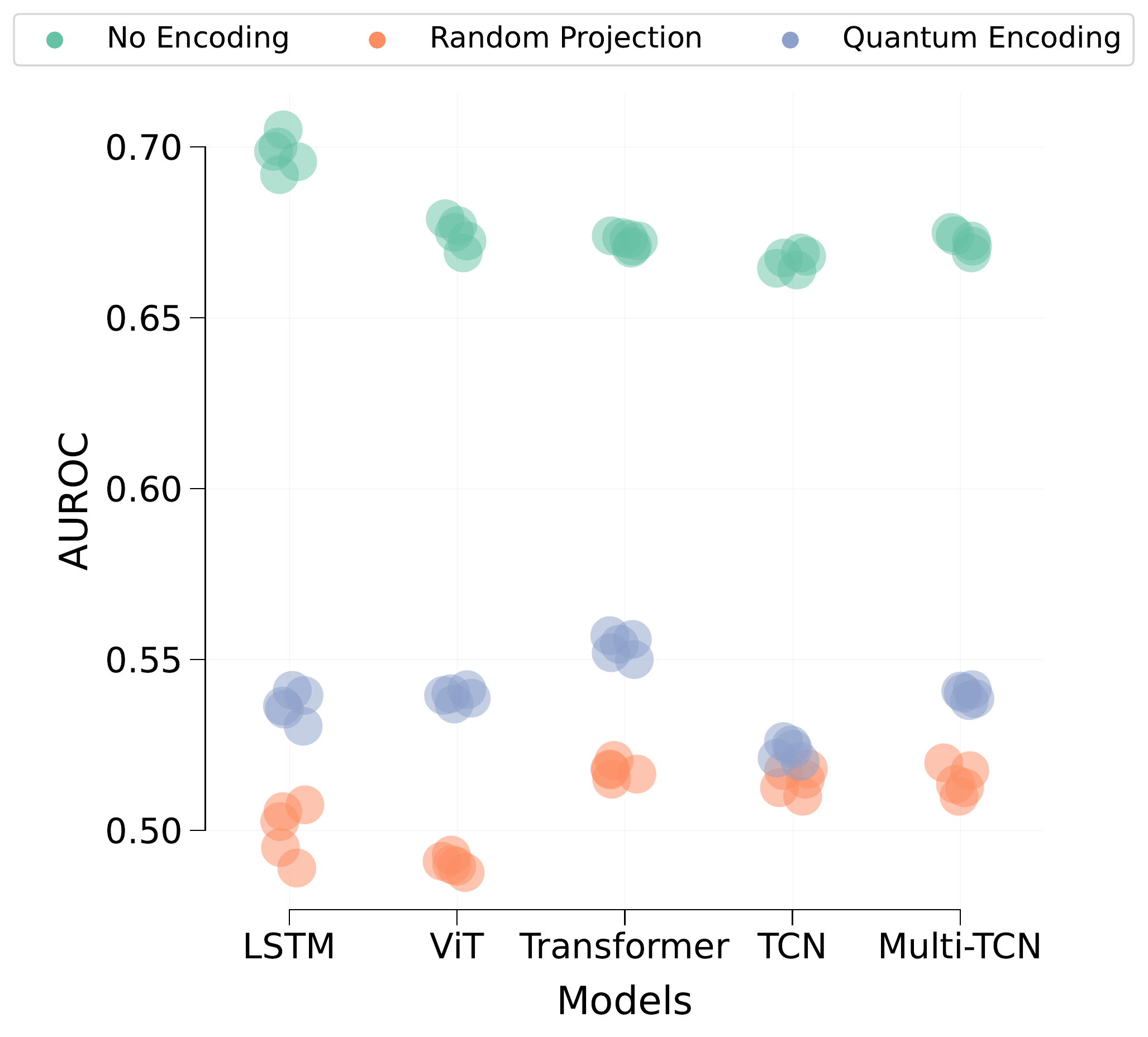}
\includegraphics[scale=0.26]{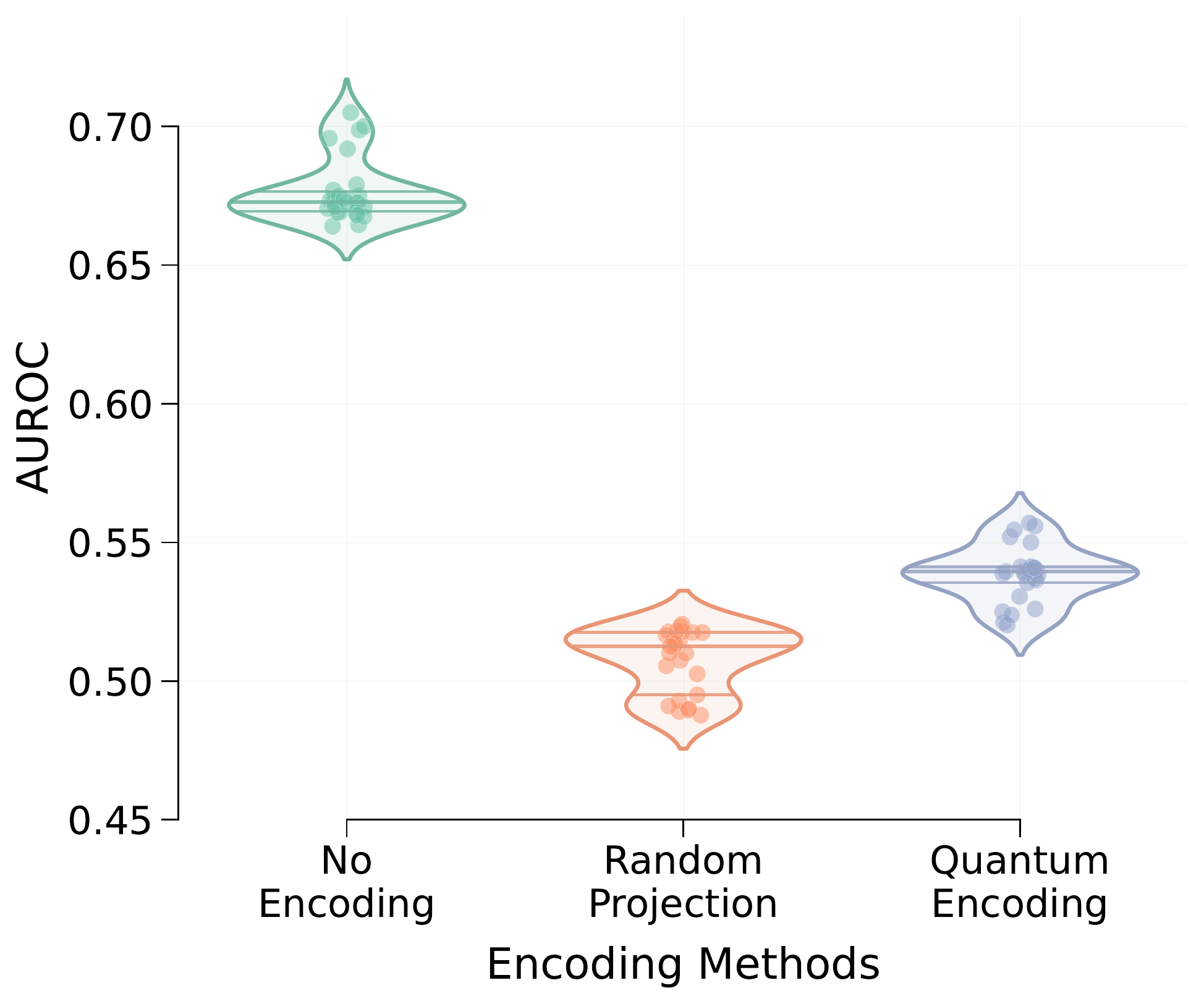}
\caption{Gender prediction from the embedding of MIMIC-III models.}
\end{subfigure}

\begin{subfigure}{\textwidth}
\centering
\includegraphics[scale=0.26]{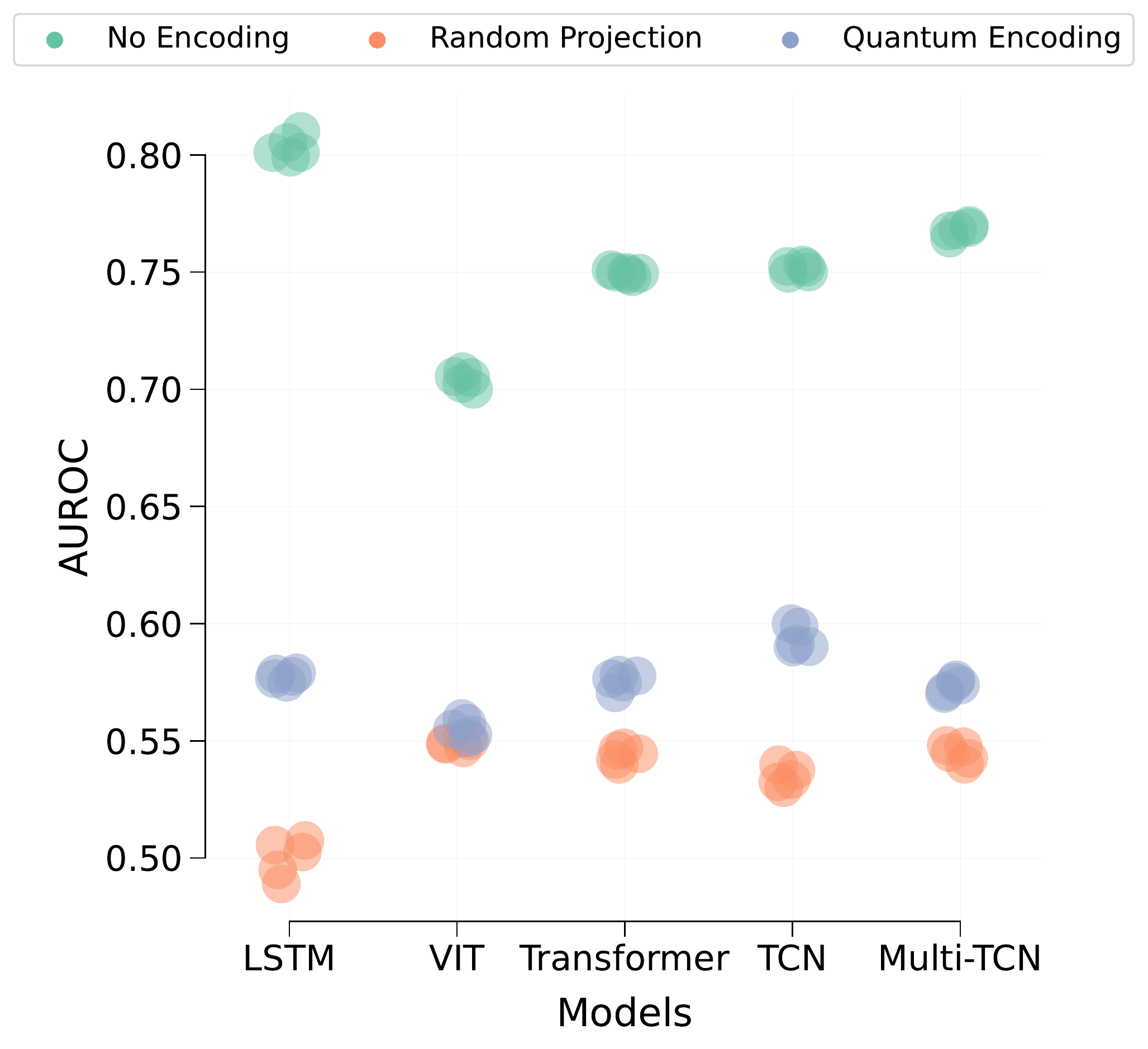}
\includegraphics[scale=0.26]{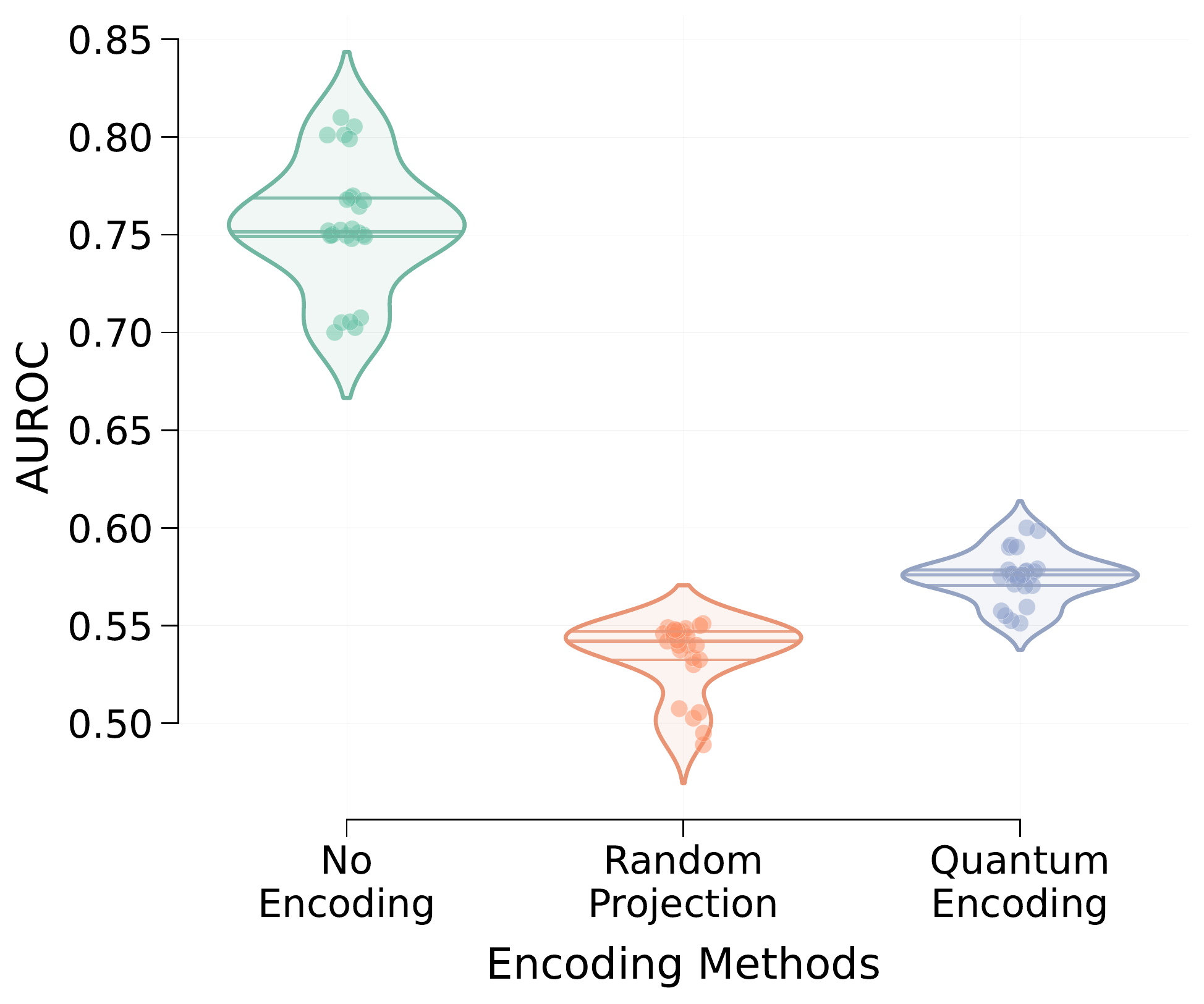}
\caption{Gender prediction from the embedding of PhysioNet models .}
\end{subfigure}

\begin{subfigure}{\textwidth}
\centering
\includegraphics[scale=0.26]{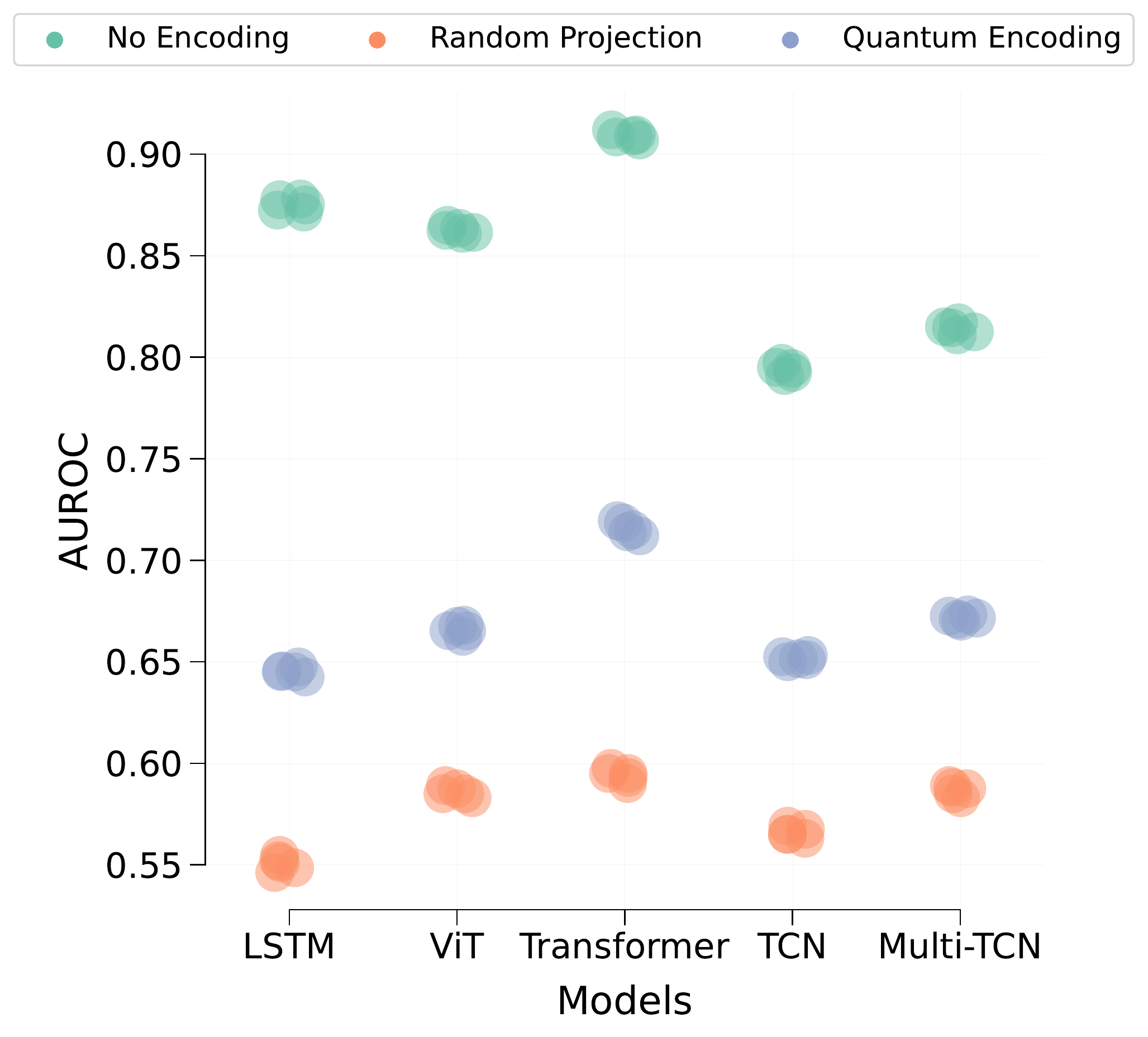}
\includegraphics[scale=0.26]{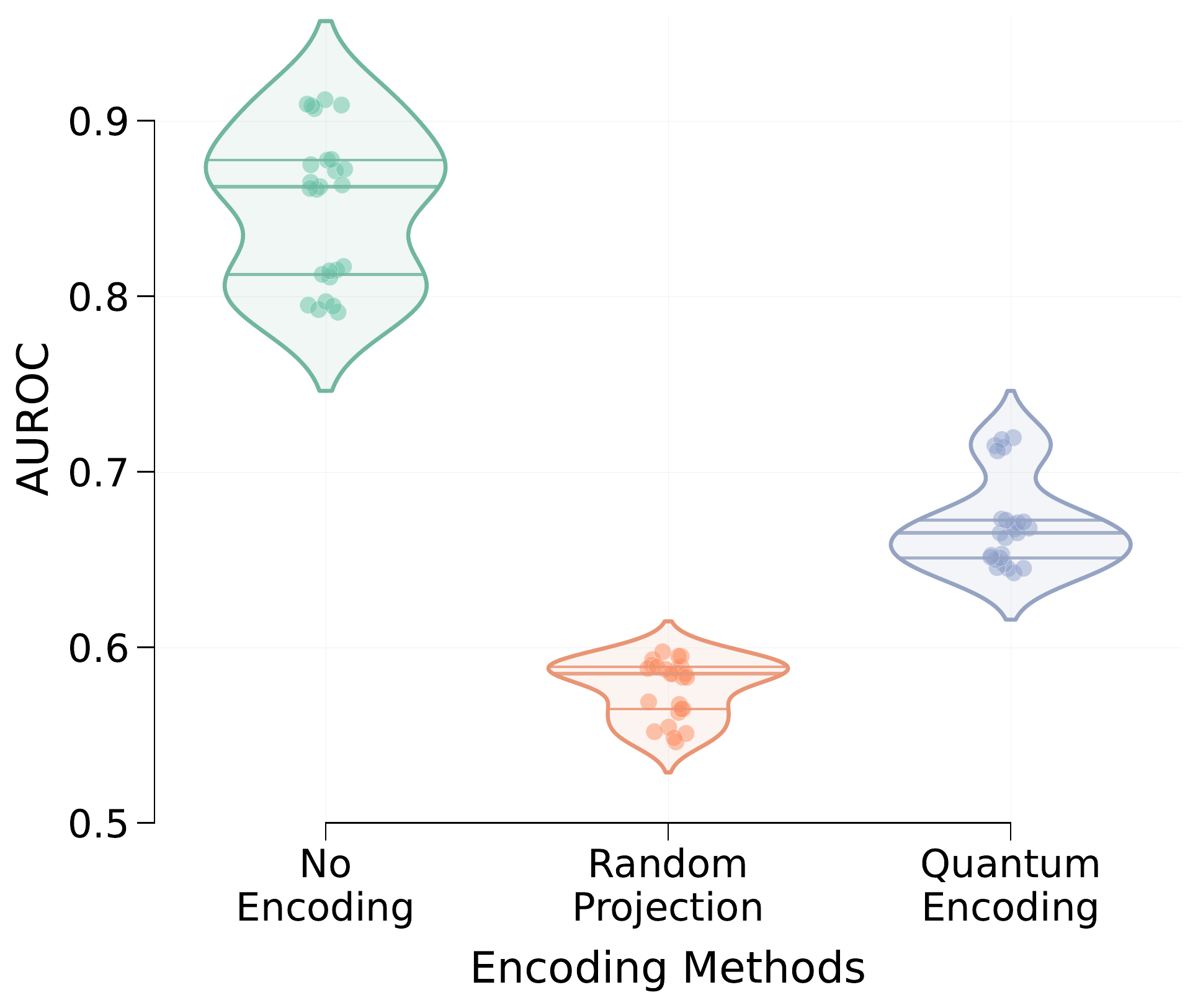}
\caption{Gender prediction from the embedding of eICU models.}
\end{subfigure}
\caption[short]{Model-specific and aggregated (across all models) performance for the task of patients' gender prediction using the penultimate embedding generated from different models trained on \textbf{(a)} MIMIC-III and \textbf{(b)} PhysioNet and \textbf{(c)} eICU datasets, respectively.}
\label{fig:res2}
\end{figure}

\begin{figure}[H]
\centering
\begin{subfigure}{\textwidth}
\centering
\includegraphics[scale=0.21]{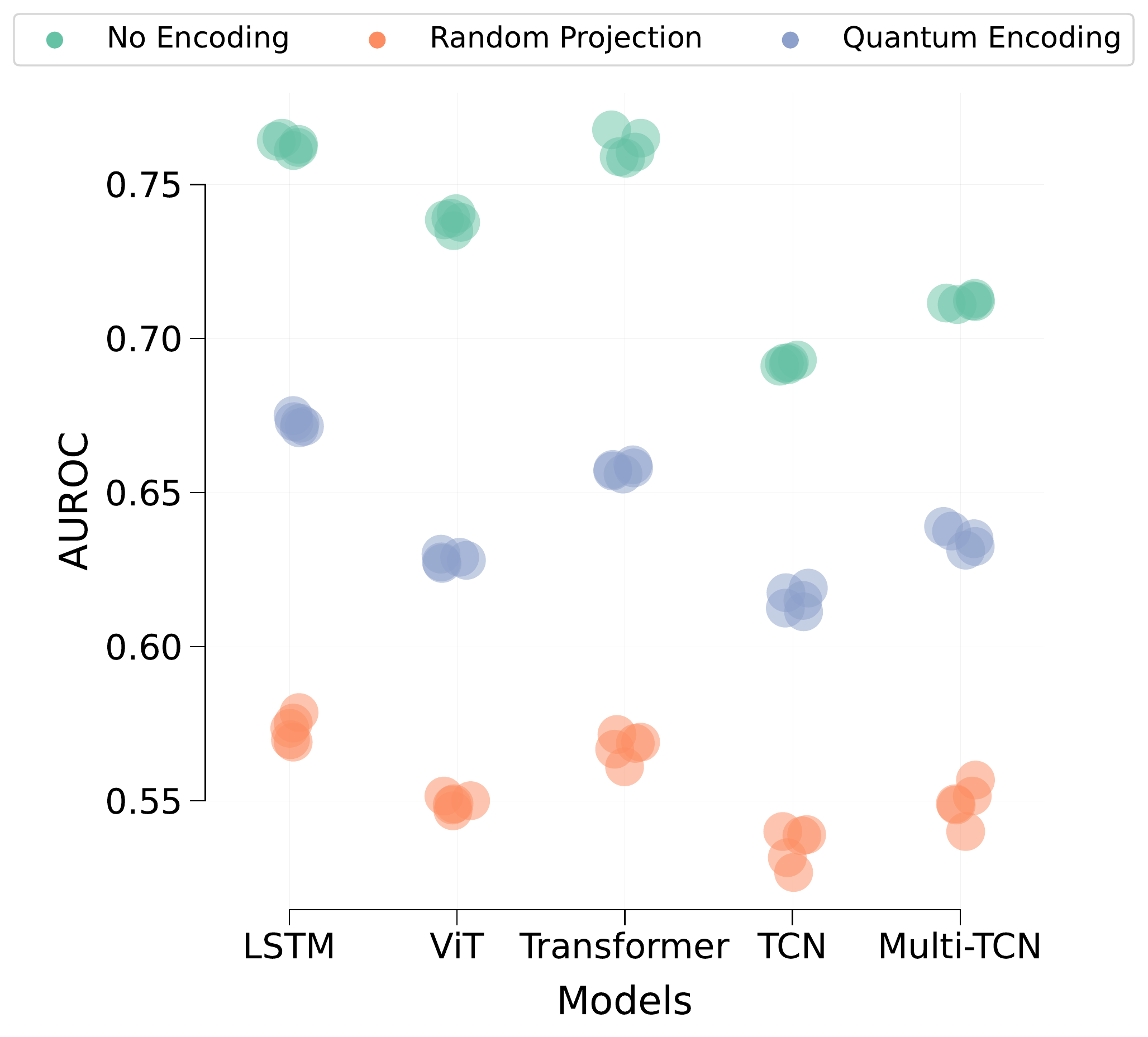}
\includegraphics[scale=0.21]{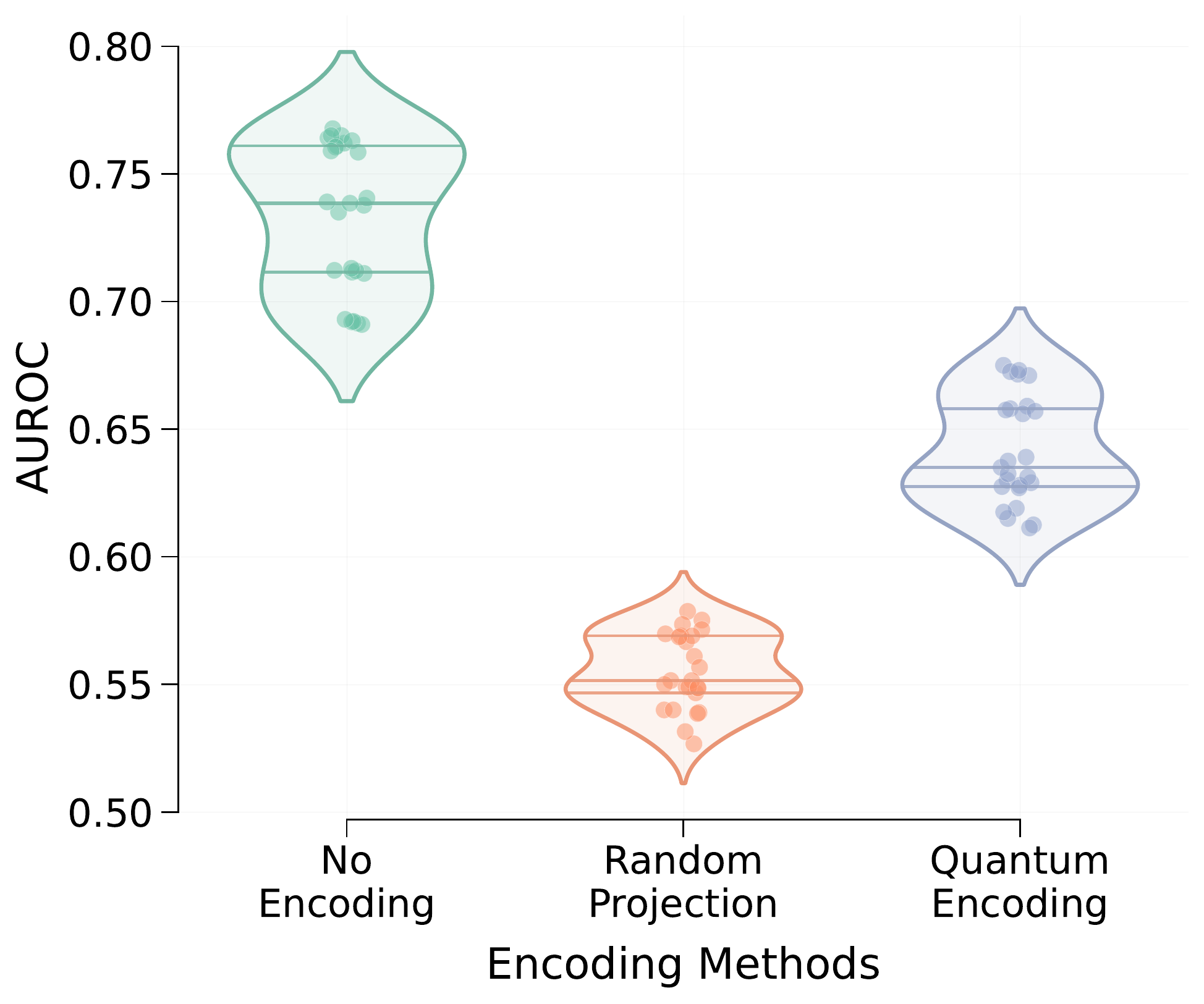}
\caption{Predicting if a patient is African-American.}
\end{subfigure}

\begin{subfigure}{\textwidth}
\centering
\includegraphics[scale=0.21]{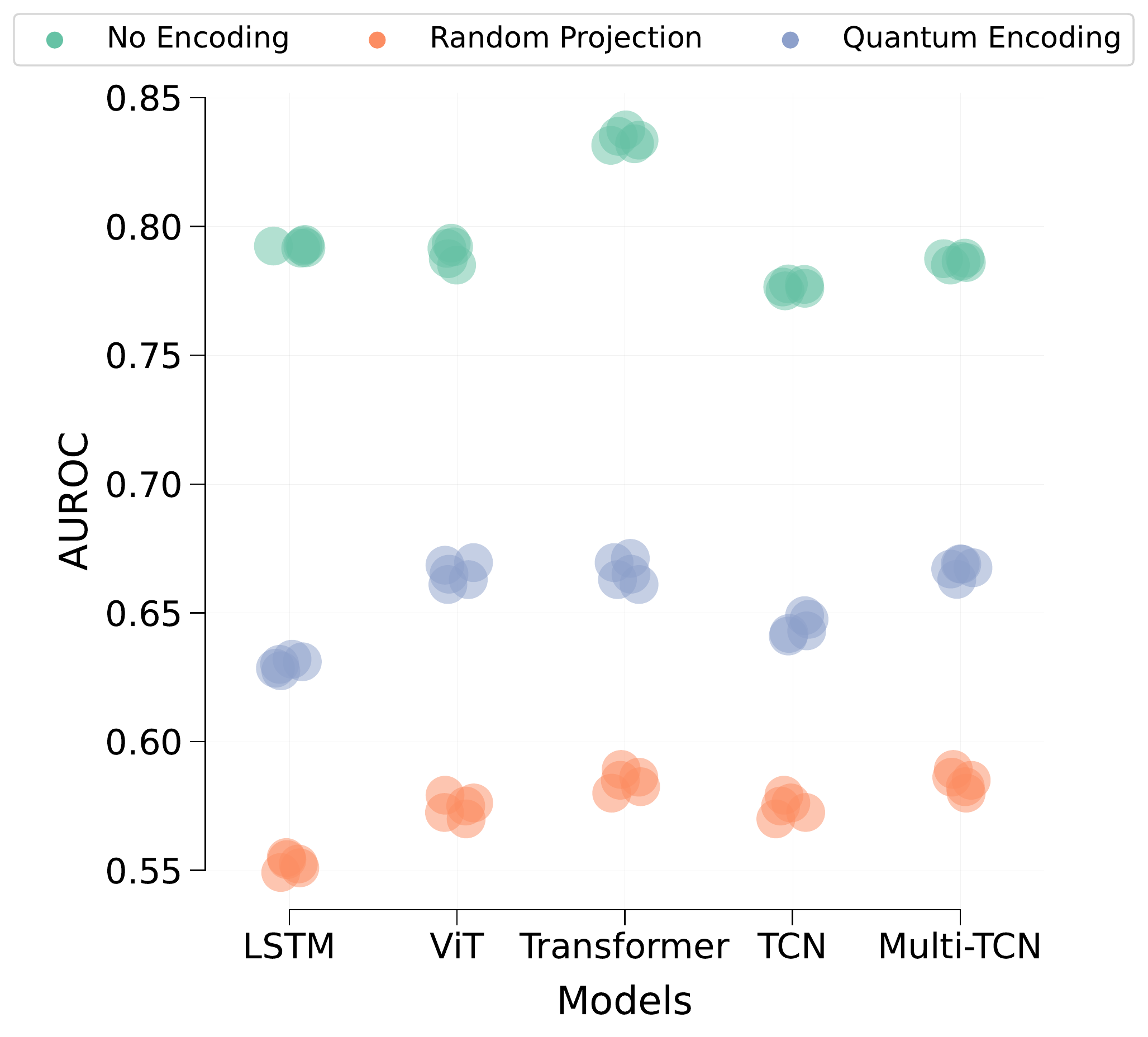}
\includegraphics[scale=0.21]{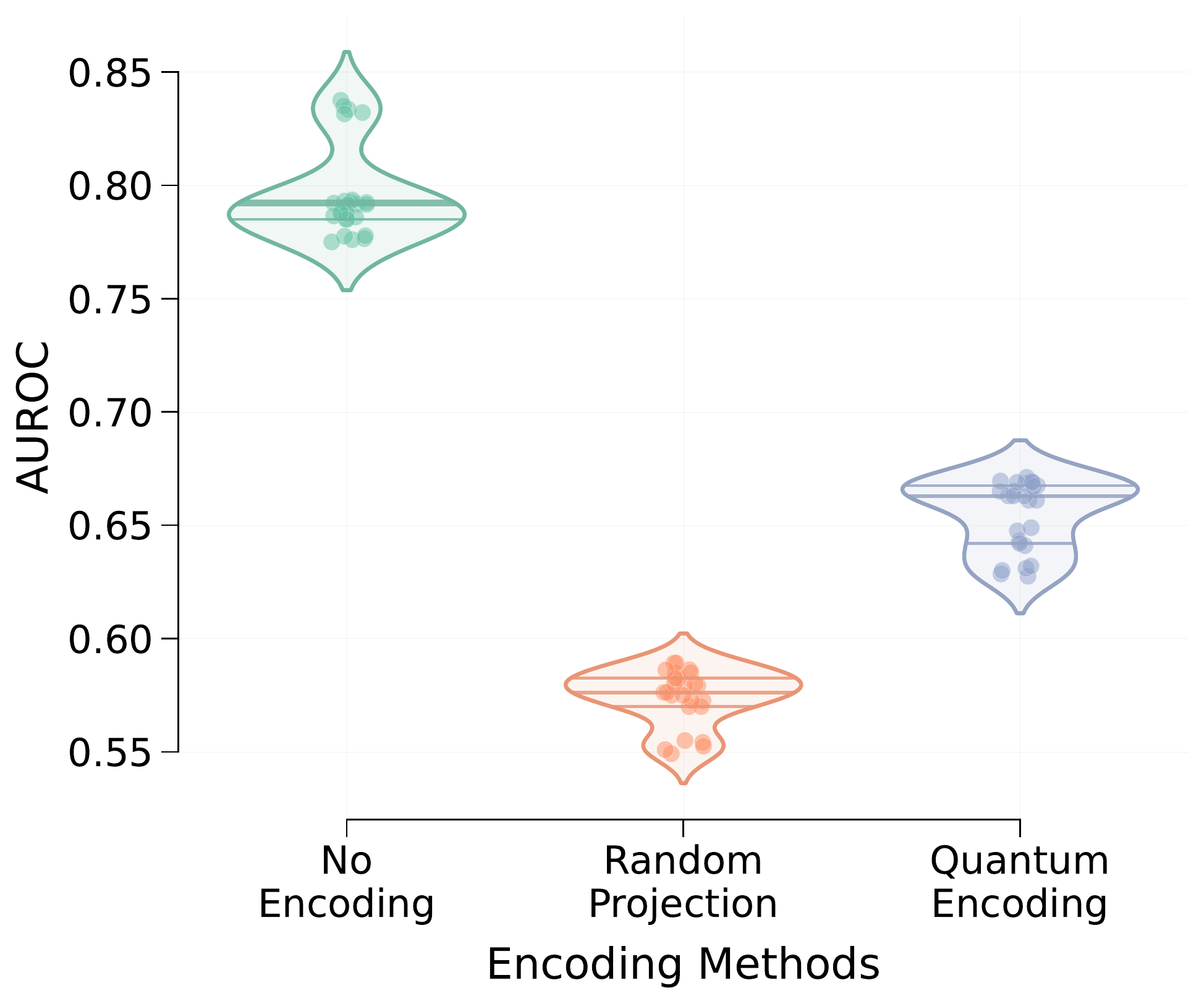}
\caption{Predicting if a patient is Asian.}
\end{subfigure}

\begin{subfigure}{\textwidth}
\centering
\includegraphics[scale=0.21]{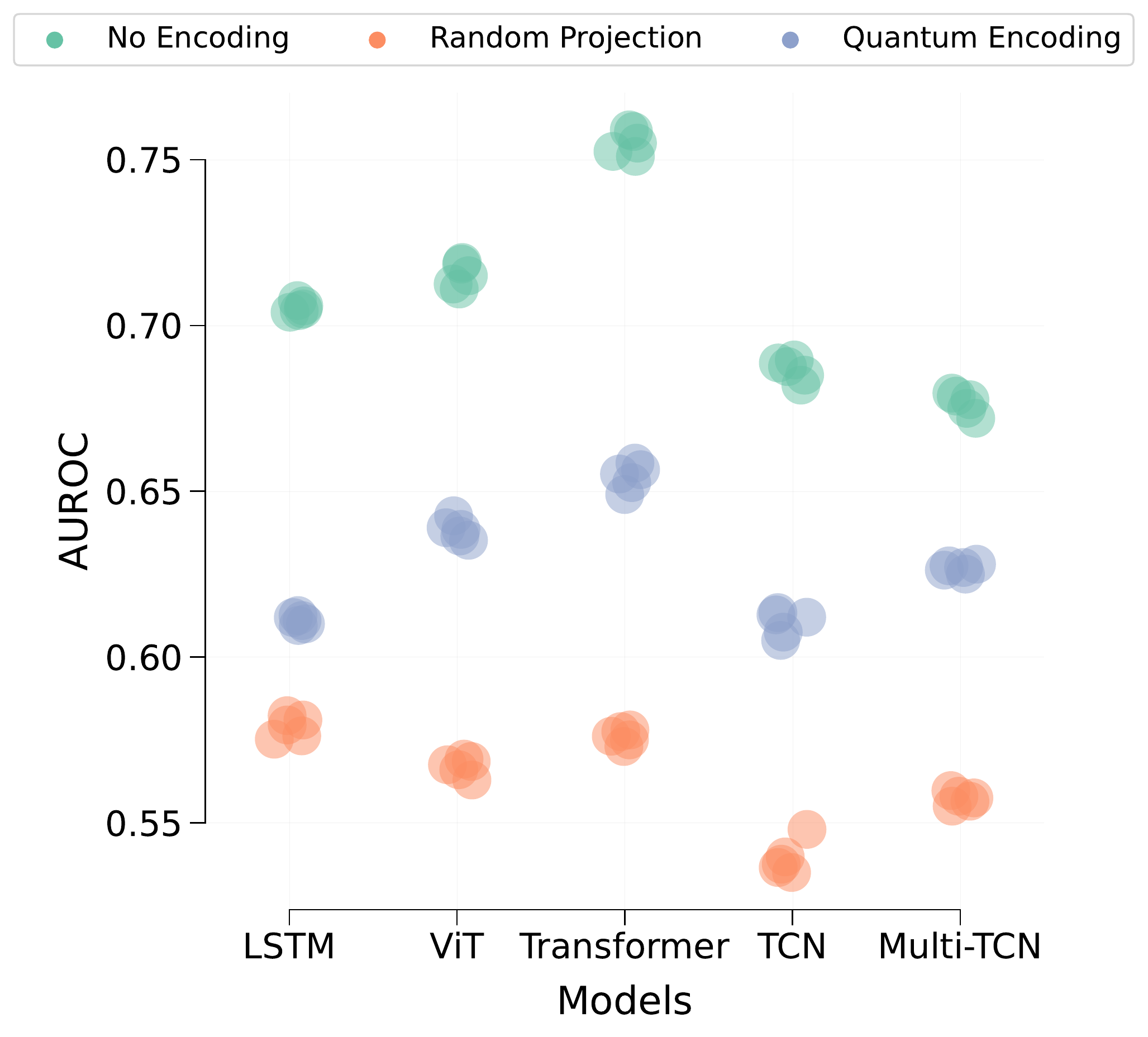}
\includegraphics[scale=0.21]{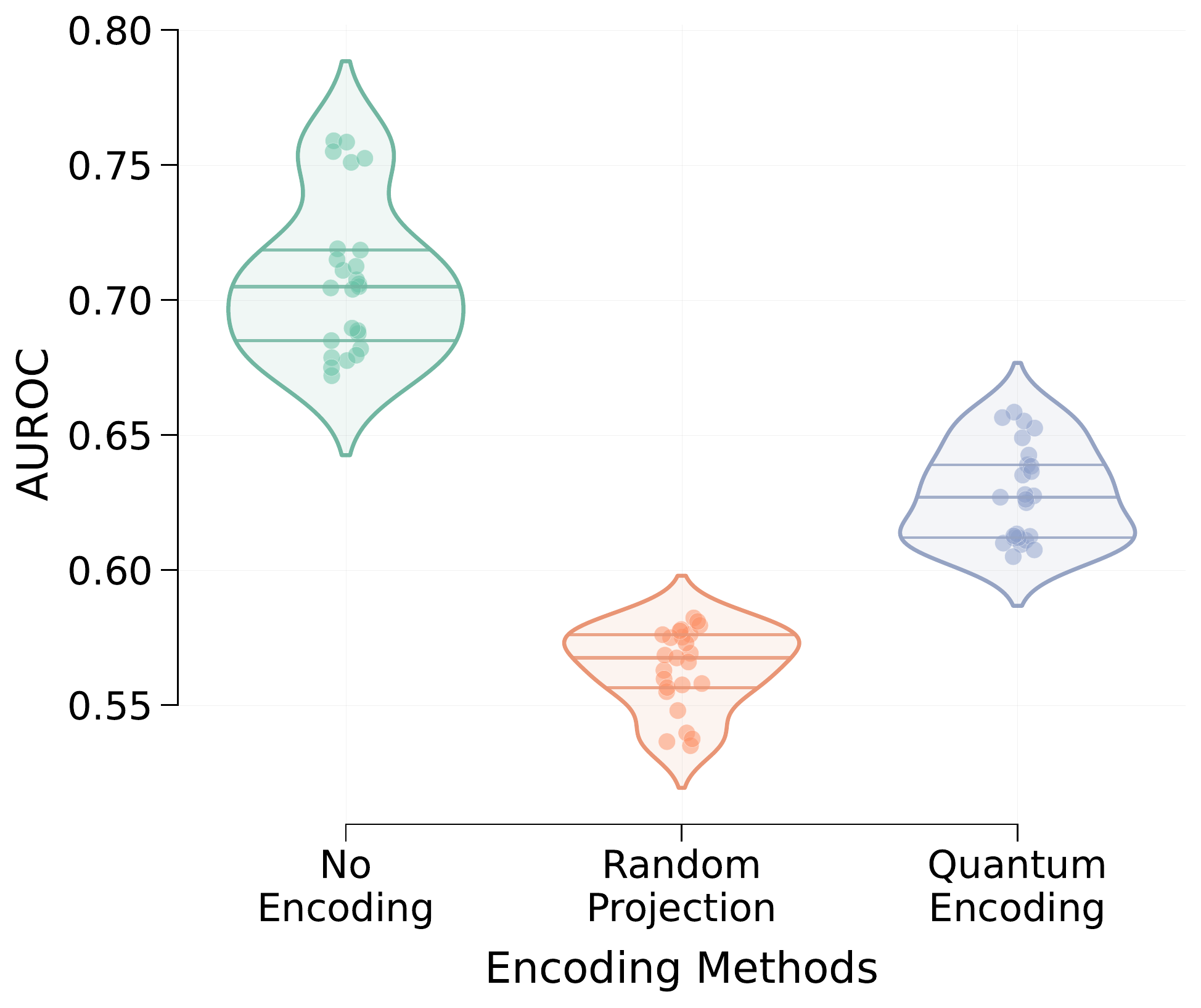}
\caption{Predicting if a patient is Caucasian.}
\end{subfigure}

\begin{subfigure}{\textwidth}
\centering
\includegraphics[scale=0.21]{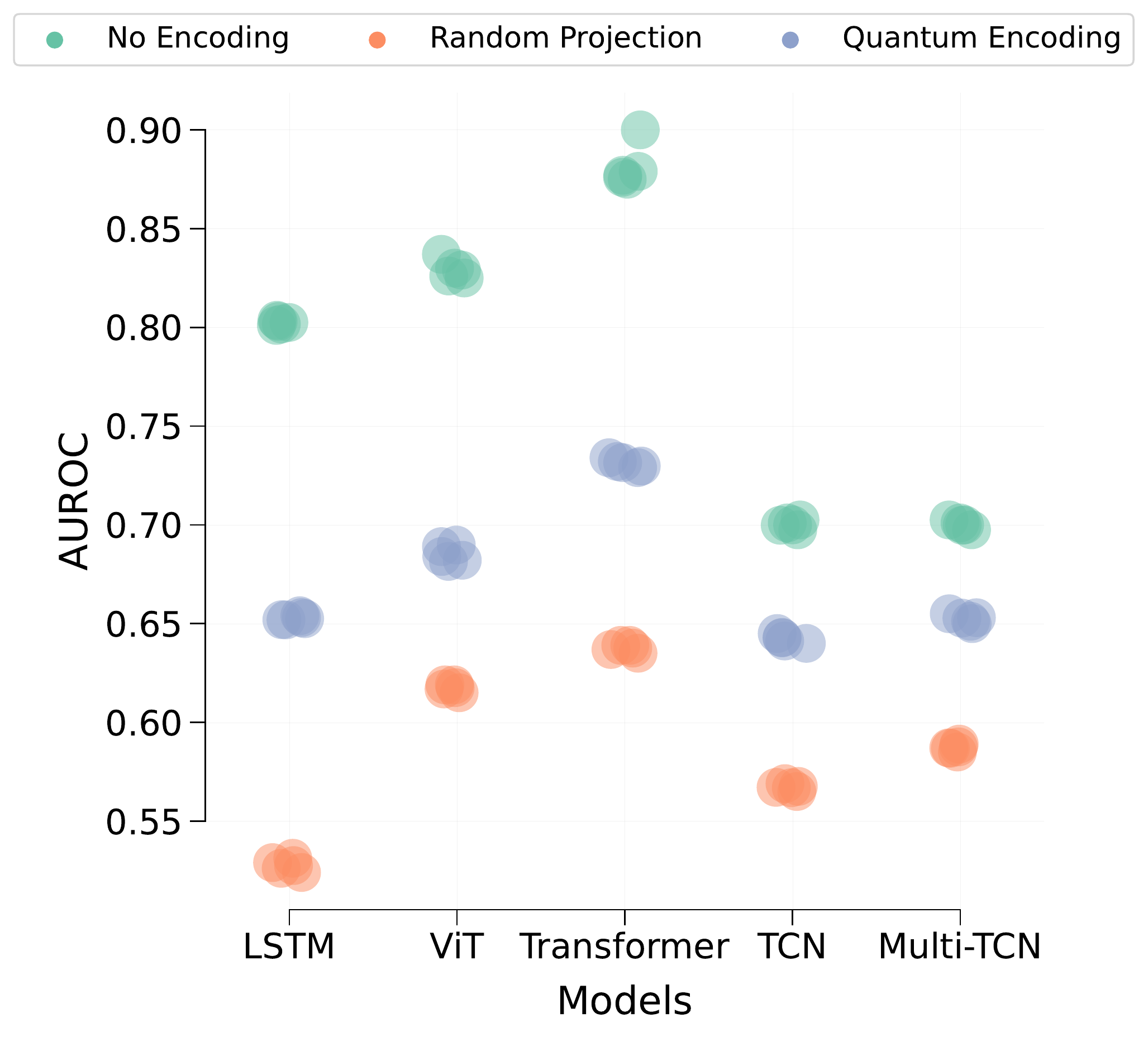}
\includegraphics[scale=0.21]{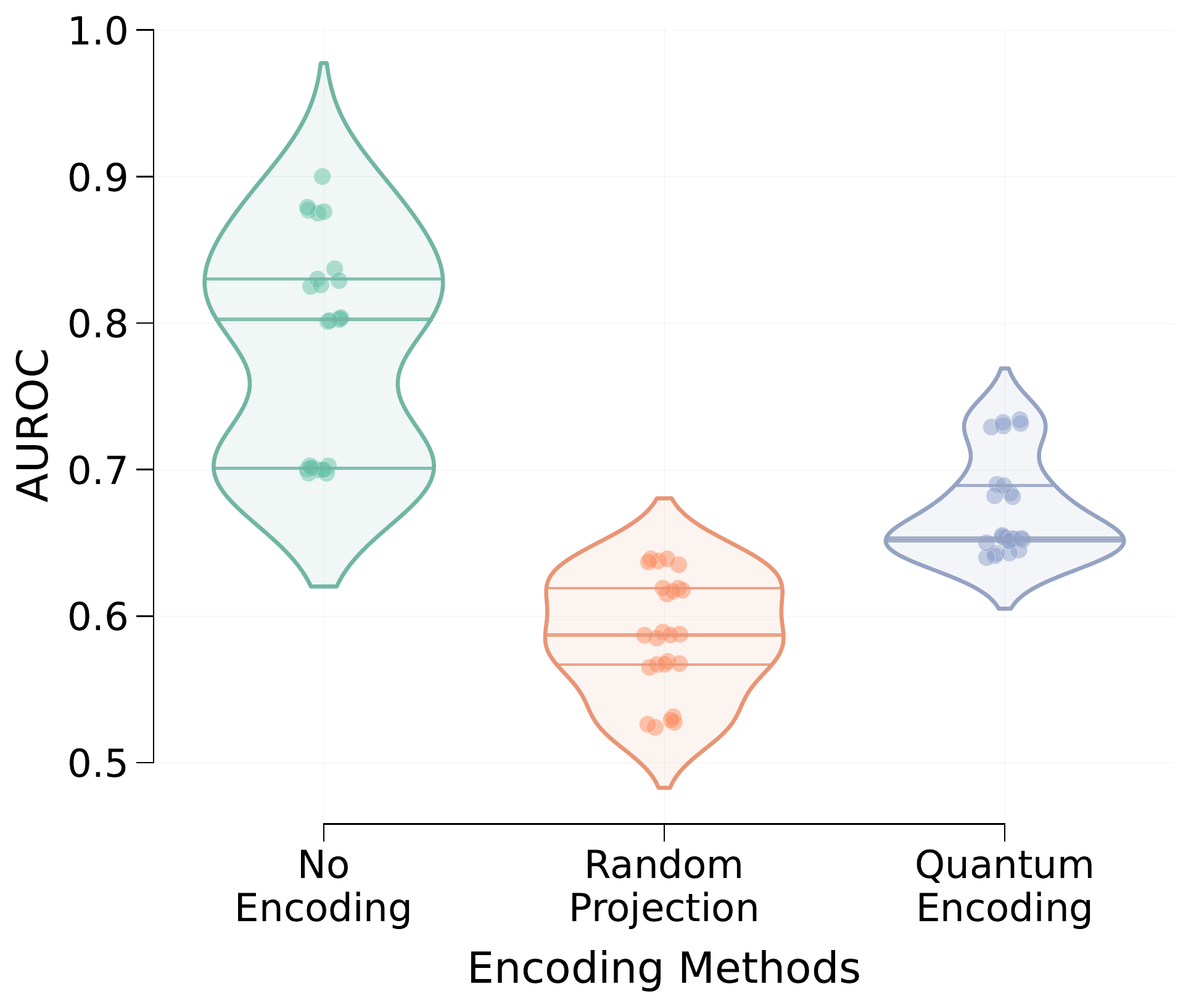}
\caption{Predicting if a patient is Hispanic.} 
\end{subfigure}
\caption[short]{Model-specific and aggregated (across all models) performance for the task of predicting patients' ethnicity from the trained \emph{acute respiratory failure} prediction models.}
\label{fig:ethn}
\end{figure}

The performance for the task of predicting a patient's gender from the trained mortality and ARF prediction models is depicted in Fig.~\ref{fig:res2}. According to the analysis of Fig.~\ref{fig:res2}, we can effectively predict patients' gender from the trained models on original or non-encoded data. The behavior is common across all datasets and all models regardless of their modeling capacity. Similarly, the analysis of Fig.~\ref{fig:ethn} illustrates that we can identify the patients' gender from the ARF models trained on the original time-series data. Although gender and ethnicity are not sensitive information, these results highlight that trained models can indeed reveal the latent non-targeted patient characteristics.

\begin{figure}[H]
\centering
\begin{subfigure}{\textwidth}
\centering
\includegraphics[scale=0.25]{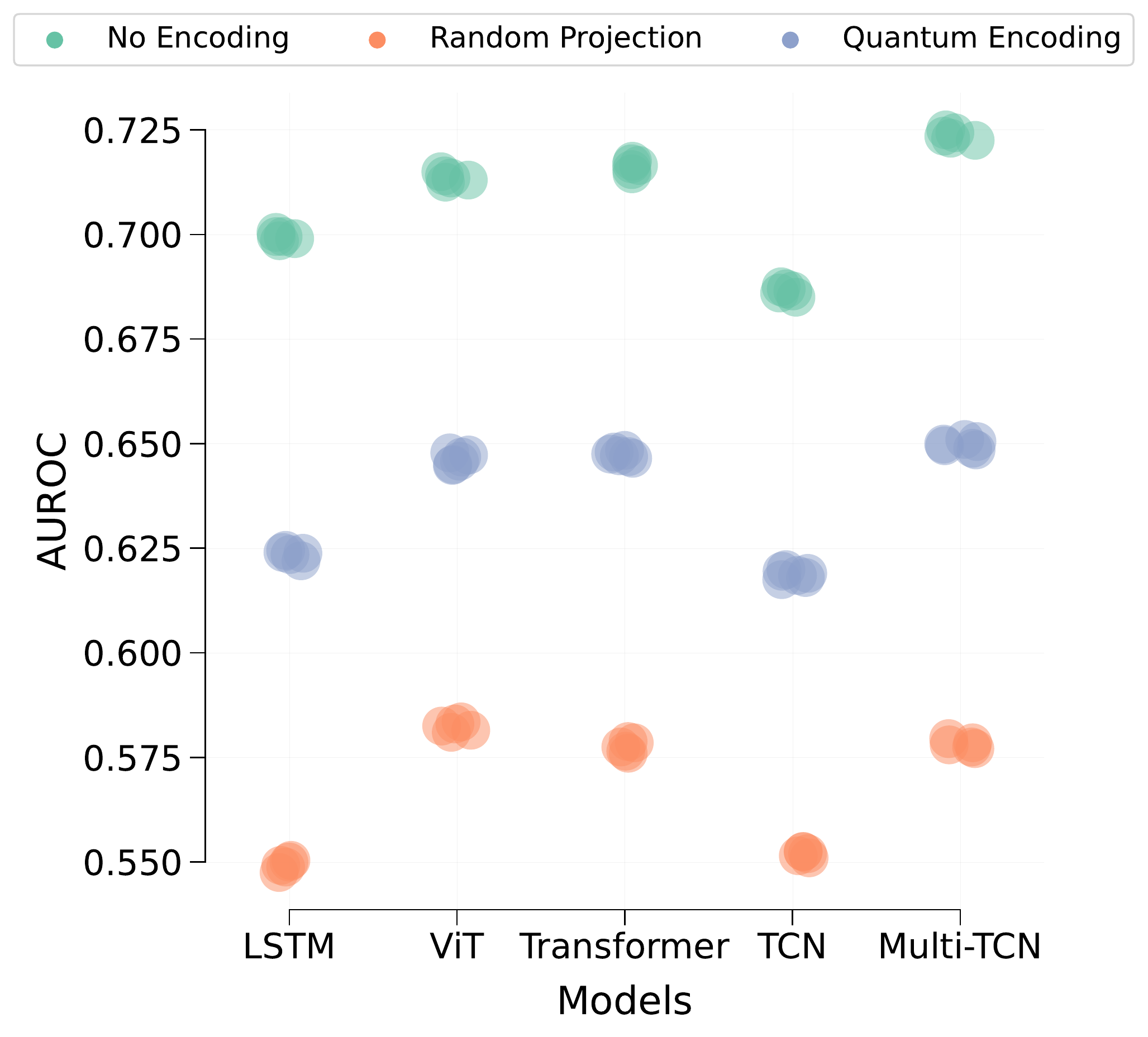}
\includegraphics[scale=0.25]{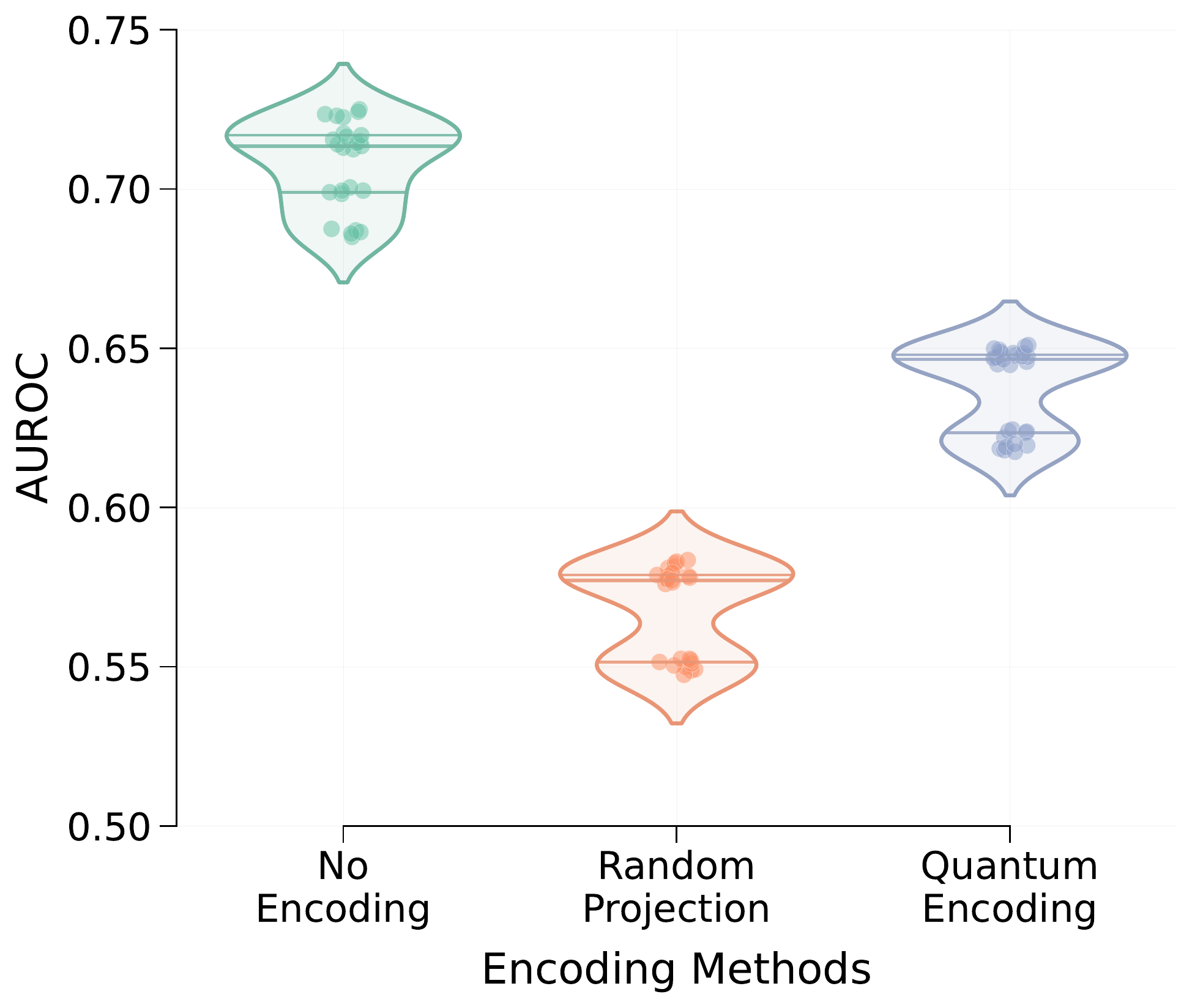}
\caption{Average performance for the prediction of all $25$ patient disorders.}
\end{subfigure}

\begin{subfigure}{\textwidth}
\centering
\includegraphics[trim={0 0.65cm 0 0},clip,scale=0.26]{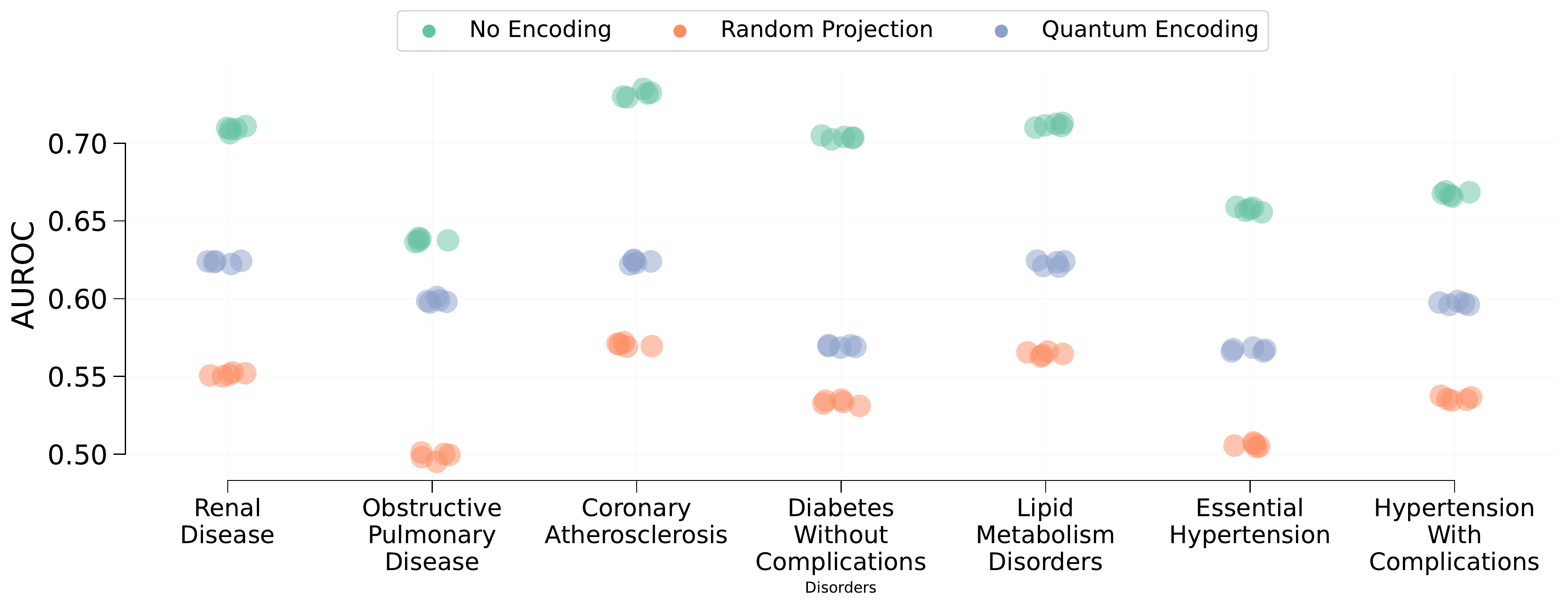}
\caption{Latent prediction of chronic disorders from LSTM-based MIMIC-III models.}
\end{subfigure}

\begin{subfigure}{\textwidth}
\centering
\includegraphics[trim={0 0.65cm 0 0},clip,scale=0.26]{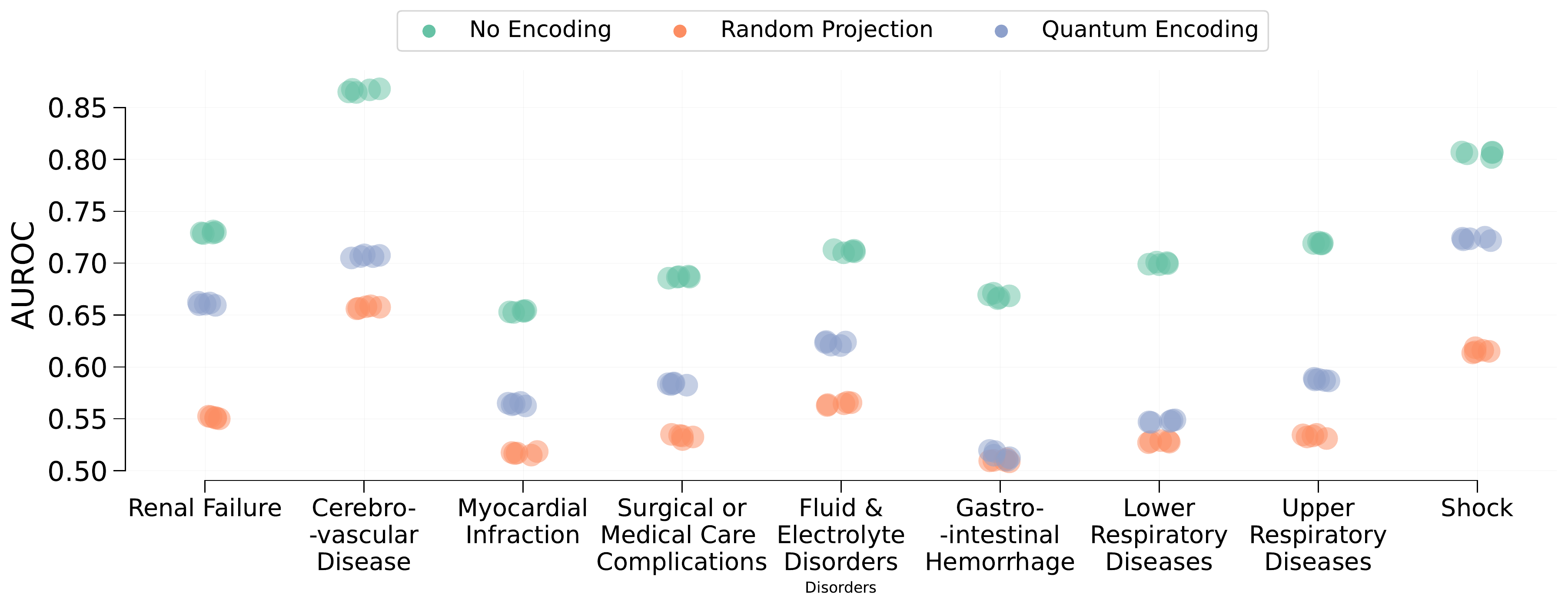}
\caption{Latent prediction of acute disorders from LSTM-based MIMIC-III models.}
\end{subfigure}
\caption[short]{Performance for the task of latent patient disorder prediction using the penultimate embedding generated from MIMIC-III mortality prediction models. The chronic and acute disorders shown in (b) and (c) are subsets of $25$ different conditions considered in this work. A single model predicts the presence/absence of all $25$ disorders.}
\label{fig:disorder}
\end{figure}

\subsubsection*{Latent disorder prediction}
Fig.~\ref{fig:disorder} (a) illustrates the performance of predicting the patient disorders from the trained MIMIC-III models in a latent manner. The analysis of this figure highlights that all models trained on the original data generate representations or embedding that reveal information regarding the patients' disorders. Across all models trained on original data, a macro AUROC of approx. $0.7$ is observed for the latent disorder prediction.  It should be noted that the macro AUROC obtained by different models within this experiment is comparable to the performance achieved by the targeted patient phenotype prediction models (see Fig.~S1 of the supplementary report). This shows that mortality prediction models are susceptible to leaking the patients' private medical information.

 \begin{figure}[t]
\centering
\begin{subfigure}{\textwidth}
\centering
\includegraphics[scale=0.33,trim={0 1.25cm 0 0},clip]{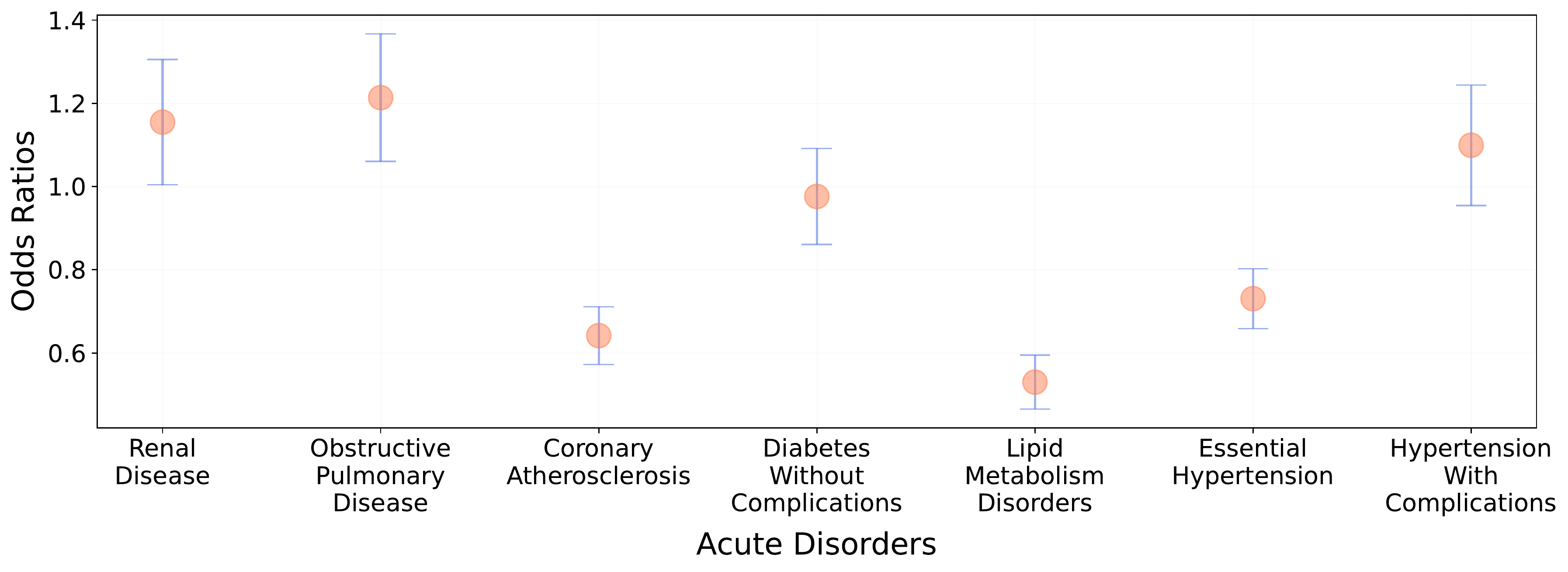}
\caption{Odds ratios between occurrences of chronic disorders and mortality.}
\end{subfigure}

\begin{subfigure}{\textwidth}
\centering
\includegraphics[scale=0.33,trim={0 1.25cm 0 0},clip]{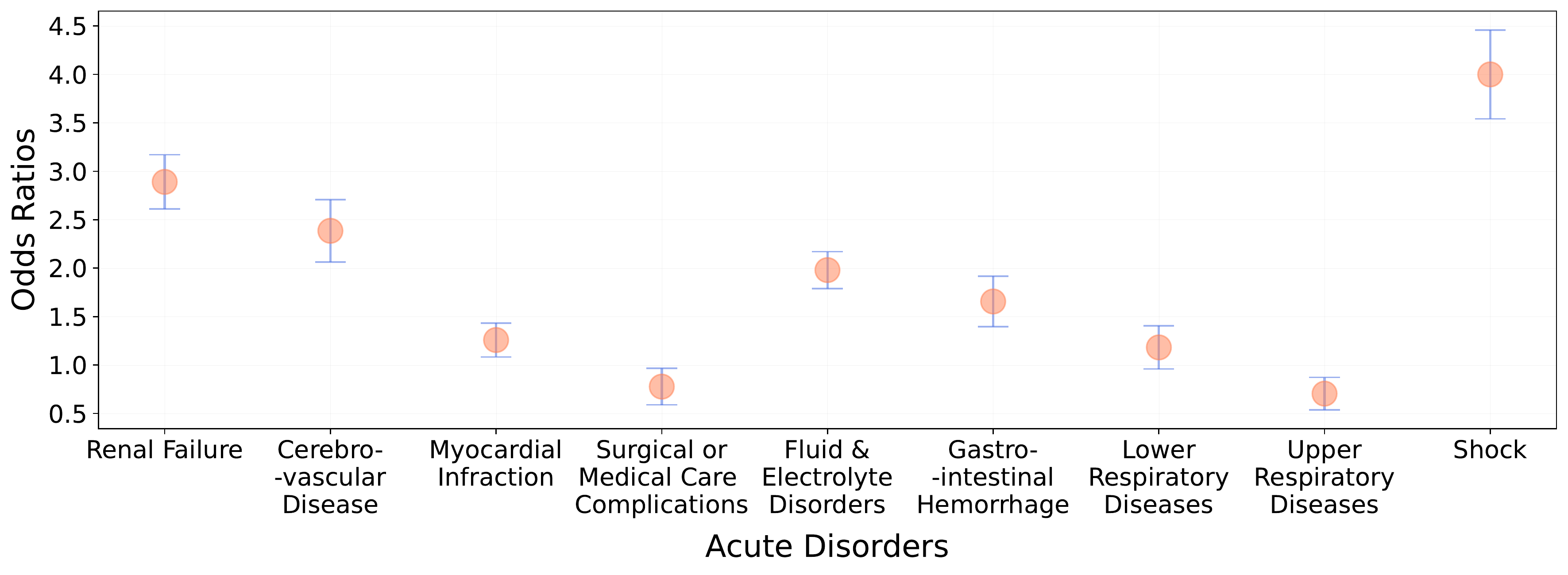}
\caption{Odds ratios between occurrences of acute disorders and mortality.}
\end{subfigure}
\caption[short]{Odds ratios comparing occurrences of \textbf{(a)} chronic and \textbf{(b)} acute disorders against the \emph{outcome} i.e.~mortality in MIMIC-III dataset.}
\label{fig:odds}
\end{figure}

Fig.~\ref{fig:disorder} (b) and (c) depict the performance of predicting the chronic and acute disorders (a subset of 25 disorders) from the trained LSTM mortality prediction models. Similar behavior is observed for all the other models considered in this study (see supplementary document Fig.~S2). The analysis of the figures shows that these models learn the characteristics that help infer or predict non-targeted patient disorders. We can predict both chronic and acute disorders that may or may not be correlated with the mortality prediction. According to the odds ratios \cite{bland2000odds} for acute and chronic disorders (Fig.~\ref{fig:odds} (a) and (b)), most acute conditions exhibit a higher risk of mortality (odds ratio $>>1$), while most chronic conditions are weakly associated with the mortality ($\approx1$).
This shows that some conditions, such as \emph{shock} and \emph{acute renal failure}, are directly associated while others, such as chronic \emph{lipid metabolism disorder} and chronic \emph{renal disease}, are not associated with mortality in the MIMIC-III patients corresponding to the ICU stays. Irrespective of odds ratios or the association between disorders and mortality, we can identify patients ailing from these ailments with an average AUROC of $>0.7$.

\subsection{Encoded data minimizes information leakage}
The analysis of Fig.~\ref{fig:res2}, \ref{fig:ethn}, and \ref{fig:disorder} further highlights that the models trained on the encoded data exhibit lesser latent information leakage than the models trained on the original data. On average, MIMIC-III models trained on data encoded using quantum circuits and random projections (rather than original data) exhibited a relative drop of $20.11\;(\pm 2.45)\%$ and $23.52\;(\pm 3.98)\%$ in performance for the latent gender prediction task. The PhysioNet models also exhibited relative drops of $22.66\;(\pm 5.45)\%$ and $28.21\;(\pm 8.98)\%$ for the data encoded using the quantum circuit and the random projections, respectively. Similar behavior is observed for the eICU models where quantum encoding and random projection-based encoding resulted in a relative drop of $23.1\;(\pm 4.25)\%$ and $31.11\;(\pm 7.6)\%$ in the performance of gender prediction task. The encoding data also resulted in a drop in the performance of the ethnicity prediction tasks. A similar trend is observed for the patient disorder prediction from MIMIC-III models. Quantum encoding and random projections resulted in a relative drop of $12.5\;(\pm 3.79)\%$ and $18.75\;(\pm 5.45)\%$ in the average macro AUROC score. 

\begin{figure}[H]
\centering
\begin{subfigure}{.5\textwidth}
\centering
\includegraphics[scale=0.45]{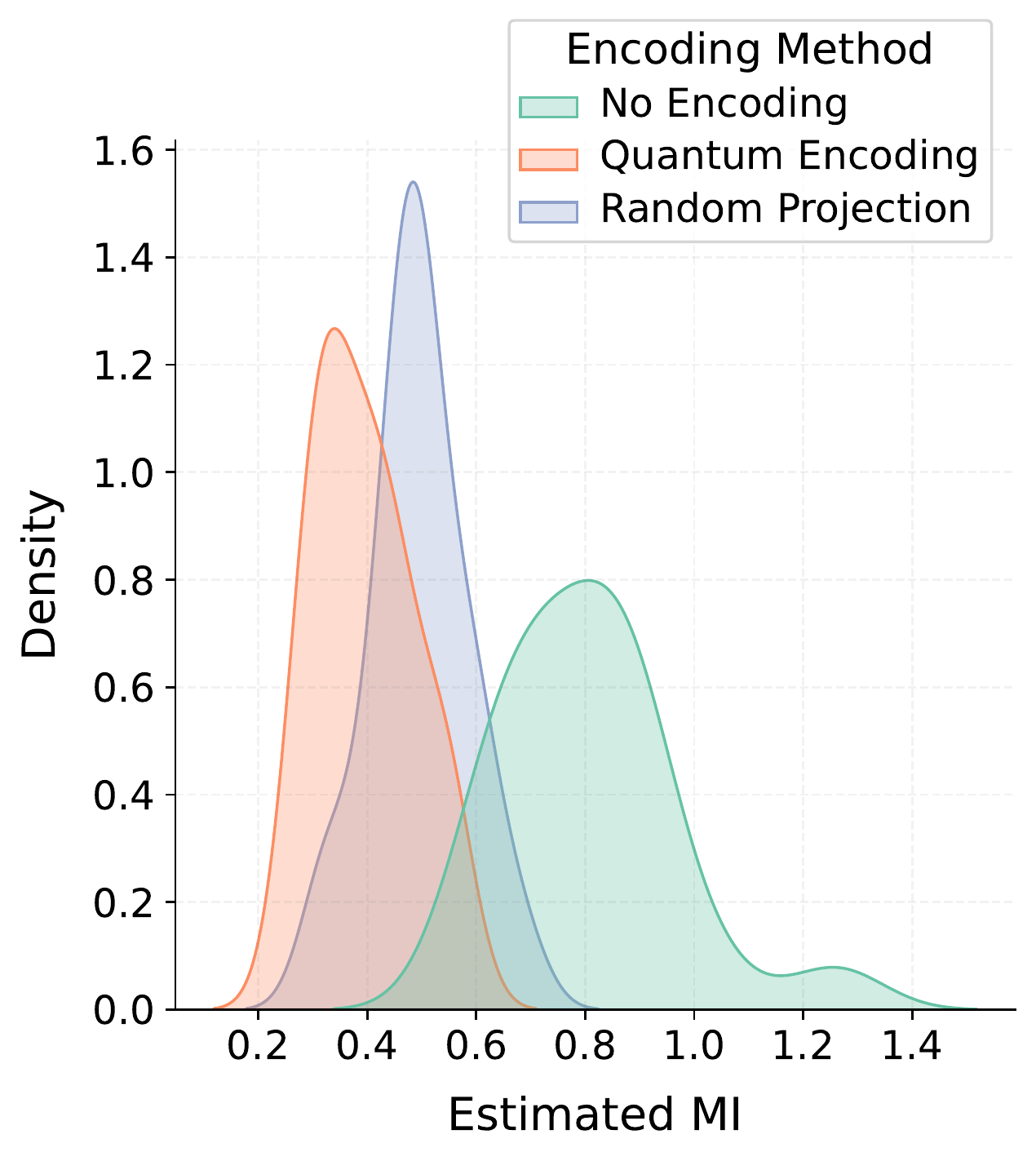}
\caption{\centering On MIMIC-III; MI using\hspace{\textwidth} averaged time-series}
\end{subfigure}%
\begin{subfigure}{.5\textwidth}
\centering
\includegraphics[scale=0.45]{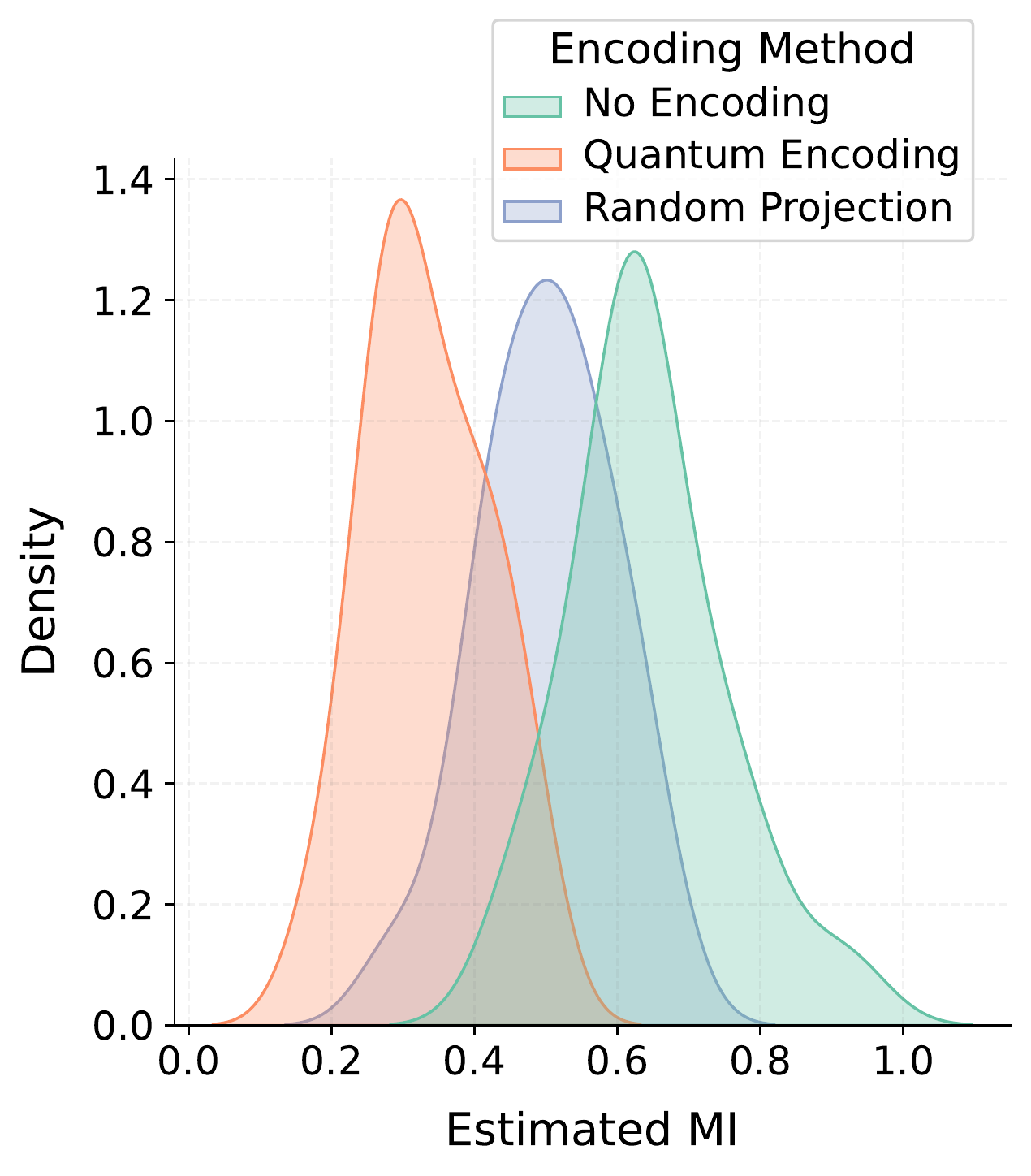}
\caption{\centering On MIMIC-III; MI using\hspace{\textwidth} unraveled time-series}
\end{subfigure}

\begin{subfigure}{.5\textwidth}
\centering
\includegraphics[scale=0.449]{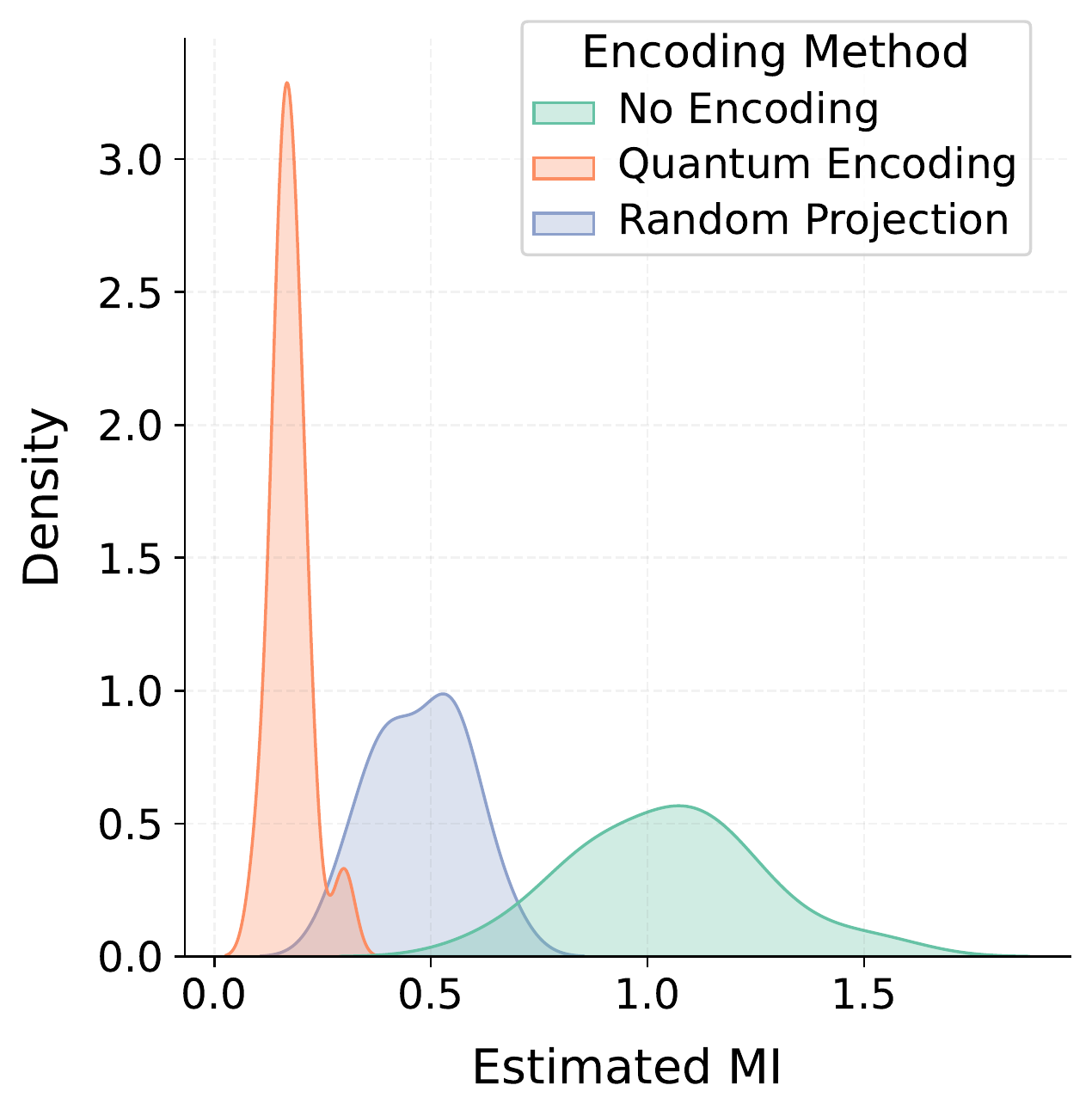}
\caption{\centering On PhysioNet, MI using\hspace{\textwidth} averaged time-series}
\end{subfigure}%
\begin{subfigure}{.5\textwidth}
\centering
\includegraphics[scale=0.45]{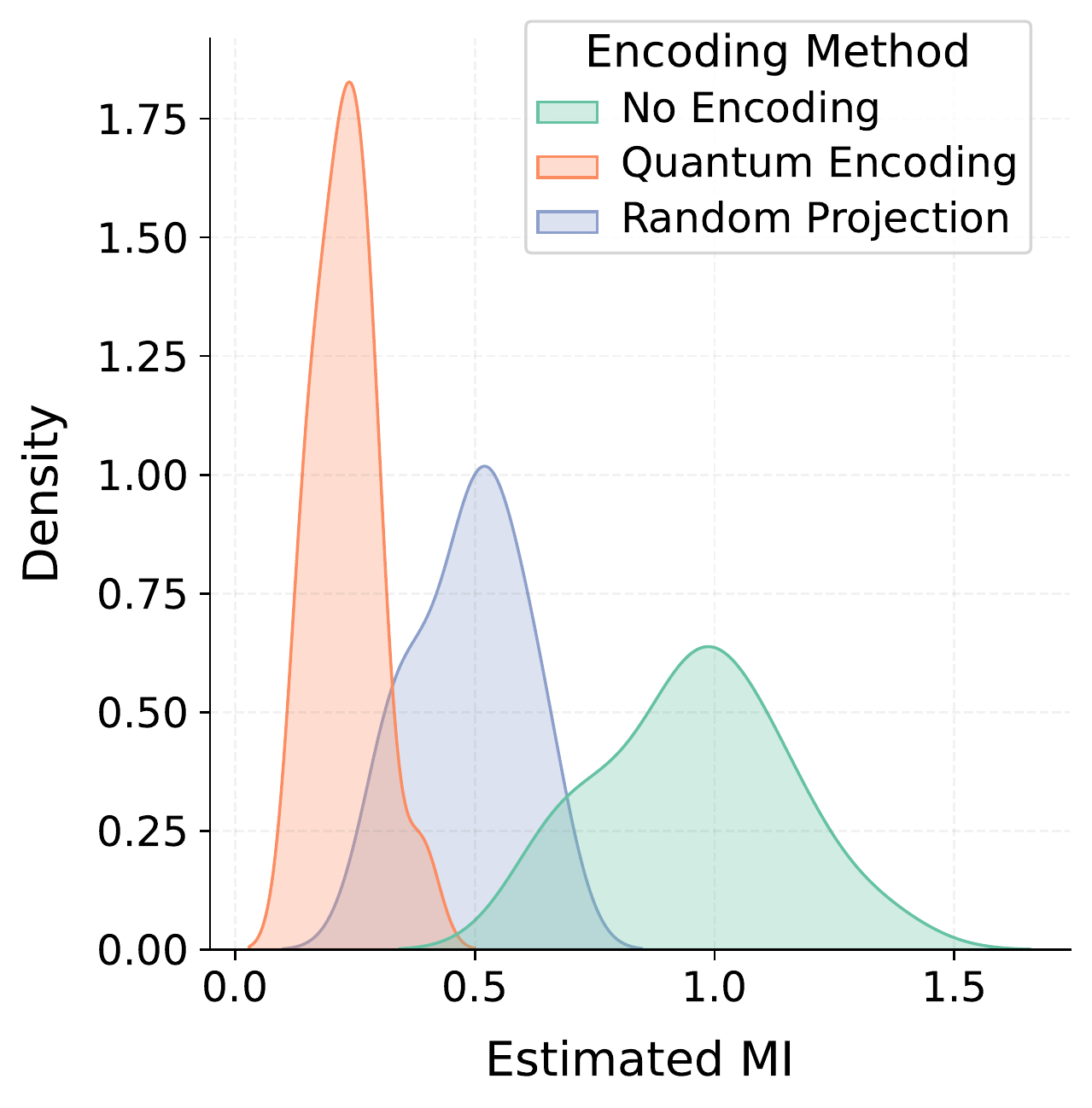}\caption{\centering On PhysioNet; MI using\hspace{\textwidth} unraveled time-series}
\end{subfigure}
\caption[short]{Kernel density estimation plots illustrating the estimated mutual information (MI) between embedding obtained from the trained LSTM models and the input time-series in MIMIC-III (\emph{first row}) and PhysioNet (\emph{second row}) as a function of the encoding methods. To facilitate the MI estimation, the input time series is either vectorized or averaged across time dimensional before computing MI with the embedding.}
\label{fig:MI}
\end{figure}

\vspace{0.1cm}
As discussed in Section \ref{sec:sec1}, models that follow the IB principle exhibit lesser information leakage. The drop in latent information leakage from models trained on the encoded data can be attributed to the lower mutual information (MI) between the model input (i.e.~original or encoded) time series and the penultimate layer embedding generated from the trained models. To uphold this claim, we estimated MI from the LSTMs trained using the original and the encoded data. For the feasibility of MI estimation, we used the average and vectorized form of the input time series to compute MI. Fig.~\ref{fig:MI} illustrates the distribution of estimated MI between the input and the penultimate embedding. It is clear from this figure that the utilization of encoded data minimizes the MI between the model input and the learned representation. As a result, it can be inferred that training models with the encoded data inherently enforce the IB principle in the training process. Hence, the learned embedding only retains the information required to predict mortality while stripping away the non-essential or non-targeted patient information. 

\vspace{0.1cm}
The above analysis shows that random projections-based encoding provides maximum prevention against latent information leakage. However, if we analyze Fig.~\ref{fig:res2} along with Fig.~\ref{fig:res1}, it is also evident that random projection-based encoding results in a larger drop in the performance of the targeted task. On the other hand, random quantum encoding provides more balance between the performance of the targeted task and the prevention of information leakage.

\begin{figure}[H]
\centering
\begin{subfigure}{\textwidth}
\centering
\includegraphics[scale=0.45]{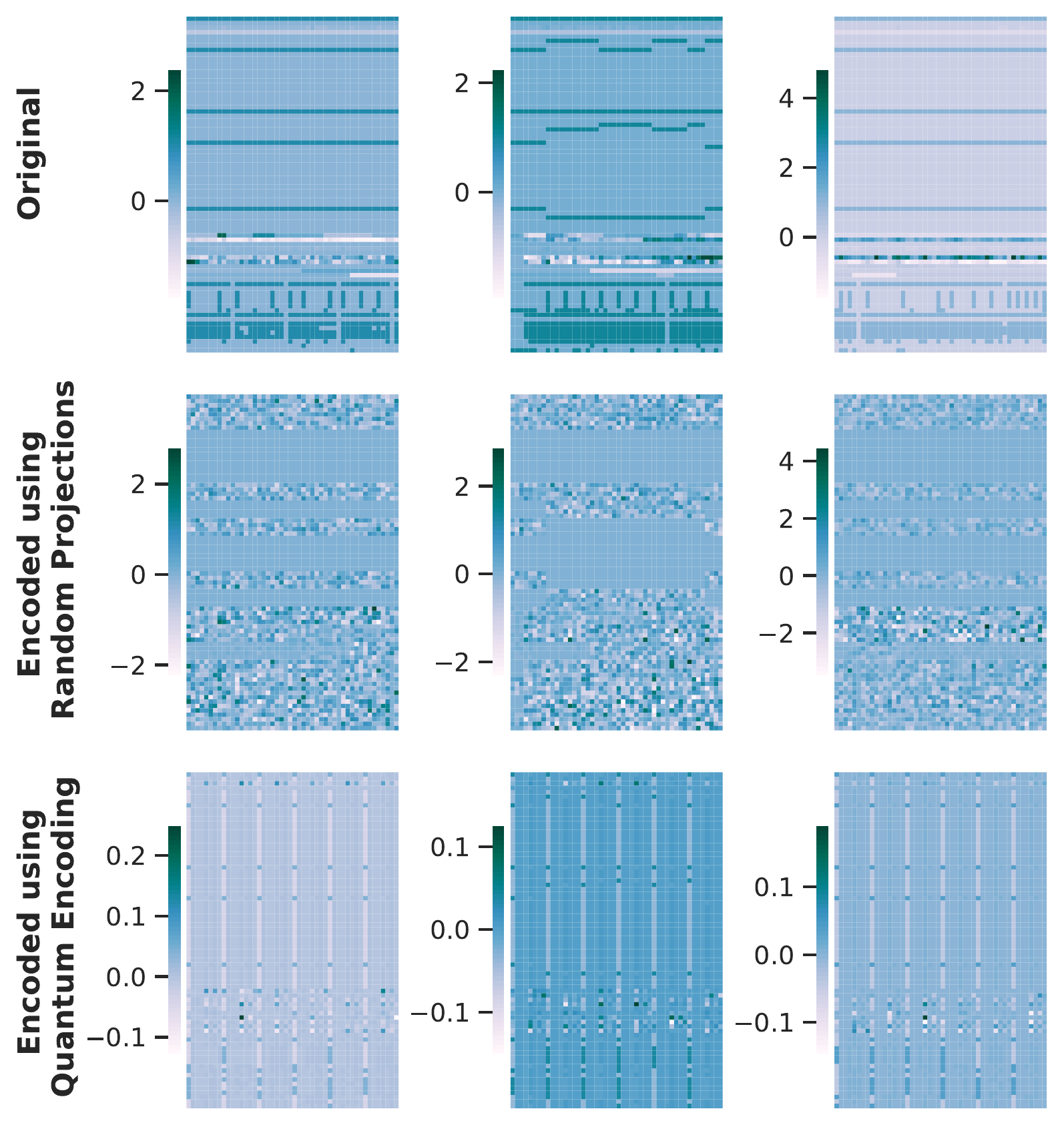}
\caption{Three negative examples from MIMIC-III.}
\end{subfigure}

\begin{subfigure}{\textwidth}
\centering
\includegraphics[scale=0.45]{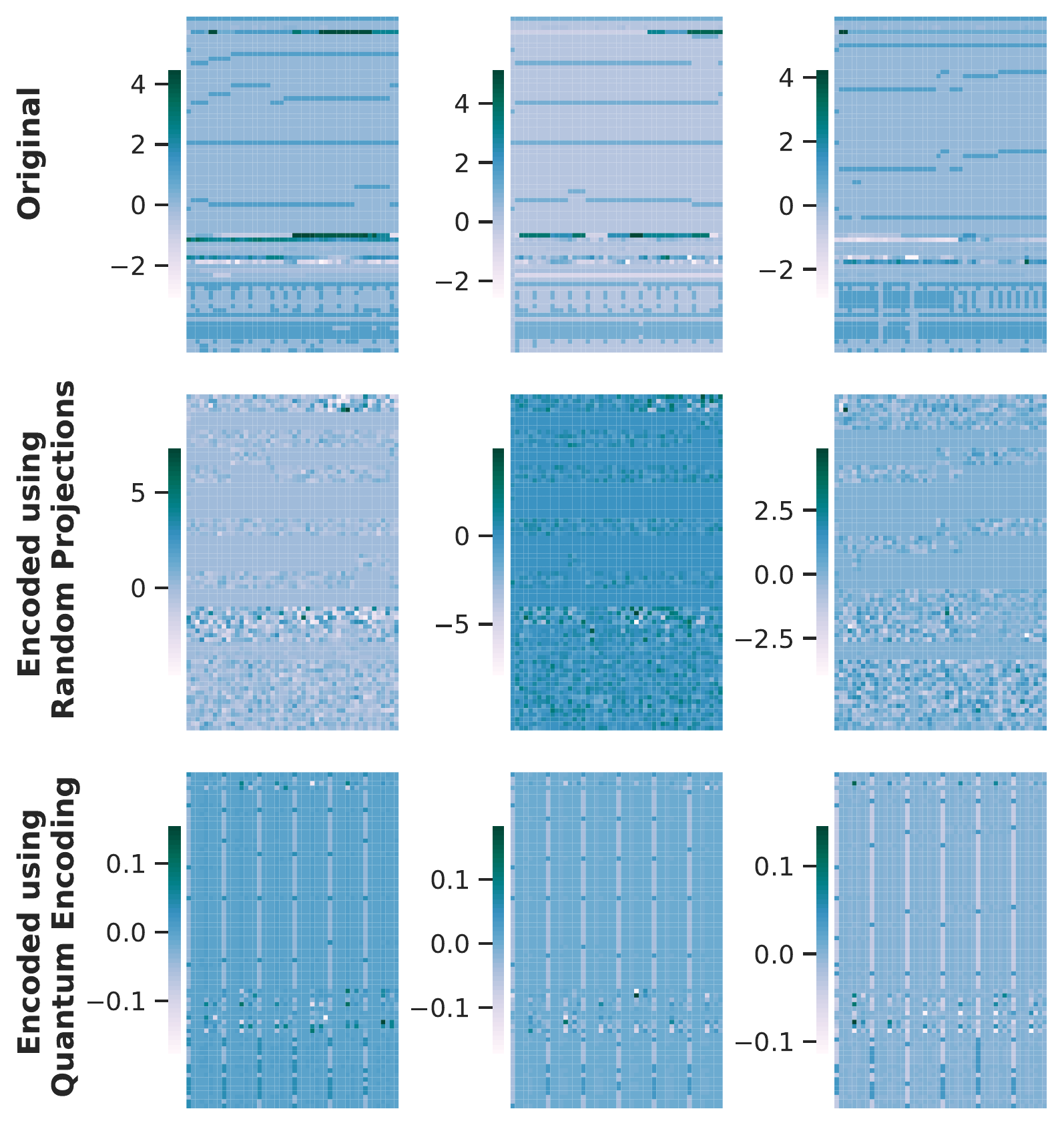}
\caption{Three positive examples from MIMIC-III.}
\end{subfigure}
\caption[short]{Heat maps illustrating the differences in magnitude and trends of the original and encoded time-series examples. Each row represents an input time series and its encoded versions.}
\label{fig:heat}
\end{figure}

\begin{figure}[t]
\centering
\begin{subfigure}{\textwidth}
\centering
\includegraphics[scale=0.3]{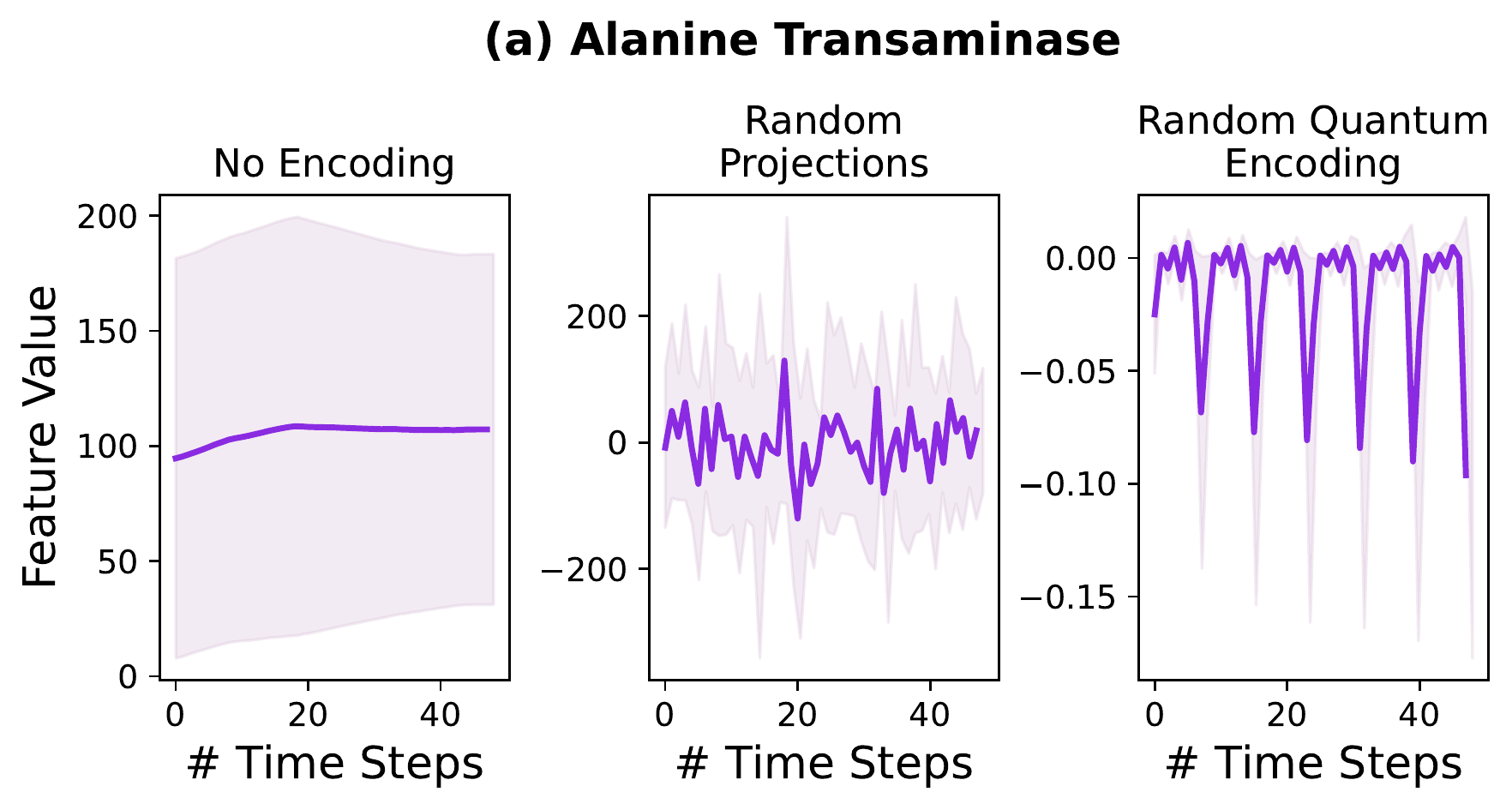}
\end{subfigure}

\begin{subfigure}{\textwidth}
\centering
\includegraphics[scale=0.3]{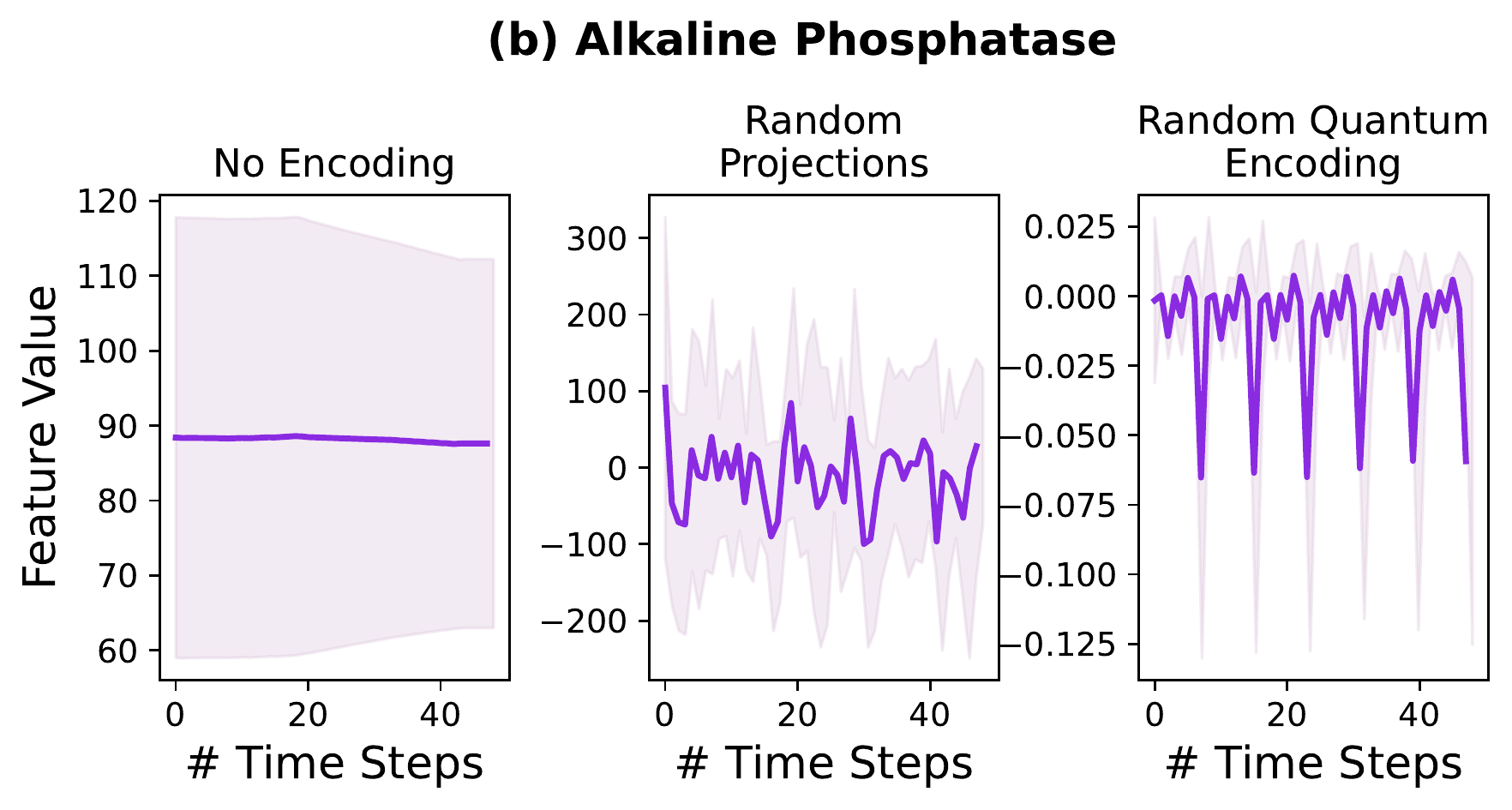}
\end{subfigure}

\begin{subfigure}{\textwidth}
\centering
\includegraphics[scale=0.3]{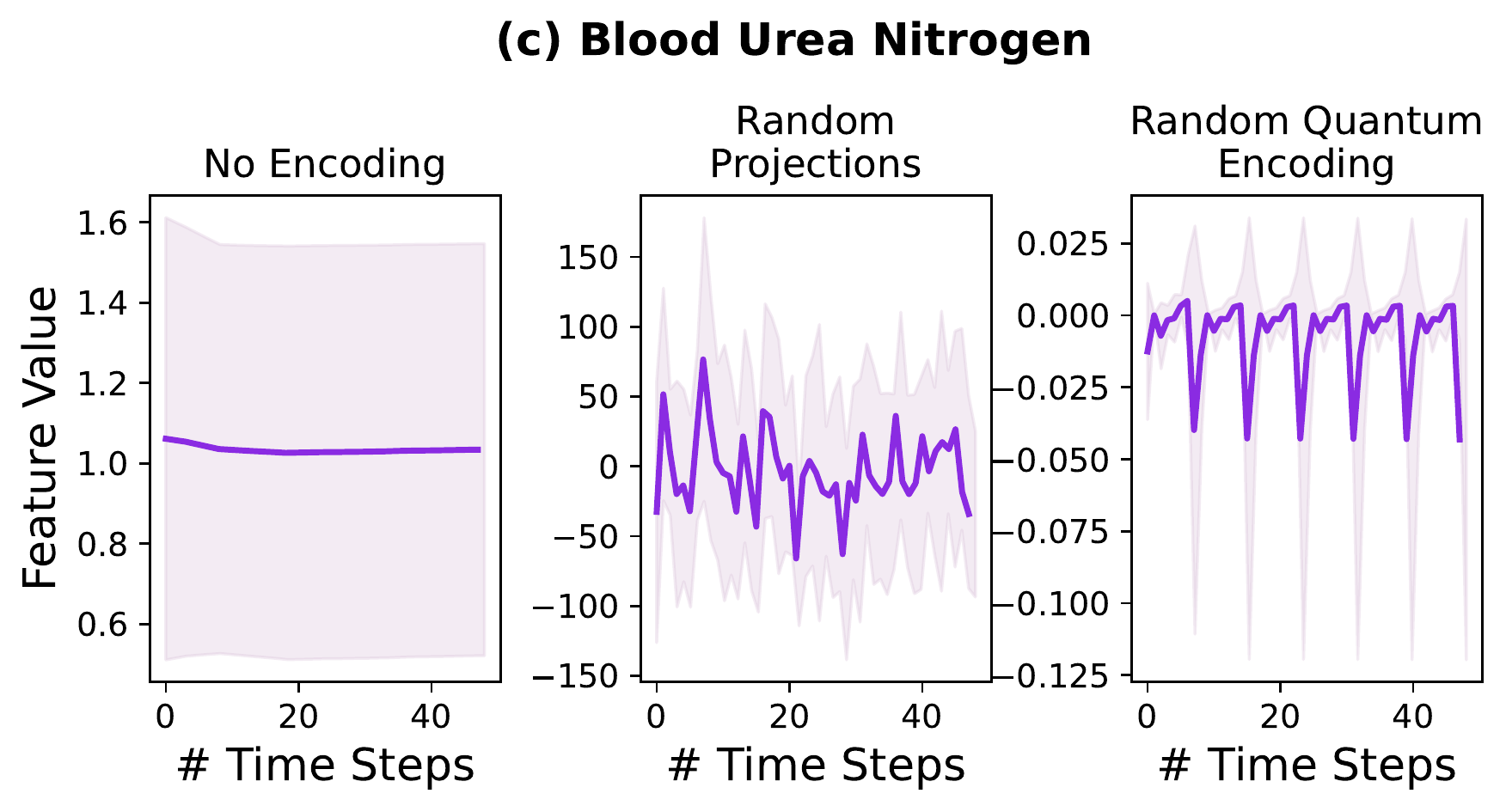}
\end{subfigure}

\begin{subfigure}{\textwidth}
\centering
\includegraphics[scale=0.3]{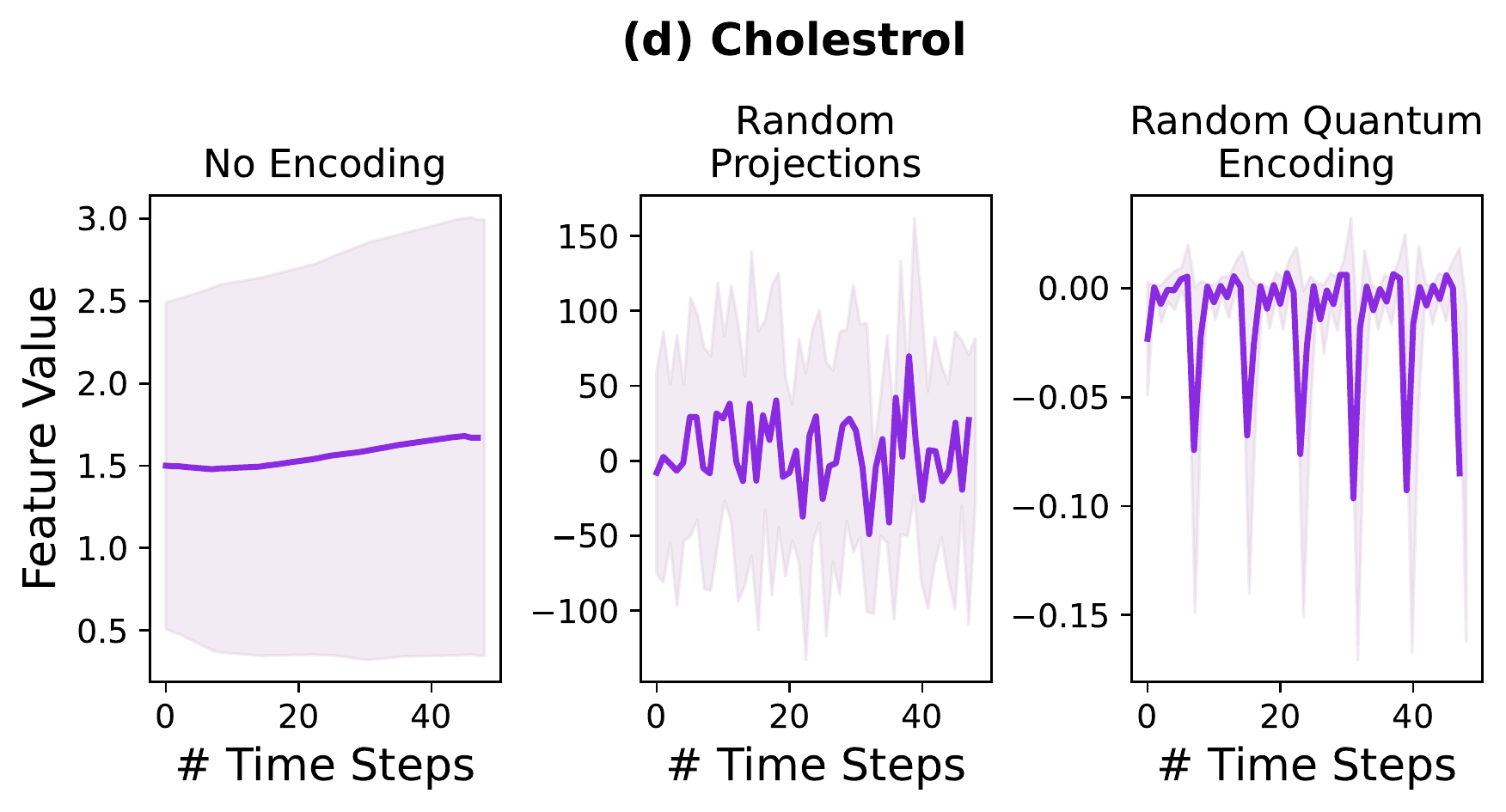}
\end{subfigure}
\caption[short]{Difference in average trends and the average magnitude of the original and encoded signals or features. These signals are obtained by averaging $50$ time series representing patients exhibiting mortality in the PhysioNet dataset.}
\label{fig:res5}
\end{figure}

\subsection{Visual inspection of the encoded data}
The difference between the original and the encoded examples from the MIMIC-III dataset is illustrated in Fig.~\ref{fig:heat}. The analysis of this figure makes it clear that both temporal trends and distribution of features in the original and the encoded time-series examples are noticeably different. To further analyze the impact of the encoding process on the time-series data, $50$ original and encoded examples from the positive (mortality) class of the PhysioNet dataset were randomly selected and averaged to obtain the original and encoded ``summary'' time series. Fig.~\ref{fig:res5} depicts the behavior of four randomly chosen features from these summarised time series. Again, the distribution of magnitude as well as temporal trends of the encoded features is different from the original time-series features. By mere visual inspection, it is near impossible to perceive any information from the encoded data (both quantum encoding and random projections). Similar behavior is observed for the other features. Hence, the encoding process provides an additional layer of privacy over the de-identified data and pushes the community a step closer to achieving data democratization.  

\subsection{Data encoding and explainability}
Encoded data is expected to retain the semantic characteristics of the original data to a large extent such that models trained on original and encoded data exhibit similar behavior. Along with similar performance, the features relevant for predictions in models trained on both the original and the encoded data should largely be the same. While encoded data does retain semantic characteristics, there is a noticeable performance drop due to data encoding (Fig.~\ref{fig:res1}). This shows that the behavior of models trained on encoded data could be different.  

\begin{figure}[t]
\centering
\begin{subfigure}{\textwidth}
\centering
\includegraphics[scale=0.28]{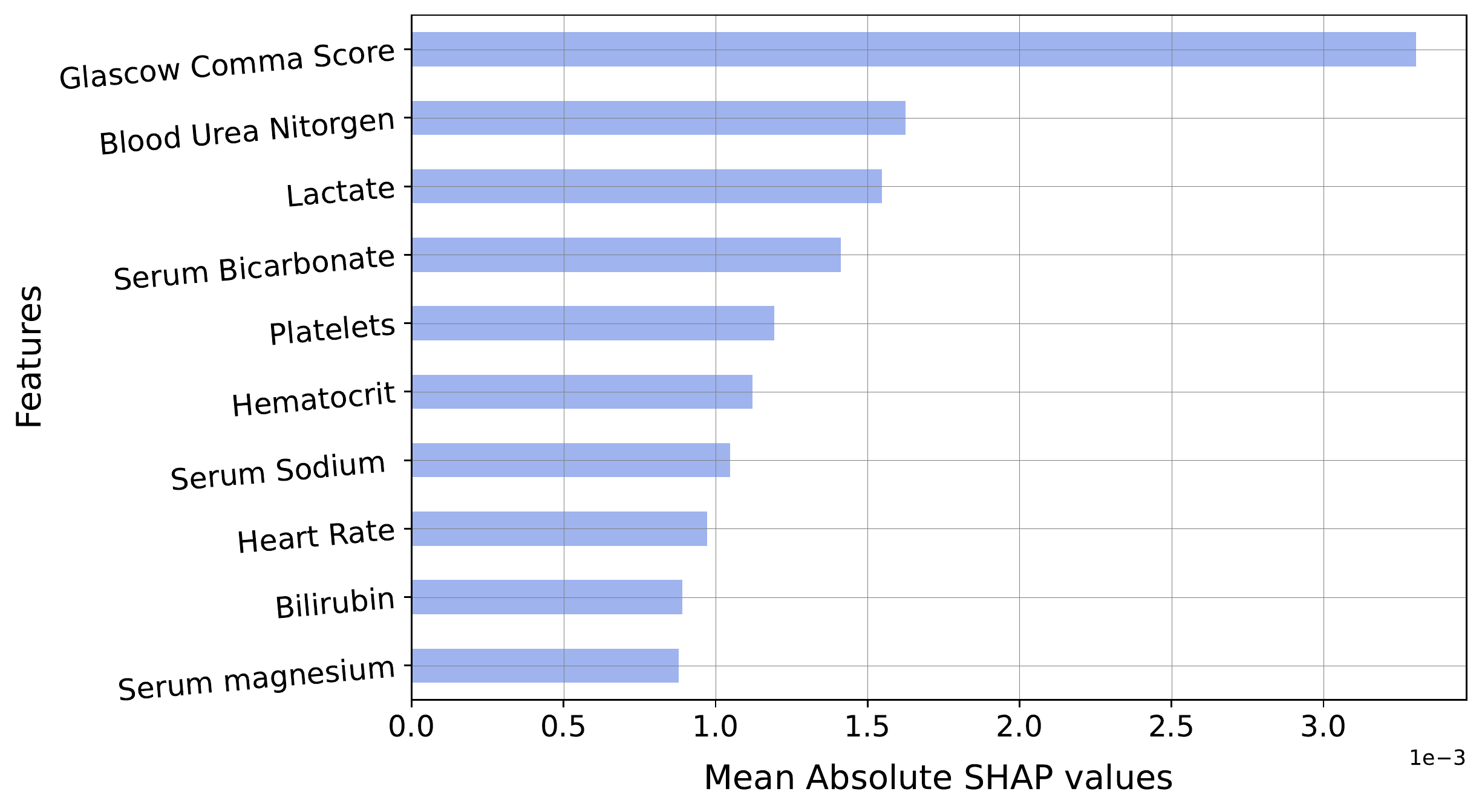}
\caption{SHAP analysis of LSTM trained on PhysioNet dataset.}
\end{subfigure}

\begin{subfigure}{\textwidth}
\centering
\includegraphics[scale=0.28]{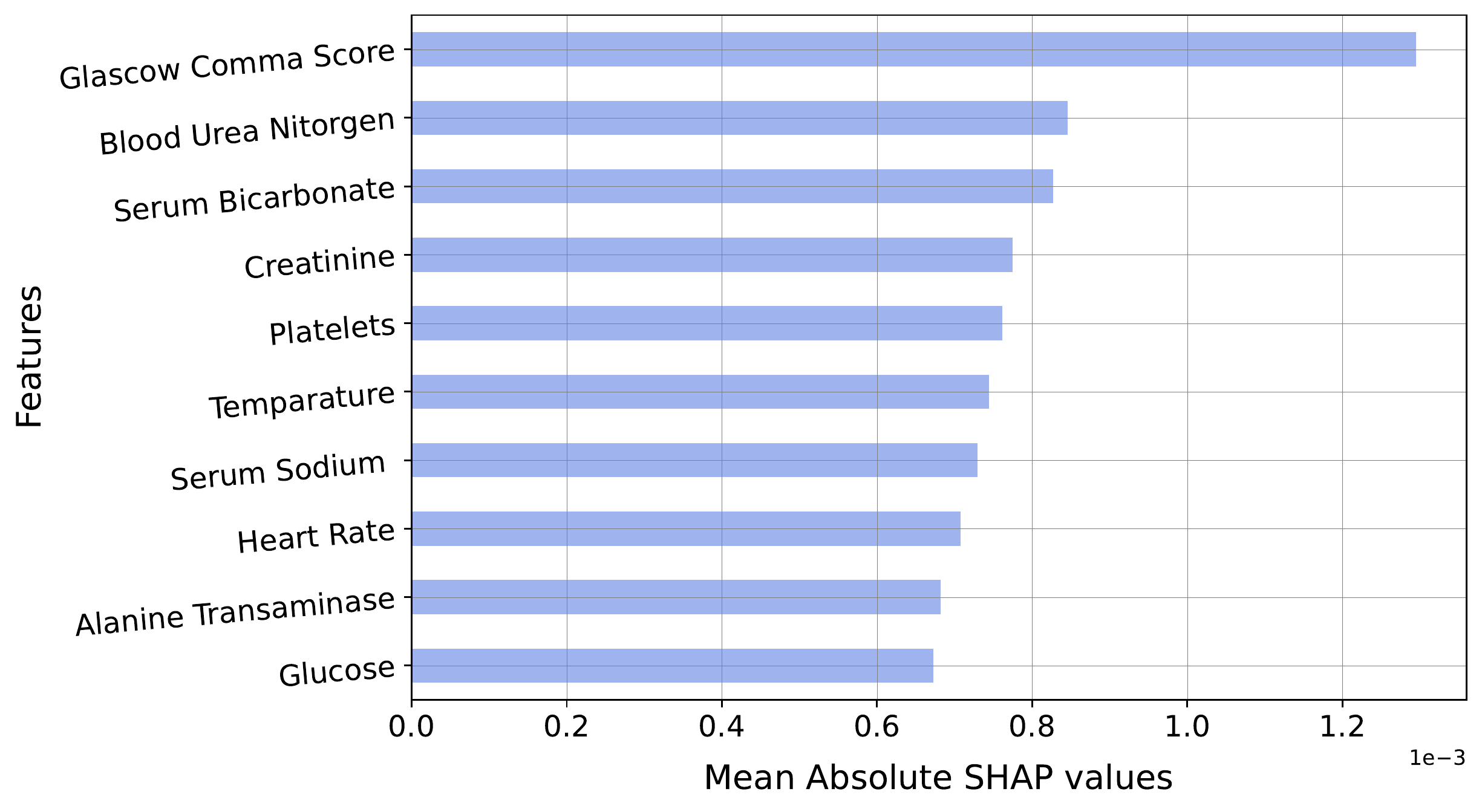}
\caption{SHAP analysis of LSTM trained on quantum encoded PhysioNet dataset.}
\end{subfigure}

\begin{subfigure}{\textwidth}
\centering
\includegraphics[scale=0.28]{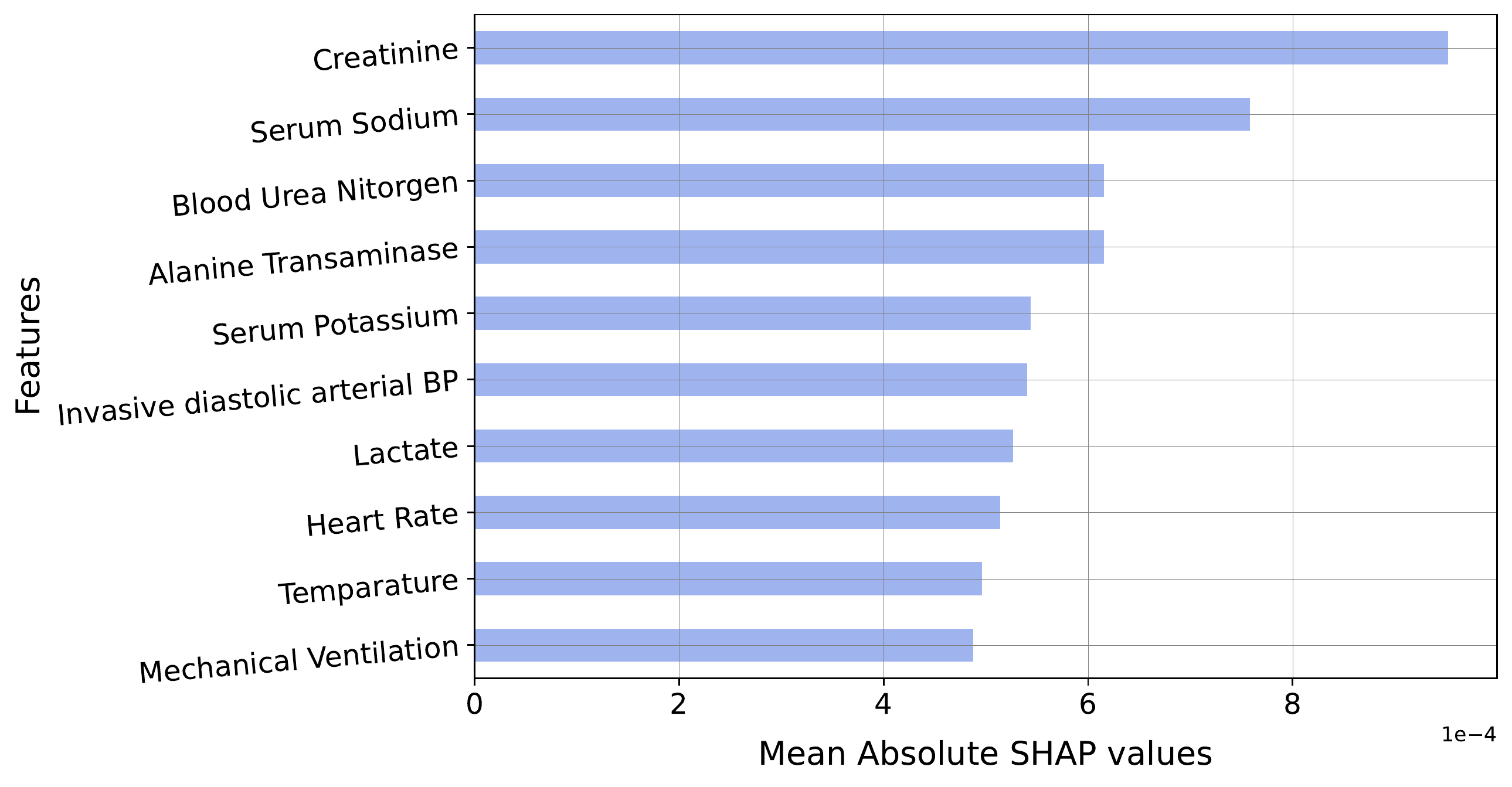}
\caption{SHAP analysis of LSTM trained on PhysioNet data encoded by random projections.}
\end{subfigure}
\caption[short]{A comparison of SHAP-based feature importance in LSTM models trained on (a) original, (b) quantum encoded, and (c) randomly projected versions of the PhysioNet dataset.}
\label{fig:shap}
\end{figure}

Shapely additive explanations (SHAP) \cite{NIPS2017_8a20a862} are employed on the LSTM models, trained using the original and encoded PhysioNet and MIMIC-III datasets, to study the impact of data encoding on the feature relevance. Fig.~\ref{fig:shap} illustrates the top $10$ relevant features identified by SHAP in each PhysioNet model. The analysis of this figure highlights that there is a huge overlap between the sets of relevant features identified for the ``original'' and the ``quantum encoded'' models. Moreover, \emph{Glascow comma score} and \emph{blood urea nitrogen} are regarded as the most relevant features in both models. Although there is some overlap between the relevant features of the original and the ``random projection-based encoded'' models, the overall behavior seems to be very different. Similar behavior is observed for the MIMIC-III models (see Fig.~S3 of the supplementary document). Hence, it can be argued that random quantum encoding has been successful in retaining semantic characteristics such that the resultant models exhibit similar behavior to the original models up to an acceptable level.

\section{Discussion}
\label{sec:dis}
This study proposes to encode the healthcare data to achieve data democratization and prevent information leakage. The \emph{irreversible} and \emph{semantic preserving} encoding process outlined in this paper allows getting an imperceptible and deformed form of healthcare data that can be shared among researchers without violating privacy constraints. Moreover, the inherent regularisation imposed on neural network training due to deformity of the training data induces the \emph{information bottleneck} (IB) principle and results in models that are less susceptible to latent information leakage (Fig.~\ref{fig:MI}).  

This paper explores random projections and random quantum operations to piece-wise encode the 1-d signals in a time series as highlighted in Section \ref{sec:methods} and Fig.~\ref{fig:intro}. Compared to the original time-series signals, the resultant encoded signals exhibit different feature distributions and follow \emph{somewhat} imperceptible trends (Fig.~\ref{fig:res5}). Models trained on the encoded data perform well, highlighting that the semantics are effectively preserved (Fig.~\ref{fig:res1}). Concomitantly, the information leakage from these models is significantly lesser than models trained on the original data (Fig.~\ref{fig:res2}, \ref{fig:ethn} and \ref{fig:disorder}). Thus, as desired, the proposed encoding framework results in encoded data that is visually imperceptible, effective for deep learning and minimizes information leakage from the trained deep models. The encoding transformation is near irreversible if the attacker does not have access to the random Gaussian matrix or random quantum circuit. To learn this transformation computationally or with deep neural networks, both encoded and original data must be available, which is clearly not the case. 

Based on the performance comparison between models trained on data encoded using random projections and random quantum circuits (Fig.~\ref{fig:res1} and \ref{fig:res2}), it is evident that random quantum encoding balances the deformation of data and preservation of the semantic characteristics, which results in better models. Apart from the better performance of quantum encoding, retrieving the original data from its encoded version is theoretically harder as outputs of the quantum circuit or the state of qubits are observed by projecting them on a pre-defined basis state \cite{keyl2002fundamentals}. These measurements become the encoded signals, and estimating the qubit state from this measurement can be ambiguous as multiple qubit states could map to the same measurement. Even if the measurement weren't an issue, one would have to estimate the structure of quantum circuits (number of layers, number of gates, and nature of gates) as well as the parameters of rotation gates to reverse the encoding process possibly. In contrast, we only need to estimate the transformation matrix ($4 \times 4$) to reverse the random projections. It will be sufficient to have access to even one pair of original and encoded data to estimate this transformation matrix accurately.      

\vspace{0.1cm}
The encoding of data is also able to facilitate collaboration among multiple research entities without infringing upon the privacy of the patients. All data collection sites can potentially share their data among themselves so that every site can access the ``global'' data. As discussed in Section \ref{sec:sec1}, the models trained on this global data are expected to be more generic and better at handling the population-specific distribution shifts. However, the random nature of encoding at each site will impede this cross-site collaboration. This problem can be solved by agreeing beforehand on the nature of data transformation, such as quantum circuit structure and rotation gate parameters. Thus, encoded data from each site will be in the same transformation space, allowing deep learning models to be trained effectively. Similar to cross-site collaboration, federated learning also allows a central server to collaborate with multiple sites for training a global model without data sharing \cite{rieke2020future}. However, the structure of models is entirely decided by the server, and sites do not have any independence. Each site is expected to perform similar operations using its local data. On the other hand, data encoding allows the researchers at each site to access the global data and work independently on any deep learning algorithm.   

\vspace{0.1cm}
As an alternative to data encoding, generative models such as generative adversarial networks have been used to generate the data points that do not represent any real patients and theoretically can be shared publicly \cite{chen2021synthetic,jordon2020synthetic}. However, generative models capture the input distribution of the data points, and it is always possible to sample data points that are extremely similar to the input points or real patients. Similar to the subjectivity around the de-identification process (as discussed in Section 1), a sampled example that is similar to real patient data may or may not be considered as a fabricated data point. Moreover, generative modeling requires extensive computational resources and a ``large amount'' of data to fabricate the data points effectively. On the other hand, the proposed encoding approach is an information-processing framework and does not require any training. 

\vspace{0.1cm}
In hindsight, the proposed encoding framework suffers from two major drawbacks:

\begin{itemize}
\itemsep 1em
\item The proposed framework was designed to encode the data for deep learning models that are known for their higher modeling capabilities. The deformations in the encoded data make it difficult for traditional machine-learning models with limited capabilities to process the encoded data effectively. Similarly, the summary statistics of the encoded and the original data are poles apart (by design). As a result, traditional statistical or epidemiological analyses are not feasible. Hence, the use cases of the encoded data are only limited to deep learning models.  
        
\item Both random projections and random quantum encoding don't provide any explicit mechanism to control the degree of deformation in the encoded data or to maintain a balance between the imperceptibility of the encoded data and retaining the semantic information. The performance drop in models trained on the encoded data can be attributed to a lack of this balancing mechanism that could have induced only the minimum required deformation to make the data imperceptible and prevent information leakage.
\end{itemize}

In the future, we will work towards coming up with non-linear or sub-linear data transformations that could either automatically balance the deformation and semantic retention trade-off or provides a hyper-parameter to control the degree of deformation in the encoded data. Using such data transformations in the proposed encoding framework will improve the performance of the target tasks while enabling data democratization and preventing information leakage.

\section{Methods}
\label{sec:methods}

\subsection{Proposed encoding framework}
A uniformly sampled multivariate time series is a collection of multiple 1-d signals representing features measured over time. Suppose $\mathbf{X} \in \mathbb{R}^{F \times T}$ is a time-series consisting of $F$ 1-d signals of length $T$, and $\mathbf{x} \in \mathbb{R}^T$ or $\mathbf{x}=\left[\mathbf{x}_1,\mathbf{x}_2, \mathbf{x}_3, \ldots \mathbf{x}_T\right]$ is one of the $F$ signals. The proposed framework transforms the time-series $\mathbf{X}$ by performing piece-wise encoding of every 1-d signal in $\mathbf{X}$. The framework divides the signal $\mathbf{x}$ into segments or chunks of length $n$ as 
$\hat{\mathbf{x}}=\left[\mathbf{x}
_{1:n},\mathbf{x}_{n+1:2n}
,\ldots \mathbf{x}_{(T-n+1):T}\right]$ and applies transformation operation $f()$ on every segment:

\begin{equation} \label{eq1}
\mathbf{e}_j=f(\hat{\mathbf{x}}_j) \;\;\;\forall\;\hat{\mathbf{x}}_j \in \hat{\mathbf{x}},
\end{equation}

\noindent where $\mathbf{e}_j \in \mathbb{R}^n$ is encoded version of $j$th segment of $\mathbf{x}$. Note that the dimensions of transformed/encoded and input segments are the same, and a segment length of $n=4$ has been used across all experiments. Each encoded segment of length $n$ is temporally concatenated to obtain the encoded version, $\mathbf{e} \in \mathbb{R}^T$, of the signal $\mathbf{x}$ as:
$\mathbf{e}=\left[\mathbf{e}_1, \mathbf{e}_2, \mathbf{e}_3 \ldots \mathbf{e}_{(T/n)} \right]$. Similarly, transformation or encoding operation is applied on all $F$ 1-d signals to transform $\mathbf{X}$ into the encoded time-series $\mathbf{E} \in \mathbb{R}^{F \times T}$. In this paper, we have used random projection and random quantum encoding as data transformation operation $f()$ in the proposed framework. Both these mechanisms are discussed below.

\subsubsection{Random projection}
Random projection is a method of projecting the input data into a random subspace using a random projection matrix whose columns are of unit length \cite{bingham2001random,vempala2005random}. It is mainly used for dimensionality reduction, and it approximately preserves the similarity among data points in the projected subspace as outlined by Johnson–Lindenstrauss lemma \cite{larsen2017optimality}. In this work, we are not interested in dimensionality reduction and are mainly concerned with projecting the input into a random subspace to make the data imperceptible. To attain this goal, we use a projection matrix $\mathbf{R} \in \mathbb{R}^{n \times n}$ whose entries are randomly sampled from Gaussian distribution $\mathcal{N}(0,1/n)$. This projection matrix can be used to encode the $j$th segment $\hat{\mathbf{x}}_j \in \mathbb{R}^{n \times 1}$ of signal $\mathbf{x}$ as:
\begin{equation}
    \mathbf{e}_j= \mathbf{R}\hat{\mathbf{x}}_j,
\end{equation}
where $\mathbf{e}_j \in \mathbb{R}^{n \times 1}$ is the encoded version of the input segment. As discussed above, we have used a segment length of $n=4$, so the projection matrix of $4 \times 4$ is used for data encoding. 

\subsubsection{Random quantum encoding}
The term random quantum encoding refers to a process of data transformation through the use of a quantum circuit containing multiple gates with random parameters \cite{yang2021decentralizing}. The Quantum circuit used in this study is shown in Fig.~\ref{fig:intro}(c). This circuit is composed of the following components \cite{kaye2006introduction}:

\begin{itemize}
  \setlength\itemsep{1em}
    \item \emph{Qubits or wires:} The circuit consists of four wires to represent four quantum bits or qubits. A qubit is a quantum system having a resting state $\ket{0}$ and an excited state $\ket{1}$. These states are mutually orthogonal and any qubit state $\ket{\psi}$ can be represented as a superposition of $\ket{0}$ and $\ket{1}$ as: $\ket{\psi}=a\ket{0}+b\ket{1}$, where $a$ and $b$ are complex numbers that must satisfy $\lvert a \rvert^2 + \lvert b \rvert^2=1$. $\lvert a\rvert^2$ and $\lvert b\rvert^2$ represent the probability of $\ket{\psi}$ being in $\ket{0}$ and $\ket{1}$, respectively. Initially, all four qubits are in a resting state. The number of wires or qubits is dictated by the length of the input segmented signal i.e.~$n=4$.
    
    \item \emph{Rotation gates (RX)}:  These gates rotate the qubit around x-axis by $\phi_k$ (radians) on its Bloch sphere projection, where $k$ is the index of RX gate in the circuit. This rotation operator with $\phi_k$ randomly chosen parameters can be defined as:    
\begin{equation}
RX(\phi_k)=\begin{bmatrix}
\textrm{cos}\frac{\phi_k}{2} &-\iota \textrm{sin} \frac{\phi_k}{2}\\
-\iota\textrm{sin}\frac{\phi_k}{2} &\textrm{cos}\frac{\phi_k}{2}
\end{bmatrix}
\end{equation}
The resultant qubit state $\ket{\psi'}$ after applying $k$th RX gate to qubit $\ket{\psi}$ is given as:    
\begin{equation}
\ket{\psi'}=\begin{bmatrix}
\textrm{cos}\frac{\phi_k}{2} &-\iota \textrm{sin} \frac{\phi_k}{2}\\
-\iota\textrm{sin}\frac{\phi_k}{2} &\textrm{cos}\frac{\phi_k}{2}
\end{bmatrix} \begin{bmatrix}
a \\
b
\end{bmatrix}
\end{equation}

\item \emph{Controlled-NOT (CNOT) gates}: A CNOT gate is used to entangle the two qubits and has no parameters. First qubit is considered as control and the second qubit is flipped if the control is $\ket{1}$. As we can see, CNOT deals with 2-qubit quantum system whose basis states are $\{\ket{00},\ket{01}, \ket{10},\ket{11}\}$. An input to CNOT gate is a linear superimposition of these basis states: $\ket{\psi}=a\ket{00}+b\ket{01}+c\ket{10}+d\ket{11}$, where $a,b,c$ and $d$ are the complex coefficients. Hence, CNOT operation can be defined as:
\begin{equation}
    \textsc{cnot}(\ket{\psi})=a\ket{00}+b\ket{01}+d\ket{10}+c\ket{11}.
\end{equation}
\end{itemize}

\noindent\textbf{Encoding using quantum circuit:} The whole quantum encoding process can be divided into three steps:

\begin{itemize}
      \setlength\itemsep{1em}
\item \emph{Encoding input segment on wires:} The first step is to infuse or project the input segment $\hat{\mathbf{x}}_j$ on wires of the circuit. Each element ($\hat{\mathbf{x}}_{j_n}$) of the input segment $\hat{\mathbf{x}}_j$ corresponds to $n$th wire or qubit. To encode the information from $\hat{\mathbf{x}}_{j_n}$ to $n$th qubit, we rotate this qubit by $\hat{\mathbf{x}}_{j_n}$ radians around $y$-axis on its Bloch sphere projection. This rotation operator is described as:
\begin{equation}
RY(\phi_n)=\begin{bmatrix}
\textrm{cos}\frac{\phi_n}{2} &-\textrm{sin} \frac{\phi_k}{2}\\
\textrm{sin}\frac{\phi_n}{2} &\textrm{cos}\frac{\phi_k}{2}
\end{bmatrix},
\end{equation} 
where $\phi_n$ is $\pi \hat{\mathbf{x}}^j_{n}$. The process of applying this operator is similar to RX gates (Equation 4).   

\item  \emph{Processing qubits by quantum circuit:} After preparing the qubits as encoded versions of the input segment, these qubits are processed by the quantum circuit (Fig.~\ref{fig:intro}C) described above. 

\item  \emph{Measuring the outputs:} This operation is used to register the state of a qubit after applying all the quantum operations. In this work, we use the expectation of the Pauli-Z operator ($\mathbf{Z}$) to measure the output state of a qubit $\ket{\psi}$. We know that $\mathbf{Z}$ can be defined as \cite{kaye2006introduction}:

\begin{equation}
\mathbf{Z} = \begin{bmatrix}
1 &0\\
0 &-1
\end{bmatrix},
\end{equation}
where $\ket{0}\bra{0}-\ket{1}\bra{1}$ is the spectral decomposition form of $\mathbf{Z}$.  Then, the expected value of Pauli-Z operator for $\ket{\psi}$ can be determined as:
\begin{equation}
\bra{\psi} \mathbf{Z} \ket{\psi}= \bra{\psi}\ket{0} \bra{0}\ket{\psi} - \bra{\psi}\ket{1} \bra{1}\ket{\psi}=\lvert \bra{0}\ket{\psi} \rvert^2 - \lvert \bra{1}\ket{\psi} \rvert^2.
\end{equation}
Here $\lvert \bra{0}\ket{\psi} \rvert^2$ and $\lvert \bra{1}\ket{\psi} \rvert^2$ represents the probabilities of $\ket{\psi}$ being in states $\ket{0}$ and $\ket{1}$, respectively. Note that $\bra{a}\ket{b}$ represents the inner product between $\ket{a}$ and $\ket{b}$ in Hilbert space. For $n$th wire or qubit, the measured value ($e_{j_n}$) is regarded as the encoded version of the corresponding element $\hat{\mathbf{x}}_{j_n}$ of the input segment $\hat{\mathbf{x}}_j$. By considering all $n$ qubit measurements, we obtain an encoded version ($\mathbf{e}_j=[e_{j_1}, e_{j_2}, \ldots e_{j_n}]$) of the input segment $\hat{\mathbf{x}}_j$. The encoded signal $\mathbf{e}$ is obtained by temporally concatenating all the encoded segments $\mathbf{e}_j$.
\end{itemize}

\subsection{Models}
Following neural network architectures have been used for the prediction tasks:

\begin{itemize}
\item \emph{Long short-term memory (LSTM) based model:} This model has also been used in \cite{thakur2021dynamic} for mortality prediction. It consists of an LSTM with 256 recurrent units followed by a linear layer with 1 node and sigmoid activation for binary prediction.   
    
\item \emph{Temporal convolution neural networks} \cite{bai2018empirical,martinez2020lipreading}: These models exploit 1-d convolution operations for modeling input time series. In this work, we have used temporal convolutional networks having four temporal blocks followed by a linear layer with $1$ node and sigmoid activation that maps the $64$ dimensional embedding to an output score. Each temporal block consists of two 1-d convolution layers having $64$ filters of size $9$. Also, each convolution layer is followed by a 1-d batch normalization layer, parametric ReLU activation, and a dropout layer with a dropout probability of $0.75$. We have also used a multi-branch temporal convolutional network (\emph{Multi-TCN}) consisting of two multi-branch temporal blocks followed by a linear layer with 1 node and sigmoid activation. Each multi-branch temporal block consists of three branches that process the same input in parallel. Each branch consists of two 1-d convolutional layers having 32 filters where each convolutional layer is followed by batch normalization and parametric ReLU activation. The kernel size of the filters in first, second and third branches is of $5,7$ and $9$, respectively. The last layer of the block is a 1-d conv layer with 96 filters of size 1 that acts as an aggregator and selects the relevant features from all three branches.    
    
\item \emph{Transformer and Vision Transformer:} A transformer \cite{vaswani2017attention} consists of an encoder and a decoder, each formulated by stacking multiple self-attention layers that can capture the global dependencies in the input signals. We employ a transformer encoder having $1$ attention layer with sixteen $256$-dimensional heads followed by two linear layers having $16$ and $F$ nodes. We get an output of shape $T \times F$, where $T$ are the time steps and $F$ is the feature dimensions of the input time series. This encoder output is temporally pooled and given as input to a two-layered MLP classifier having 128 and 1 node to obtain the binary prediction.  Vision transformer (ViT) \cite{dosovitskiy2020image} is another prevalent variant of transformer that is explicitly designed for images. In this work, we also exploit a vision transformer for modeling time series. The architecture is almost identical to the previously described transformer.  In ViT, a learnable $F$-dimensional token is appended to the input time series and is given as input to the MLP classifier (instead of temporally pooled representation as done in classical transformer).  
\end{itemize}

As discussed earlier, we have used single-layer models having either $1$ node for latent binary prediction tasks or $25$ nodes followed by sigmoid activation for latent disorder prediction from the mortality prediction models.

\subsection{Training mortality prediction models}
Irrespective of the data encoding strategy or model architectures, all prediction models are trained using the same parameter setting. Binary cross-entropy is used as the loss function. Adam optimizer with a fixed learning rate of $0.001$ and a batch size of $64$ are used for training the models. Each model is trained to provide the best performance on the validation examples, and the best-performing model configuration is used for evaluating the test or held-out dataset. 

\subsection{Training latent information prediction models}
For training information leakage or latent prediction models, we again followed the same train, validation, and test split that is available for the prediction tasks. For estimating the information leakage from a trained model, we obtained the penultimate layer embedding for all examples. These embedding are used as input representation for training and evaluating the latent information prediction models i.e. gender, ethnicity and disorder prediction models. Binary cross-entropy loss, Adam optimizer with a fixed learning rate of $0.001$ and a batch size of $256$ are used for training the models.

\subsection{Implementation details}
All the experiments are performed using Python. PyTorch is used as a deep learning library. Quantum operations have been simulated using PennyLane \cite{bergholm2018pennylane}. Mutual information for the IB analysis (Fig.~\ref{fig:MI}) has been estimated using \cite{noshad2019scalable}.

\backmatter
\section*{Declarations}

\subsection*{Funding}
This work was supported in part by the National Institute for Health Research (NIHR) Oxford Biomedical Research Centre (BRC), and in part by an InnoHK Project at the Hong Kong Centre for Cerebro-cardiovascular Health Engineering (COCHE). AT is supported by an EPSRC Healthcare Technologies Challenge Award (EP/N020774/1). The views expressed are those of the authors and not necessarily those of the NHS, the NIHR, the Department of Health, InnoHK – ITC, or the University of Oxford.

\subsection*{Conflict of interest} Nothing to disclose.

\subsection*{Ethical approval} Not required. 

\subsection*{Data availability} All datasets used in this study are publicly available. 

\subsection*{Code availability} The code repository is publicly available at \url{github.com/AnshThakur/Quantum-Encoding}.

\subsection*{Authors' Contribution} AT and DAC were responsible for developing the idea of data encoding for data democratization. AT and VA implemented the proposed encoding framework including random projections and random quantum encoding. AT and TZ studied the latent information leakage from the clinical models. AT, JA, and YW designed the experiments and evaluated the impact of data encoding. AT, VA, TZ, and DAC were responsible for drafting the manuscript.

\bibliography{sn-bibliography}
\end{document}


\title[Data Encoding]{Supplementary Document:  Data Encoding For Healthcare Data Democratisation and Information Leakage Prevention}

\maketitle

\section{Dataset Pre-processing}
We have used the existing open source pipelines to obtain the pre-processed form of all datasets. For MIMIC-III, the benchmark\footnote{\url{github.com/YerevaNN/mimic3-benchmarks}} provided in \cite{harutyunyan2019multitask} was used for obtaining the pre-processed time-series representing the ICU stays. FIDDLE pipeline\footnote{\url{physionet.org/content/mimic-eicu-fiddle-feature/1.0.0/}} \cite{tang2020democratizing} was used to obtain the pre-processed version of the eICU dataset for the ARF prediction tasks.   

\noindent Apart from that, we impute the missing values in PhysioNet dataset using \emph{carry forward} approach\footnote{\url{github.com/Ghadeer-Ghosheh/physionet2012-timeseries}}.

\section{List of features in PhysioNet 2012 dataset}
\begin{multicols}{2}
\begin{enumerate}
    \item Alkaline phosphatase
\item Alanine transaminase 
\item Aspartate transaminase
\item Albumin
\item Blood urea nitrogen
\item Bilirubin
\item Cholesterol
\item Creatinine
\item Invasive diastolic arterial blood pressure
\item Fractional inspired oxygen
\item Glasgow Comma Score
\item Glucose
\item Serum bicarbonate
\item Hematocrit
\item Heart rate
\item Height
\item ICU Type - Coronary Care Unit
\item ICU Type - Cardiac Surgery Recovery Unit
\item ICU Type - Medical ICU
\item ICU Type - Surgical ICU
\item Serum potassium 
\item Lactate
\item Invasive mean arterial blood pressure
\item Mechanical ventilation respiration
\item Serum magnesium 
\item Non-invasive diastolic arterial blood pressure
\item Non-invasive mean arterial blood pressure
\item Non-invasive systolic arterial blood pressur
\item Serum sodium
\item Partial pressure of arterial $CO_2$
\item Partial pressure of arterial $O_2$
\item Platelets
\item Respiration rate
\item SAPS-I score
\item SOFA score
\item $O_2$ saturation in hemoglobin 
\item Invasive systolic arterial blood pressure
\item Temperature
\item Troponin-I
\item Troponin-T
\item Urine output
\item White blood cell count
\item Weight
\item Arterial pH

\end{enumerate}
\end{multicols}

\section{List of features in MIMIC-III dataset}
\begin{multicols}{2}
\begin{enumerate}

\item Capillary refill rate-0.0
\item Capillary refill rate-1.0
\item Diastolic blood pressure
\item Fraction inspired oxygen
 \item Glascow coma scale eye opening-2 To Pain
 \item Glascow coma scale eye opening-3 To speech
\item Glascow coma scale eye opening-1 No Response
\item Glascow coma scale eye opening-4 Spontaneously
\item Glascow coma scale eye opening-0 None
 
 \item Glascow coma scale motor response-1 No Movement
 \item Glascow coma scale motor response-3 Abnormal flexion
 \item Glascow coma scale motor response-2 Abnormal extension
 \item Glascow coma scale motor response-4 Flex-withdraws
 \item Glascow coma scale motor response-5 Localizes Pain
 \item Glascow coma scale motor response-6 Obeys Commands
 
 \item Glascow coma scale total-11
 \item Glascow coma scale total-10
 \item Glascow coma scale total-13
 \item Glascow coma scale total-12
 \item Glascow coma scale total-15
 \item Glascow coma scale total-14
 \item Glascow coma scale total-3
 \item Glascow coma scale total-5
 \item Glascow coma scale total-4
 \item Glascow coma scale total-7
 \item Glascow coma scale total-6
 \item Glascow coma scale total-9
 \item Glascow coma scale total-8
 
 \item Glascow coma scale verbal response-1 No Response
 \item Glascow coma scale verbal response-4 Confused
 \item Glascow coma scale verbal response-2 Incomprehensible sounds
 \item Glascow coma scale verbal response-3 Inappropriate Words
 \item Glascow coma scale verbal response-5 Oriented
 
 \item Glucose
 \item Heart Rate
 \item Height
 \item Mean blood pressure
 \item Oxygen saturation
 \item Respiratory rate
 \item Systolic blood pressure
 \item Temperature
 \item Weight
 \item pH
 \item mask-Capillary refill rate
 \item mask-Diastolic blood pressure
 \item mask-Fraction inspired oxygen
 \item mask-Glascow coma scale eye opening
 \item mask-Glascow coma scale motor response
 \item mask-Glascow coma scale total
 \item mask-Glascow coma scale verbal response
 \item mask-Glucose
 \item mask-Heart Rate
 \item mask-Height
 \item mask-Mean blood pressure
 \item mask-Oxygen saturation
 \item mask-Respiratory rate
 \item mask-Systolic blood pressure
 \item mask-Temperature
 \item mask-Weight
 \item mask-pH
    
\end{enumerate}
\end{multicols}

\section{List of features used in the eICU dataset}
\begin{multicols}{2}
\begin{enumerate}

\item Height: (0, 160.0]
\item Height: (160.0, 167.0]
\item Height: (167.0, 172.7]
\item Height: (172.7, 180.0]
\item Height: (180.0, 612.6]
 \item Weight: (0.0, 63.0]
 \item Weight: (63.0, 74.0]
 \item Weight: (74.0, 85.0]
 \item Weight: (85.0, 100.7]
 \item Weight: (100.7, 953.0]
 \item age: (17.999, 49.0]
 \item age: (49.0, 60.0]
 \item age: (60.0, 69.0]
 \item age: (69.0, 78.0]
 \item age: (78.0, 89.0]
 \item age: $>$89
 
 \item Airway type: No Artificial Airway
 \item Apache Admission value: Acid-base/electrolyte disturbance
 \item Apache Admission value: Angina, unstable 
\item Apache Admission value: Bleeding, GI-location unknown
 \item Apache Admission value: Bleeding, lower GI
 \item Apache Admission value: Bleeding, upper GI
 \item Apache Admission value: CHF, congestive heart failure
 \item Apache Admission value: CVA, cerebrovascular accident
 \item Apache Admission value: Coma/change in level of consciousness
 \item Apache Admission value: Diabetic ketoacidosis
 \item Apache Admission value: Embolus, pulmonary
 \item Apache Admission value: Emphysema/bronchitis
 \item Apache Admission value: Endarterectomy, carotid
 \item Apache Admission value: Hemorrhage/hematoma, intracranial
 \item Apache Admission value: Hypertension, uncontrolled
 \item Apache Admission value: Infarction, acute myocardial
 \item Apache Admission value: Pneumonia, bacterial
 \item Apache Admission value: Renal failure, acute
 \item Apache Admission value: Rhythm disturbance (atrial, supraventricular)
 \item Apache Admission value: Rhythm disturbance (conduction defect)
 \item Apache Admission value: Seizures (primary-no structural brain disease)
 \item Apache Admission value: Sepsis, GI
 \item Apache Admission value: Sepsis, cutaneous/soft tissue
 \item Apache Admission value: Sepsis, pulmonary
 \item Apache Admission value: Sepsis, renal/UTI (including bladder)
 \item Apache Admission value: Sepsis, unknown
 \item Hospital admit offset: (-529268.001, -1790.0]
 \item Hospital admit offset: (-1790.0, -389.0]
 \item Hospital admit offset: (-389.0, -173.0]
  \item Hospital admit offset: (-173.0, -50.0]
  \item Hospital admit offset: (-50.0, 118121.0]
  \item Hospital admit source: Acute Care/Floor
 \item Hospital admit source: Direct Admit
 \item Hospital admit source: Emergency Department
 \item Hospital admit source: Floor
 \item Hospital admit source: Operating Room
 \item Hospital admit source: Other Hospital 
 \item Hospital admit source: Recovery Room
 \item Hospital admit source: Step-Down Unit (SDU)
 \item Hospital Id: 110
 \item Hospital Id: 122
 \item Hospital Id: 141
 \item Hospital Id: 148
 \item Hospital Id: 165
 \item Hospital Id: 167
 \item Hospital Id: 171
 \item Hospital Id: 176
 \item Hospital Id: 183
 \item Hospital Id: 199
 \item Hospital Id: 208
 \item Hospital Id: 243
 \item Hospital Id: 252
 \item Hospital Id: 264
 \item Hospital Id: 281
 \item Hospital Id: 283
 \item Hospital Id: 300
 \item Hospital Id: 307
 \item Hospital Id: 331
 \item Hospital Id: 338
 \item Hospital Id: 365
 \item Hospital Id: 394
 \item Hospital Id: 411
 \item Hospital Id: 413
 \item Hospital Id: 417
 \item Hospital Id: 420
 \item Hospital Id: 435
 \item Hospital Id: 443
 \item Hospital Id: 458
 \item Hospital Id: 73
 \item Unit admit source: Acute Care/Floor
 \item Unit admit source: Direct Admit
 \item Unit admit source: Emergency Department
 \item Unit admit source: Floor
 \item Unit admit source: ICU
 \item Unit admit source: ICU to SDU
 \item Unit admit source: Operating Room
 \item Unit admit source: Other Hospital
 \item Unit admit source: Other ICU
 \item Unit admit source: PACU
 \item Unit admit source: Recovery Room
 \item Unit admit source: Step-Down Unit (SDU)',
 \item Unit stay type: admit
 \item Unit stay type: readmit
 \item Unit stay type: stepdown/other
 \item Unit stay type: transfer
 \item Unit type: CCU-CTICU
 \item Unit type: CSICU
 \item Unit type: CTICU
 \item Unit type: Cardiac ICU
 \item Unit type: MICU
 \item Unit type: Med-Surg ICU
 \item Unit type: Neuro ICU

\item Heart Rate mask
\item Non-Invasive BP Diastolic mask
\item Non-Invasive BP Systolic mask
\item $O_2$ Saturation mask
\item Respiratory Rate mask
 \item CVP (-34.001, 5.0]
 \item CVP (5.0, 8.0]
 \item CVP (8.0, 11.0]
 \item CVP (11.0, 15.0]
 \item CVP (15.0, 396.0]
 \item Invasive BP Diastolic ($<$48.0)
 \item Invasive BP Diastolic (48.0, 56.0]
 \item Invasive BP Diastolic (56.0, 62.0]
 \item Invasive BP Diastolic (62.0, 71.0]
 \item Invasive BP Diastolic ($>$71.0)

 \item Invasive BP Mean (-50.001, 68.0]
 \item Invasive BP Mean (68.0, 76.0]
 \item Invasive BP Mean (76.0, 84.0]
 \item Invasive BP Mean (84.0, 94.0]
 \item Invasive BP Mean ($>$94.0)

 \item Invasive BP Systolic ($<$104.0)
 \item Invasive BP Systolic (104.0, 118.0]
 \item Invasive BP Systolic (118.0, 131.0]
 \item Invasive BP Systolic (131.0, 146.0]
 \item Invasive BP Systolic ($>$146.0)
  
 \item Non-Invasive BP Mean ($<$67.0)
 \item Non-Invasive BP Mean (67.0, 76.0]
 \item Non-Invasive BP Mean (76.0, 84.0]
 \item Non-Invasive BP Mean (84.0, 95.0]
 \item Non-Invasive BP Mean ($>$95.0)

 \item $O_2$ Admin Device: BiPAP
 \item $O_2$ Admin Device: BiPAP/CPAP
 \item $O_2$ Admin Device: NC
 \item $O_2$ Admin Device: RA
 \item $O_2$ Admin Device: nasal cannula
 \item $O_2$ Admin Device: nc
 \item $O_2$ Admin Device: non-rebreather
 \item $O_2$ Admin Device: other
 \item $O_2$ Admin Device: ra
 \item $O_2$ Admin Device: room air
 \item $O_2$ Admin Device: trach collar
 \item $O_2$ Admin Device: ventilator
 \item $O_2$ Admin Device: venturi mask
  \item $O_2$ L\%: ($<$2.0)
 \item $O_2$ L\%: (2.0, 3.0]
 \item $O_2$ L\%: (3.0, 6.0]
 \item $O_2$ L\%: $>$6.0
 \item Temperature (C): ($<$36.4)
 \item Temperature (C): (36.4, 36.7]
 \item Temperature (C): (36.7, 36.9]
  \item Temperature (C): (36.9, 37.2]
  \item Temperature (C): ($>$37.2)
 \item Temperature (F): ($<$97.5)
 \item Temperature (F): (97.5, 98.1]
 \item Temperature (F): (98.1, 98.4]
 \item Temperature (F): (98.4, 99.0]
 \item Temperature (F): ($>$99.0)

 \item Temperature Location: (-0.001, 1.0]
  \item Temperature Location: (1.0, 4.0]
  \item Temperature Location: TA
  \item Temperature Location: AXILLARY
 \item Temperature Location: BLADDER
 \item Temperature Location: Core urinary catheter
 \item Temperature Location: Forehead
 \item Temperature Location: Oral
\item Temperature Location: PA CATHETER
 \item Temperature Location: Rectal
 \item Temperature Location: Skin Sensor
 \item Temperature Location: TEMPORAL
 \item Temperature Location: TEMPORAL ARTERY
 \item Temperature Location: TYMPANIC
 \item Temperature Location: Temporal Artery Scan
 \item Temperature Location: Temporal scan
 \item Temperature Location: core
 \item Temperature Location: undocumented

 \item Non-Invasive BP Diastolic delta time: (-0.001, 1.0]
 \item Non-Invasive BP Diastolic delta time: (-0.001, 1.0]
 \item Heart Rate: (-0.001, 68.0]
 \item Heart Rate: (68.0, 78.0]
 \item Heart Rate: (78.0, 88.0]
 \item Heart Rate: (88.0, 100.0]
 \item Heart Rate: (100.0, 300.0]
 \item Non-Invasive BP Diastolic: (-0.001, 54.0]
\item Non-Invasive BP Diastolic: (54.0, 61.0]
 \item Non-Invasive BP Diastolic: (61.0, 69.0]
 \item Non-Invasive BP Diastolic: (69.0, 79.0]
 \item Non-Invasive BP Diastolic: (79.0, 866.0]
 \item Non-Invasive BP Systolic: (-0.001, 102.0]
  \item Non-Invasive BP Diastolic: (102.0, 114.0]
  \item Non-Invasive BP Diastolic: (114.0, 127.0]
 \item Non-Invasive BP Systolic: (127.0, 142.0]
  \item Non-Invasive BP Diastolic: (142.0, 12065.0]

\item $O_2$ Saturation: (-0.001, 95.0]
\item $O_2$ Saturation: (95.0, 96.0]
\item $O_2$ Saturation: (96.0, 98.0]
 \item $O_2$ Saturation: (98.0, 99.0]
 \item $O_2$ Saturation: (99.0, 999.0]
\item Respiratory Rate: (-0.001, 15.0]
 \item Respiratory Rate: (15.0, 17.0]
 \item Respiratory Rate: (17.0, 20.0]
 \item Respiratory Rate:  (20.0, 24.0]
 \item Respiratory Rate: (24.0, 912.0]
  \item Heart Rate min: (-0.001, 67.0]
 \item Heart Rate min: (67.0, 77.0]
 \item Heart Rate min: (77.0, 87.0]
 \item Heart Rate min: (87.0, 99.0]
 \item Heart Rate min: (99.0, 293.0]
 \item Heart Rate max: (-0.001, 69.0]
  \item Heart Rate max: (69.0, 79.0]
  \item Heart Rate max: (79.0, 88.0]
  \item Heart Rate max: (88.0, 101.0]
  \item Heart Rate max: (101.0, 959.0]
 \item Heart Rate mean: (-0.001, 68.0]
 \item Heart Rate mean: (68.0, 78.0]
 \item Heart Rate mean: (78.0, 87.5]
 \item Heart Rate mean: (87.5, 100.0]
\item Heart Rate mean: (100.0, 527.0]

 \item Non-Invasive BP Diastolic min: (-0.001, 52.0]
  \item Non-Invasive BP Diastolic min: (52.0, 60.0]
  \item Non-Invasive BP Diastolic min: (60.0, 68.0]
  \item Non-Invasive BP Diastolic min: (68.0, 78.0]
  \item Non-Invasive BP Diastolic min: (78.0, 777.0]
  \item Non-Invasive BP Diastolic max: (-0.001, 55.0]
 \item Non-Invasive BP Diastolic max: (55.0, 63.0]
 \item Non-Invasive BP Diastolic max: (63.0, 70.0]
 \item Non-Invasive BP Diastolic max: (70.0, 80.0]
 \item Non-Invasive BP Diastolic max: (80.0, 6078.0]
 \item Non-Invasive BP Diastolic mean: (-0.001, 54.0]
 \item Non-Invasive BP Diastolic mean: (54.0, 61.25]
 \item Non-Invasive BP Diastolic mean: (61.25, 69.0]
 \item Non-Invasive BP Diastolic mean: (69.0, 79.0]
 \item Non-Invasive BP Diastolic mean: (79.0, 1578.5]
 
 \item Non-Invasive BP Systolic min: (-0.001, 100.0]
 \item Non-Invasive BP Systolic min: (100.0, 113.0]
 \item Non-Invasive BP Systolic min:(113.0, 125.0]
 \item Non-Invasive BP Systolic min: (125.0, 141.0]
 \item Non-Invasive BP Systolic min: (141.0, 12065.0]
 
 \item Non-Invasive BP Systolic max: (-0.001, 104.0]
 \item Non-Invasive BP Systolic max: (104.0, 116.0]
 \item Non-Invasive BP Systolic max: (116.0, 128.0]
 \item Non-Invasive BP Systolic max: (128.0, 144.0]
 \item Non-Invasive BP Systolic max: (144.0, 12065.0]
 \item Non-Invasive BP Systolic mean: (-0.001, 102.0]
 \item Non-Invasive BP Systolic mean: (102.0, 114.0]
 \item Non-Invasive BP Systolic mean: (114.0, 126.5]
 \item Non-Invasive BP Systolic mean: (126.5, 142.0]
 \item Non-Invasive BP Systolic mean: (142.0, 12065.0]

 \item $O_2$ Saturation min: (-0.001, 94.0]
 \item $O_2$ Saturation min: (94.0, 96.0]
 \item $O_2$ Saturation min: (96.0, 98.0]
 \item $O_2$ Saturation min: (98.0, 99.0]
 \item $O_2$ Saturation min: (99.0, 999.0]
 \item $O_2$ Saturation max: (-0.001, 95.0]
 \item $O_2$ Saturation max: (95.0, 97.0]
 \item $O_2$ Saturation max: (97.0, 98.0]
 \item $O_2$ Saturation max: (98.0, 100.0]
 \item $O_2$ Saturation mean: (-0.001, 95.0]
 \item $O_2$ Saturation mean: (95.0, 96.0]
 \item $O_2$ Saturation mean: (96.0, 98.0]
 \item $O_2$ Saturation mean: (98.0, 99.0]
 \item $O_2$ Saturation mean: (99.0, 999.0]
 \item Respiratory Rate min: (-0.001, 14.0]
 \item Respiratory Rate min: (14.0, 17.0]
 \item Respiratory Rate min: (17.0, 19.0]
 \item Respiratory Rate min: (19.0, 23.0]
 \item Respiratory Rate min: (23.0, 912.0]
 \item Respiratory Rate max: (-0.001, 15.0]
 \item Respiratory Rate max: (15.0, 18.0]
 \item Respiratory Rate max: (18.0, 20.0]
 \item Respiratory Rate max: (20.0, 24.0]
 \item Respiratory Rate max: (24.0, 2122.0]
 \item Respiratory Rate mean: (-0.001, 15.0]
 \item Respiratory Rate mean: (15.0, 17.667]
  \item Respiratory Rate mean: (17.667, 20.0]
  \item Respiratory Rate mean: (20.0, 23.25]
  \item Respiratory Rate mean: (23.25, 912.0]

\end{enumerate}
\end{multicols}

\newpage
\section{List of phenotypes or patient disorders in MIMIC-III dataset}
\begin{multicols}{2}
\begin{enumerate}
    \item Acute and unspecified renal failure
\item Acute cerebrovascular disease
\item Acute myocardial infarction
\item Cardiac dysrhythmias 
\item Chronic kidney disease
\item Chronic obstructive pulmonary disease
\item Complications of surgical/medical care 
\item Conduction disorders 
\item Congestive heart failure; nonhypertensive
\item Coronary atherosclerosis and related
\item Diabetes mellitus with complications
\item Diabetes mellitus without complication 
\item Disorders of lipid metabolism
\item Essential hypertension
\item Fluid and electrolyte disorders
\item Gastrointestinal hemorrhage
\item Hypertension with complications
\item Other liver diseases 
\item Other lower respiratory disease
\item Other upper respiratory disease
\item Pleurisy; pneumothorax; pulmonary collapse
\item Pneumonia
\item Respiratory failure; insufficiency; arrest
\item Septicemia (except in labor)
\item Shock

\end{enumerate}
\end{multicols}

\begin{figure}[H]
\centering
\begin{subfigure}{.495\textwidth}
    \centering
    \includegraphics[scale=0.24]{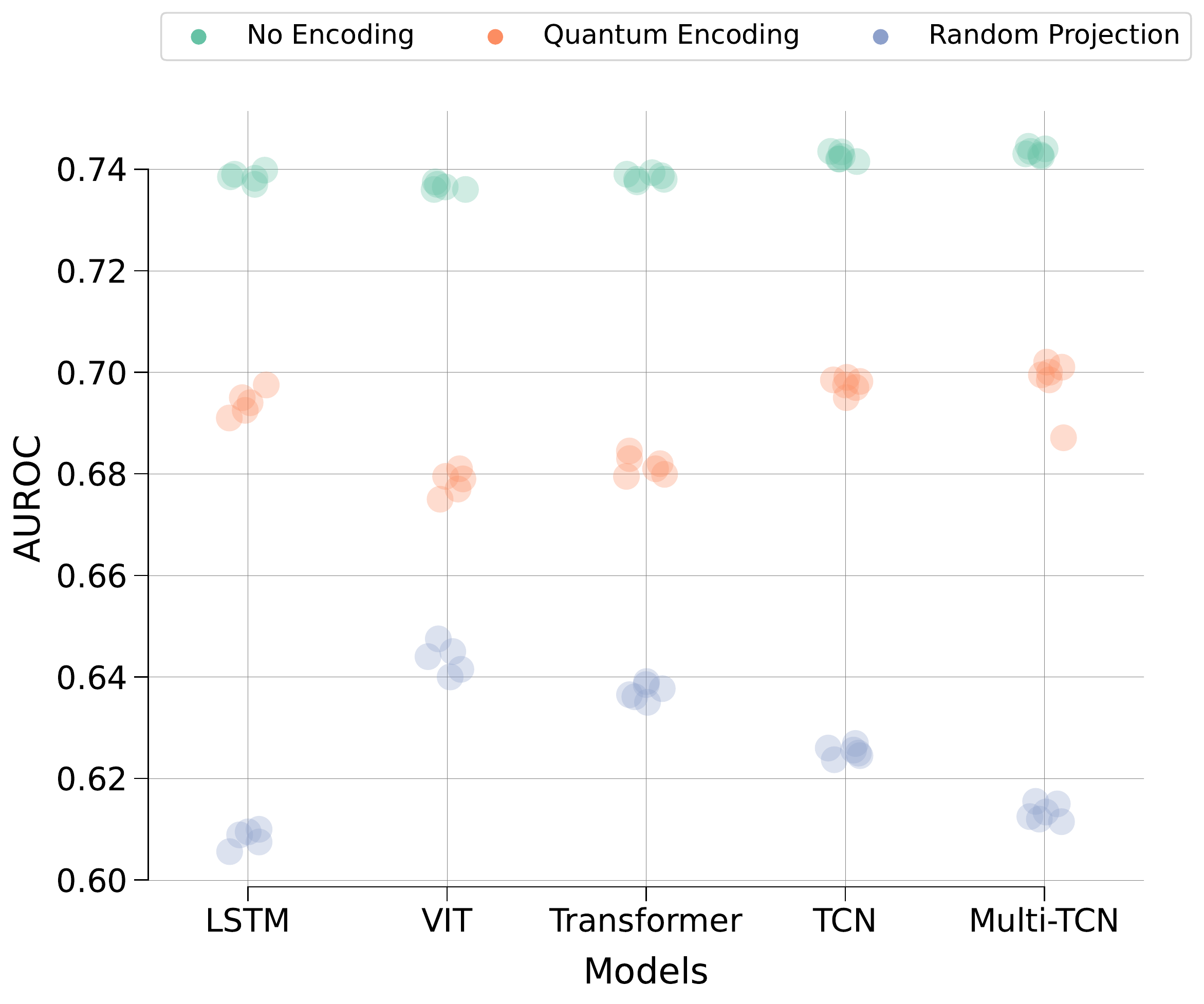}
    \vspace*{0.001mm}
\end{subfigure}%
\begin{subfigure}{.505\textwidth}
    \centering
    \includegraphics[scale=0.232]{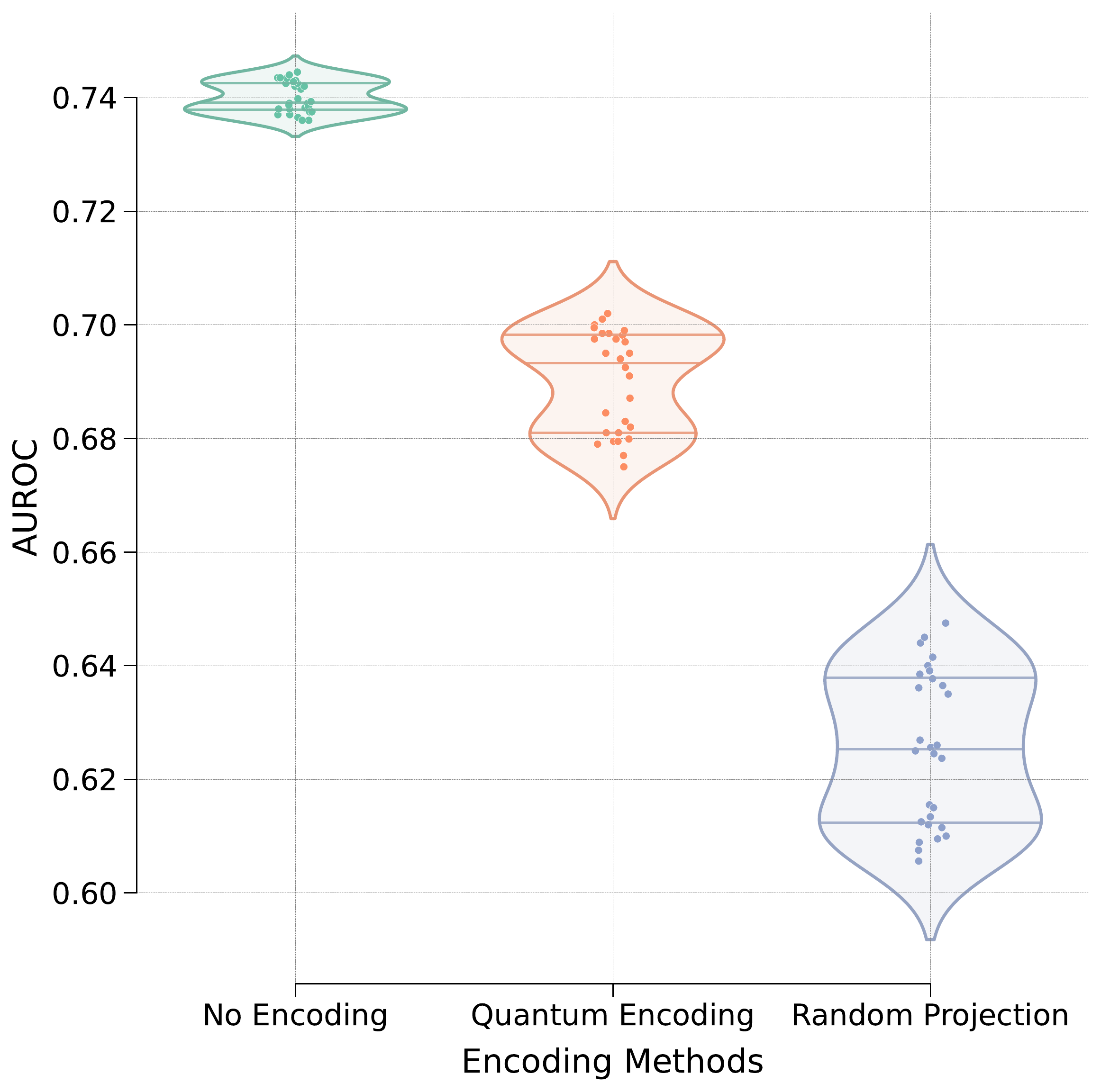}
\end{subfigure}

\caption[short]{(a) Performance of LSTM, vision transformer (ViT), transformer, temporal convolutional network (TCN) and multi-branch temporal convolutional network (Multi-TCN) for phenotype prediction. (b) Aggregate performance of all models as a function of encoding method.}
\label{fig:s1}
\end{figure}

\section{Performance on the encoded data for the task of Phenotyping}

As discussed in Results (main document), along with mortality labels, we also have information about the $25$ phenotypes or disorders corresponding to each ICU stay. In the main text, we used these disorders for evaluating the information leakage from the trained mortality prediction models using a multi-label multi-class setup. 

\vspace{0.1cm}
Here, we train all the models discussed in the main text for directly phenotyping each ICU stay. The last layers of these models were changed to have $25$ nodes followed by sigmoid activation to favour multi-label multi-class predictions. The train, test and validation setup used for mortality prediction is also used here. Adam optimiser with a learning of $0.001$ and a batch-size of $64$ is used for training all models.

\vspace{0.1cm}
Fig.~\ref{fig:s1} illustrates the performance of different models as a function of encoding method for the task of phenotyping. Similar to the binary prediction experiments, the relation between encoding methods and model performance is almost identical. Both random projection and quantum encoding results in a noticeable drop in performance. However, this drop is much more bearable in the case of quantum encoding.

\begin{figure}[h]
    \centering
    \includegraphics[scale=0.35]{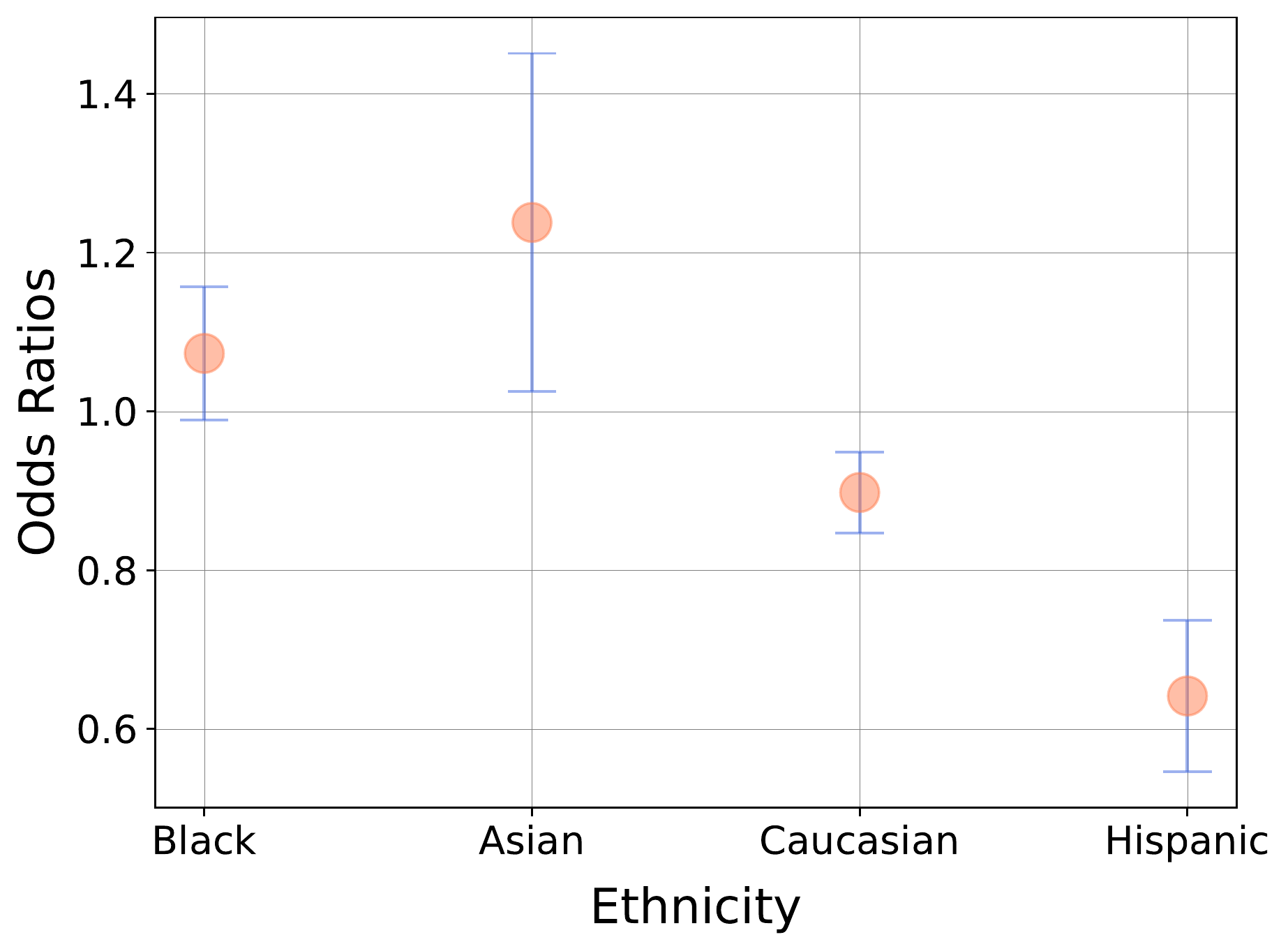}

    \caption{Odds ratios of different ethnicity to the acute respiratory failure in the eICU dataset.}
    \label{fig:odds_eth}
\end{figure}%

\section{Odds Ratio: Ethnicity vs Acute Respiration Failure in eICU dataset}
The analysis of Fig.~\ref{fig:odds_eth} highlights the odds ratio for \emph{Black}, \emph{Asian} and \emph{Caucasian} are close to $1$. This shows that there is no apparent association between ethnicity and ARF in the eICU dataset. Despite that, we are able to predict the ethnicity of the patients from trained ARF models effectively (Fig 4 of the manuscript).  

\vspace{-1cm}
\begin{figure}[H]
\centering
\begin{subfigure}{\textwidth}
\centering
\includegraphics[scale=0.28]{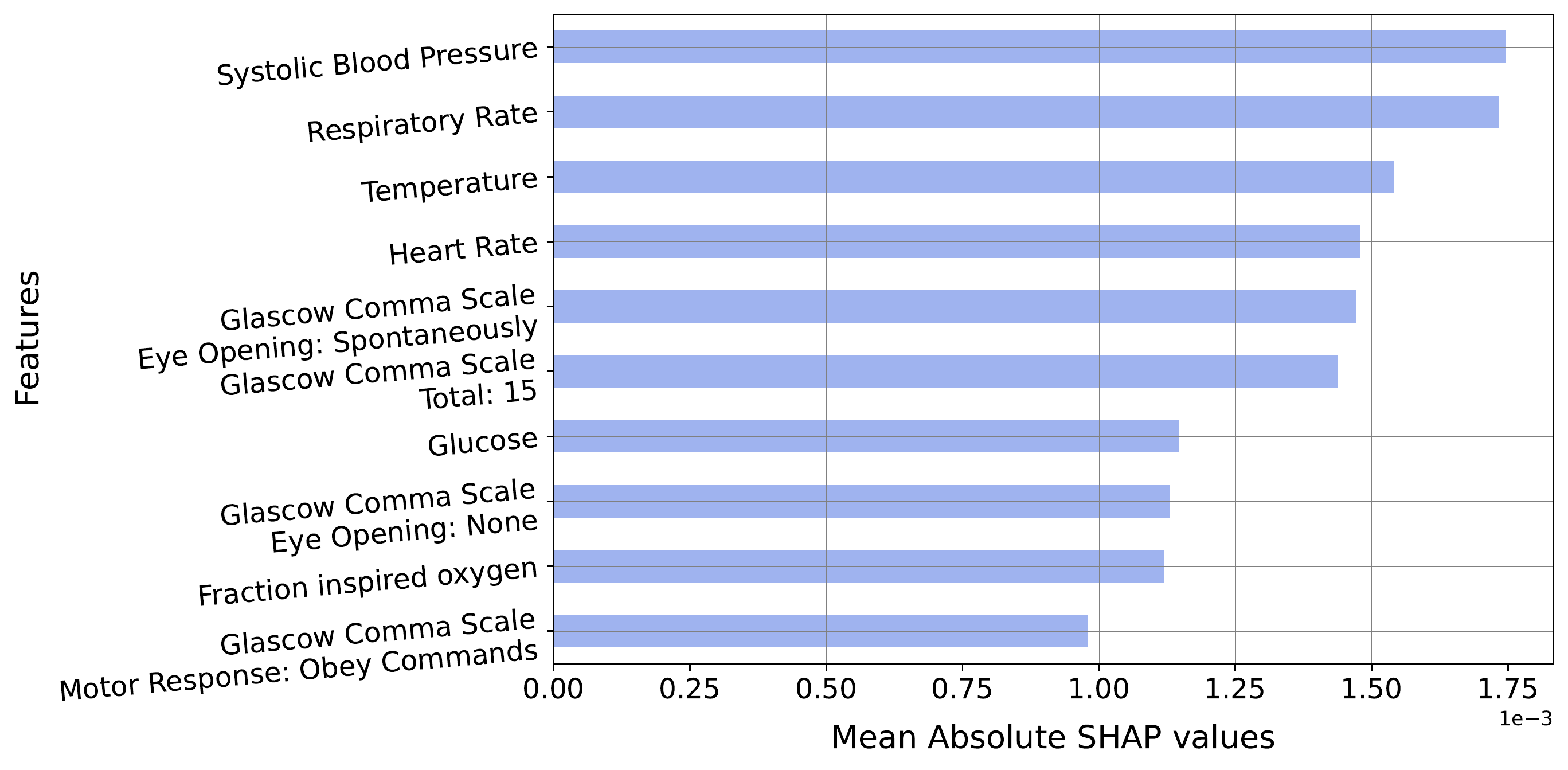}
\caption{SHAP analysis of LSTM trained on MIMIC-III.}
\end{subfigure}

\begin{subfigure}{\textwidth}
\centering
\includegraphics[scale=0.28]{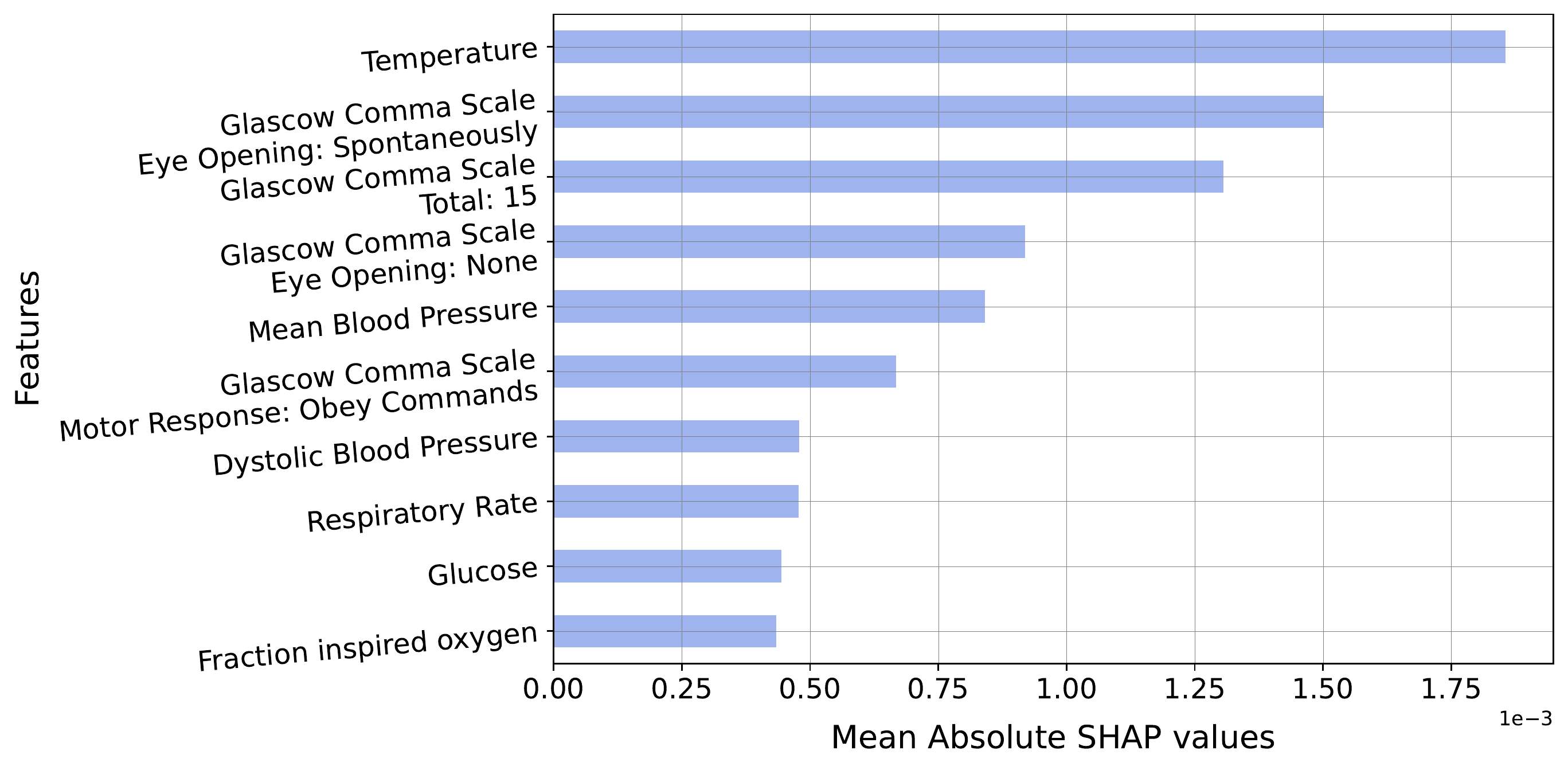}
\caption{SHAP analysis of LSTM trained on quantum encoded MIMIC-III.}
\end{subfigure}

\begin{subfigure}{\textwidth}
\centering
\includegraphics[scale=0.28]{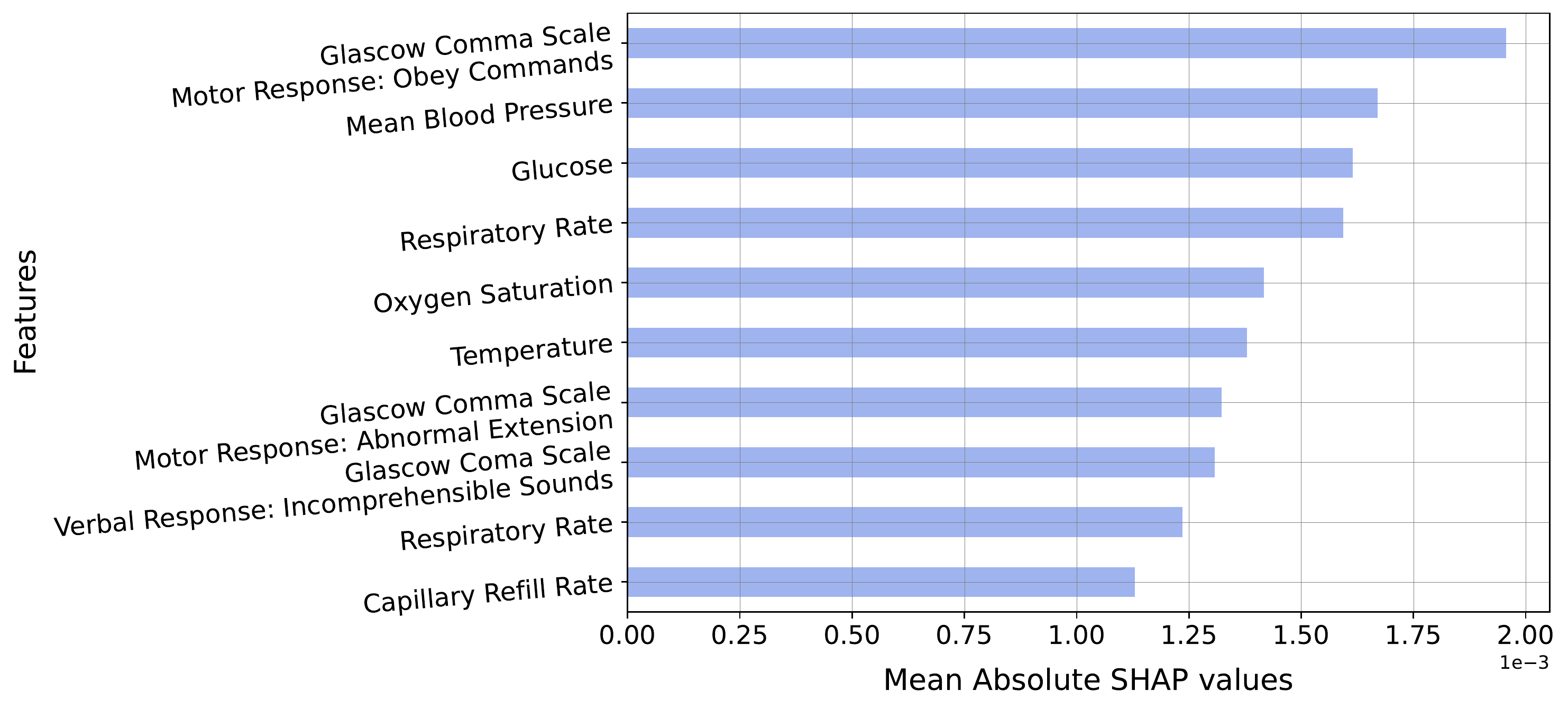}
\caption{SHAP analysis of LSTM trained on MIMIC-III data encoded by random projections.}
\end{subfigure}
\caption[short]{A comparison of SHAP-based feature importance in LSTM models trained on (a) original, (b) quantum encoded, and (c) randomly projected versions of the MIMIC-III dataset.}
\label{fig:shap1}
\end{figure}
\section{SHAP Analysis of LSTM models trained on MIMIC-III}
Fig.\ref{fig:shap1} illustrates SHAP plots for LSTM models trained on original and encoded MIMIC-III data.

\section{Visual inspection of encoded PhysioNet examples}

\begin{figure}[H]
\centering
\begin{subfigure}{\textwidth}
    \centering
    \includegraphics[scale=0.2]{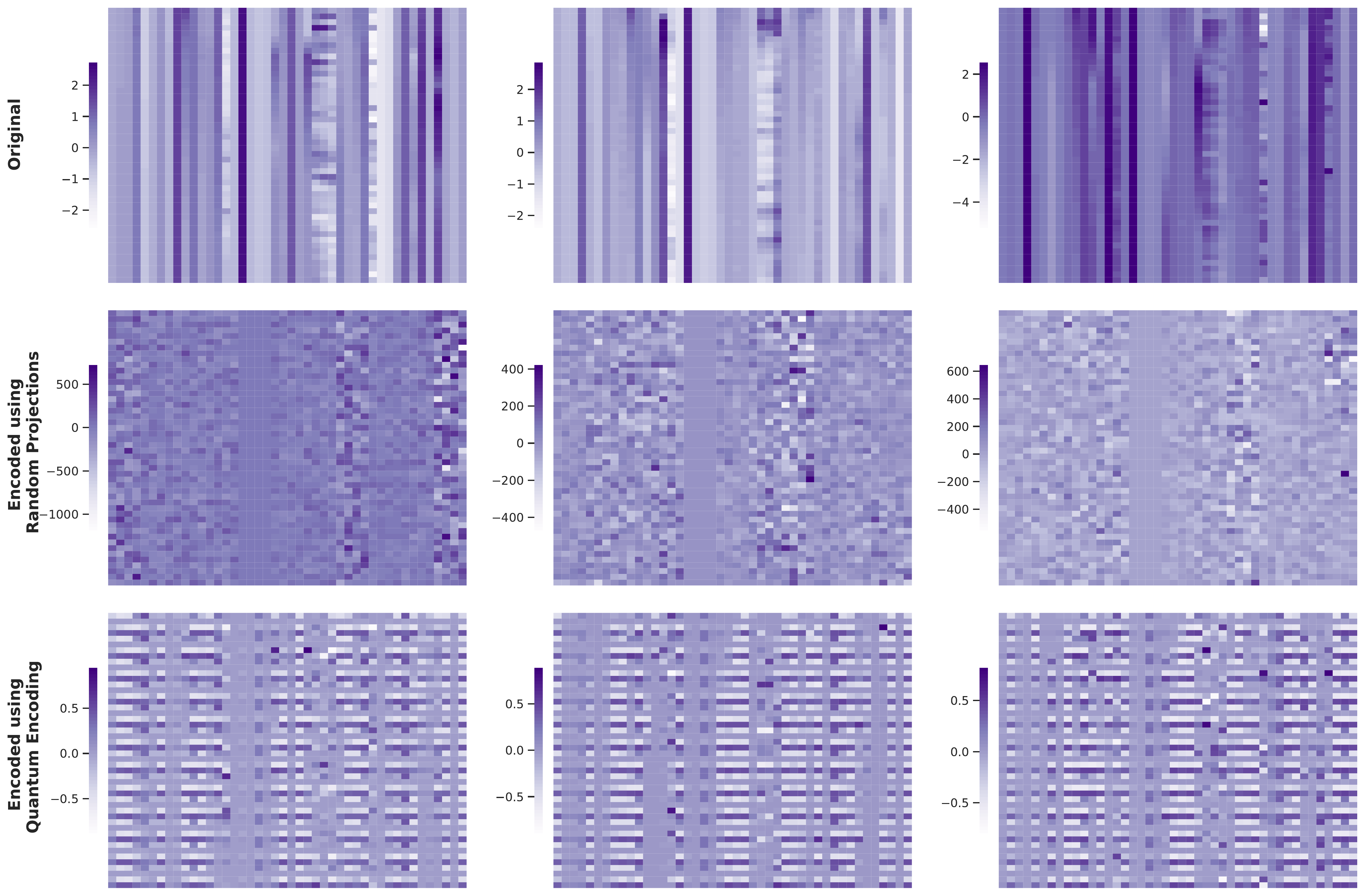}
    \caption{Three negative examples from PhysioNet.}
\end{subfigure}%

\begin{subfigure}{\textwidth}
    \centering
    \includegraphics[scale=0.2]{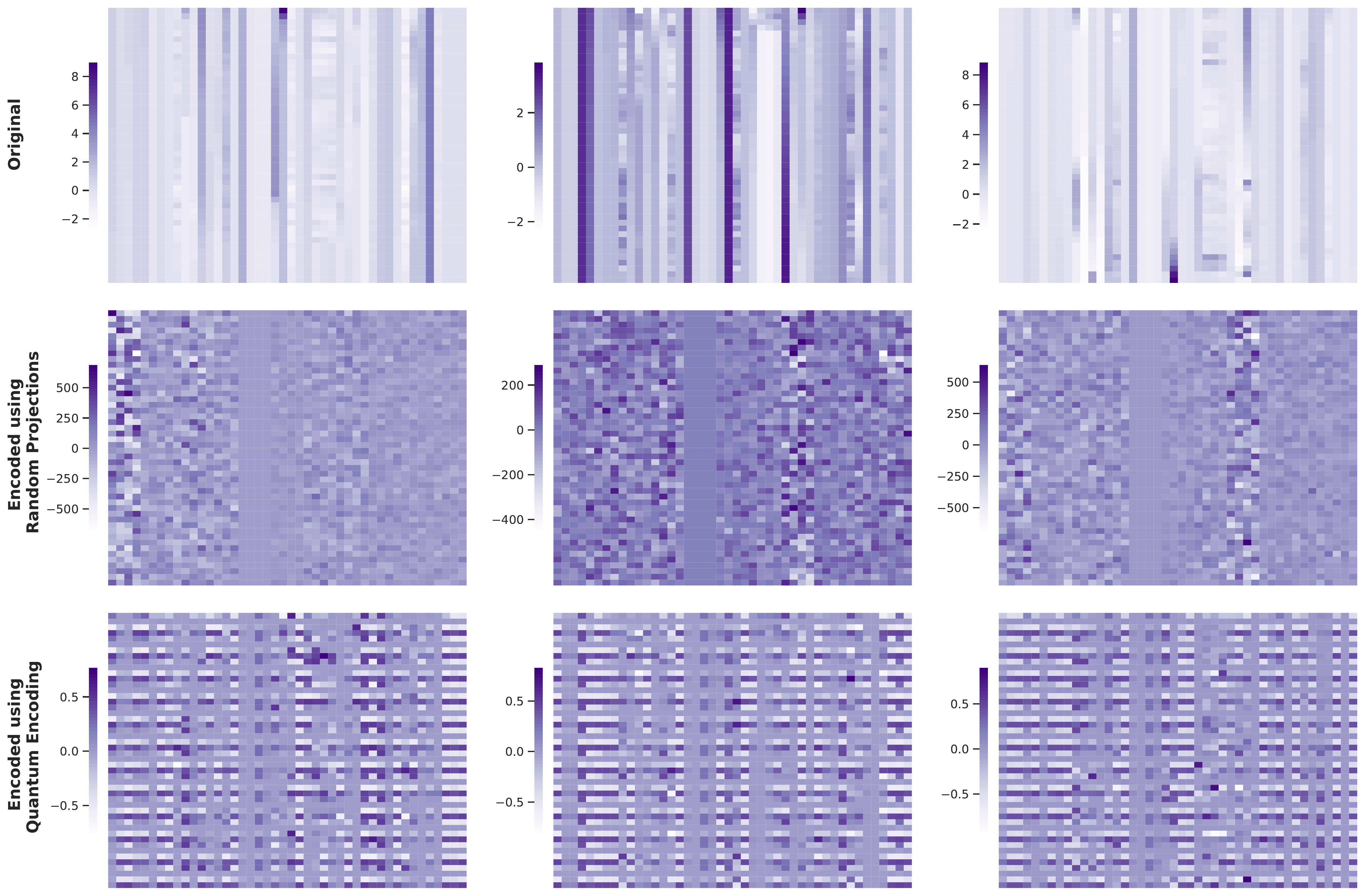}

    \caption{Three positive examples from PhysioNet.}
\end{subfigure}%

\caption[short]{Heat maps illustrating the differences in magnitude and trends of the original and encoded time-series examples. Each row represents an input time-series and its encoded versions.}
\label{fig:s4}
\end{figure}

Fig.\ref{fig:s4} illustrates the differences in encoded and original time-series examples from PhysioNet dataset. 

\newpage 
\section{Latent prediction of chronic and acute disorders from all models}

\begin{figure}[H]
\centering
\begin{subfigure}{\textwidth}
    \centering
    \includegraphics[scale=0.15,trim={0 1.25cm 0 0},clip]{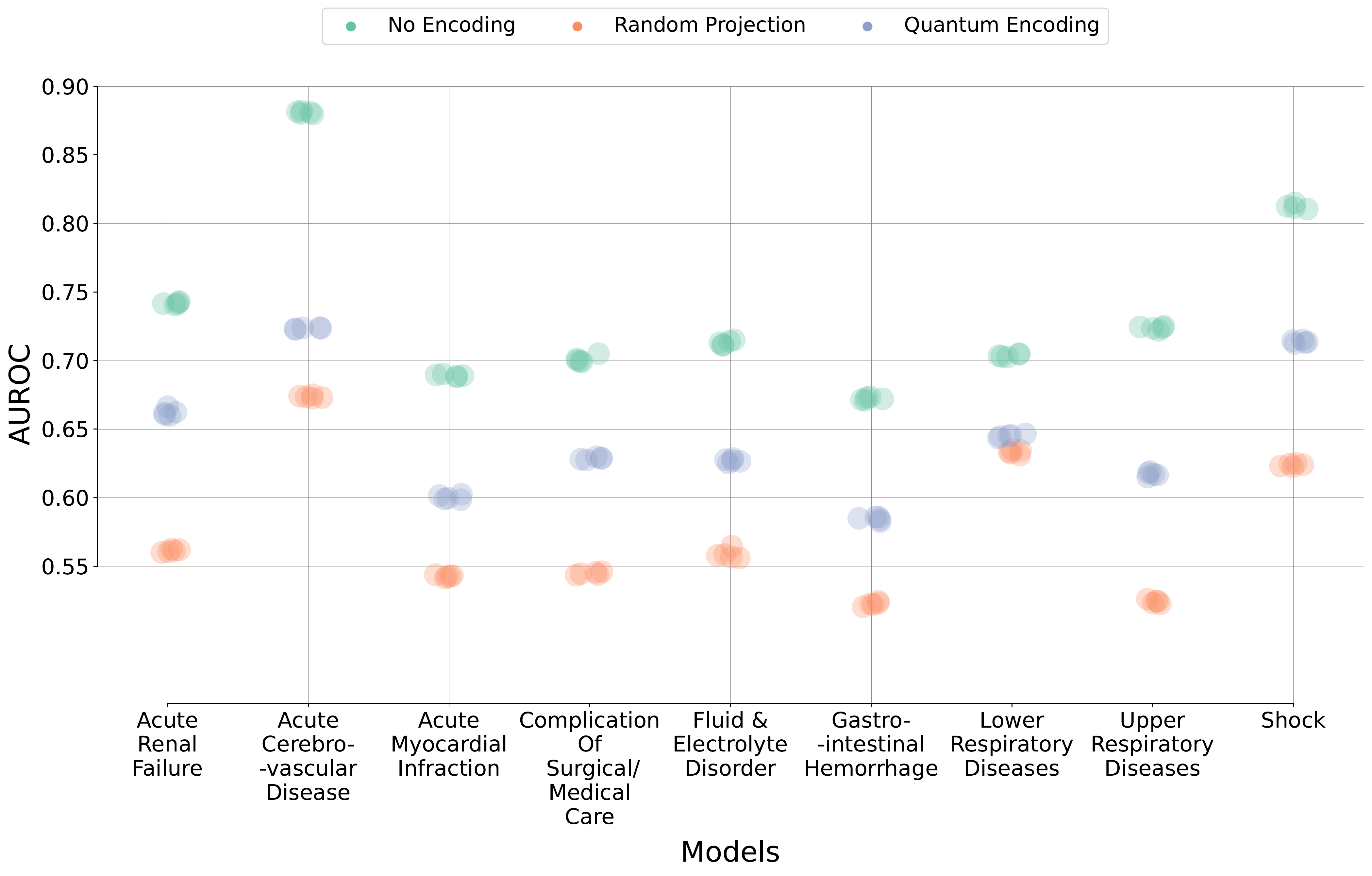}
    \caption{Transformer}
\end{subfigure}%


\begin{subfigure}{\textwidth}
    \centering
    \includegraphics[scale=0.15,trim={0 1.25cm 0 0},clip]{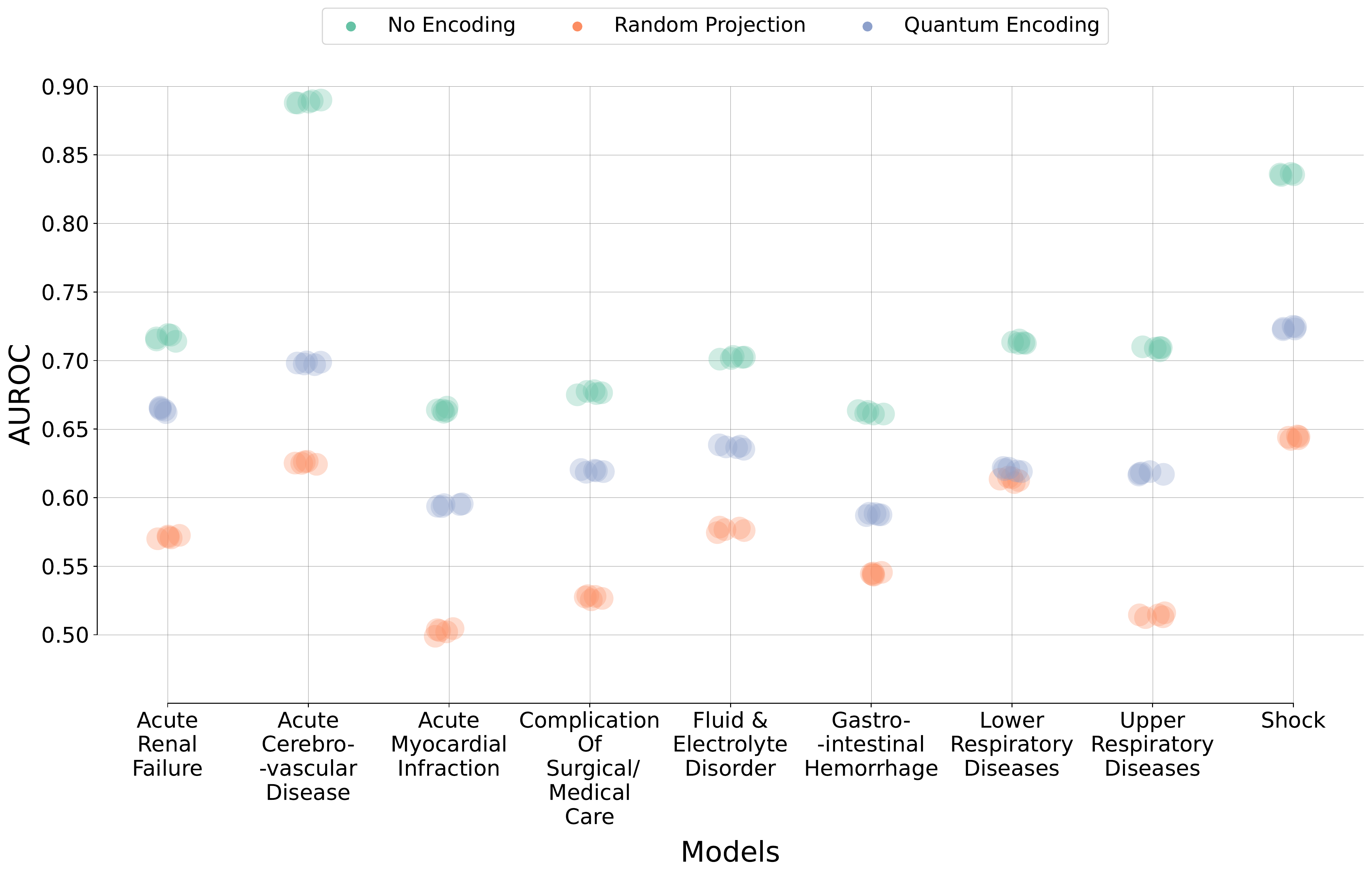}
    \caption{Temporal convolutional network}
\end{subfigure}%


\begin{subfigure}{\textwidth}
    \centering
    \includegraphics[scale=0.15,trim={0 1.25cm 0 0},clip]{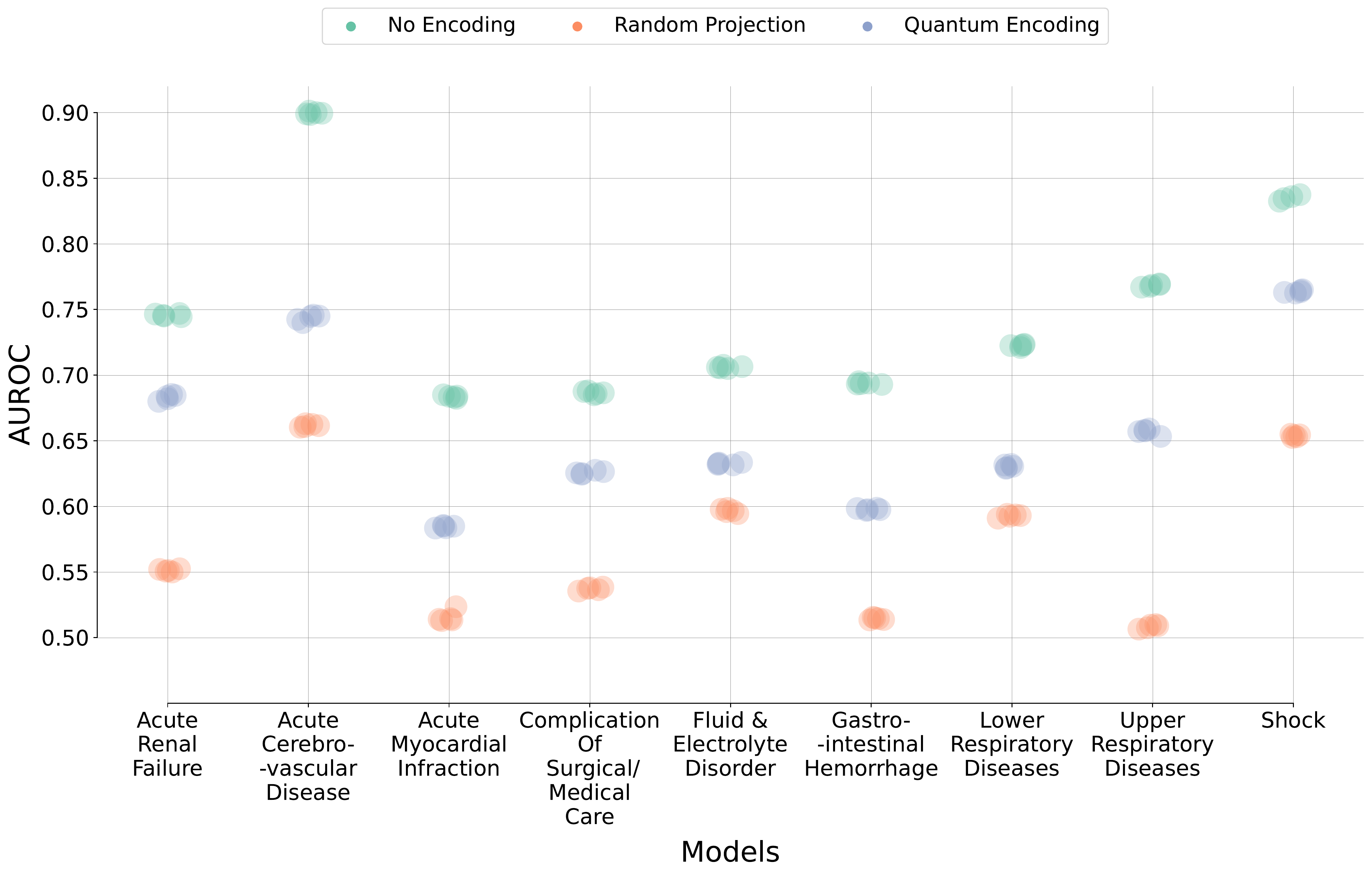}
    \caption{Multi-branch temporal convolutional network.}
\end{subfigure}%

\caption[short]{Latent acute disorder prediction from different mortality prediction models.}
\label{fig:s2}
\end{figure}

\begin{figure}[H]
\centering
\begin{subfigure}{\textwidth}
    \centering
    \includegraphics[scale=0.18,trim={0 1.25cm 0 0},clip]{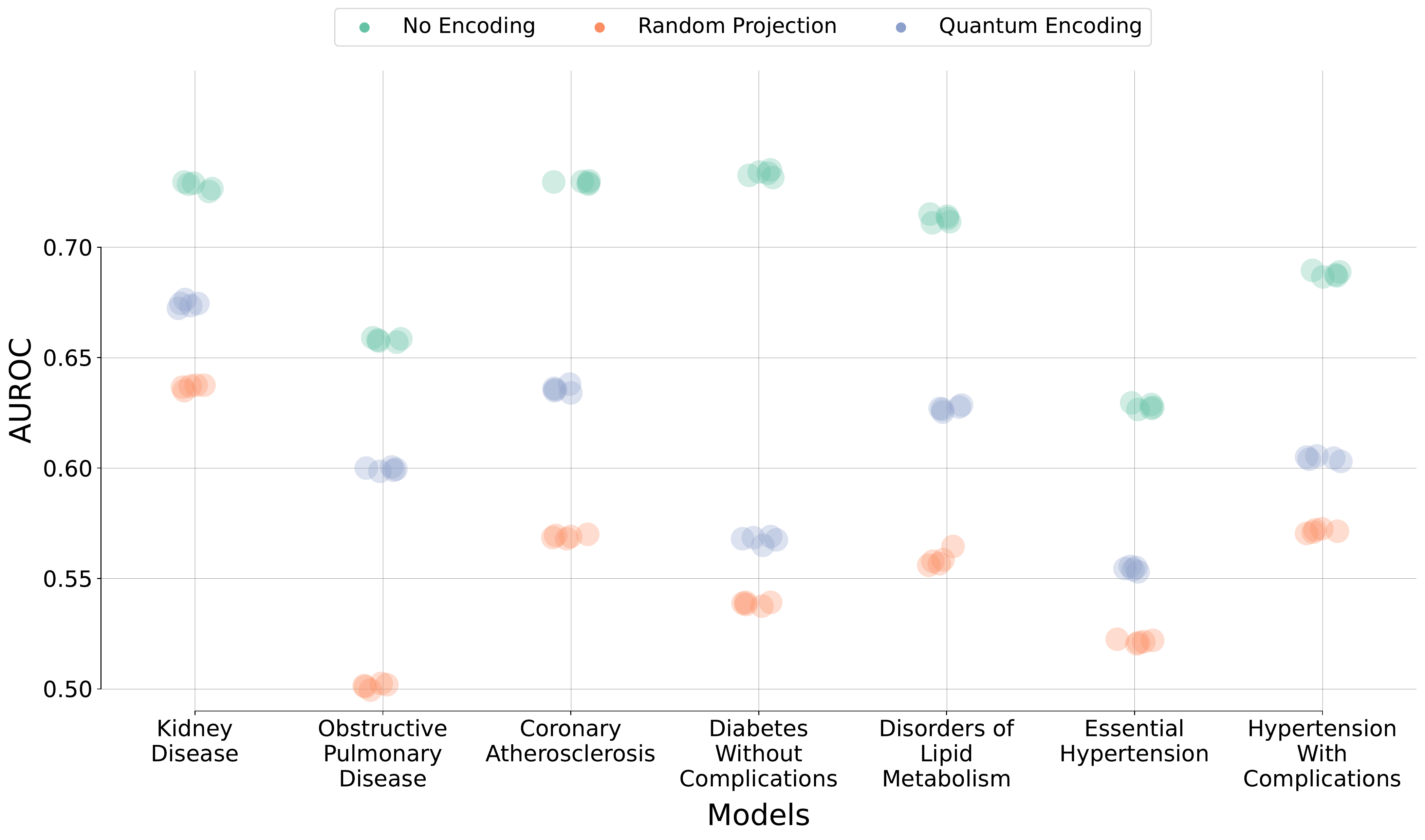}
    \caption{Transformer}
\end{subfigure}%


\begin{subfigure}{\textwidth}
    \centering
    \includegraphics[scale=0.18,trim={0 1.25cm 0 0},clip]{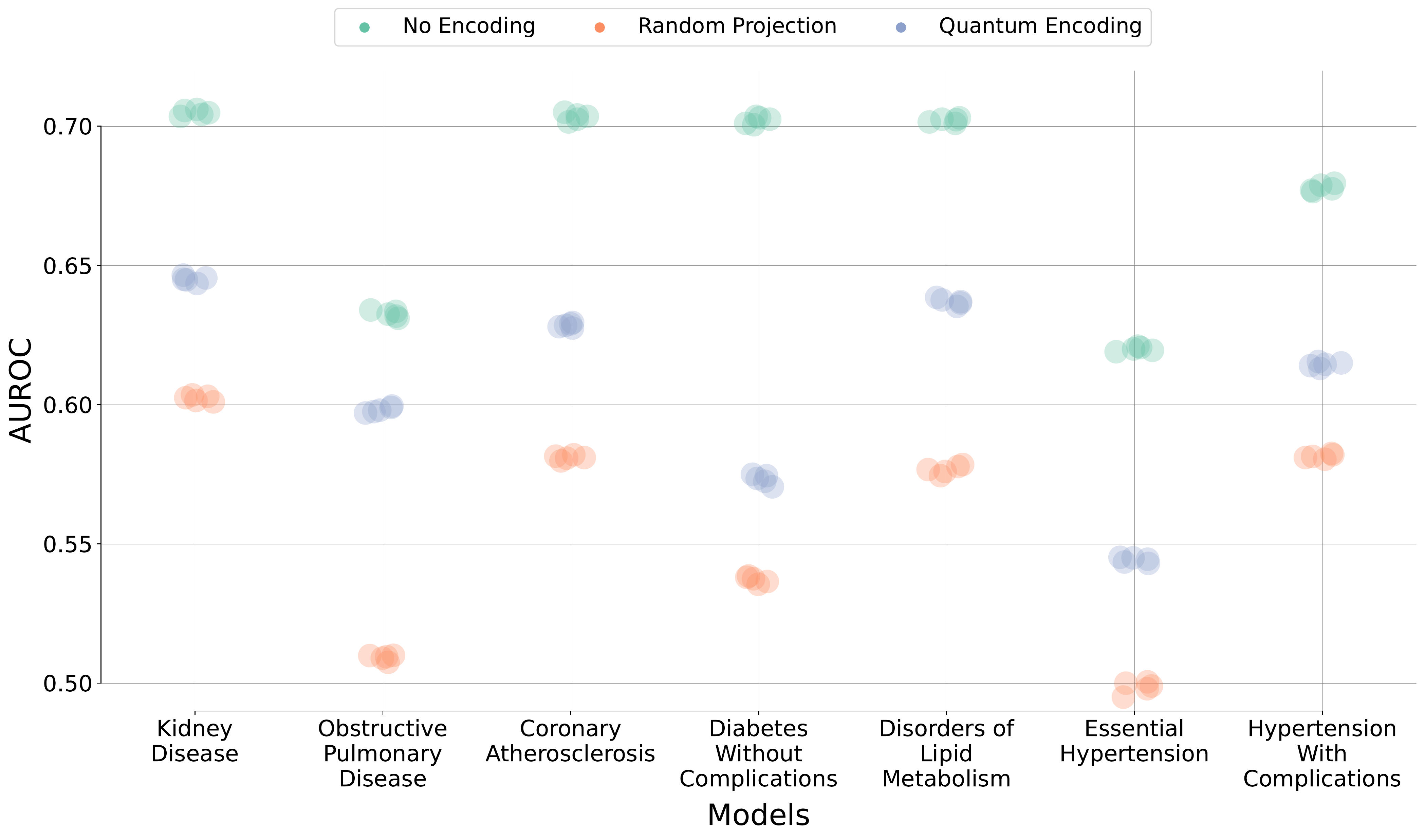}
    \caption{Temporal convolutional network}
\end{subfigure}%

\medskip

\begin{subfigure}{\textwidth}
    \centering
    \includegraphics[scale=0.18,trim={0 1.25cm 0 0},clip]{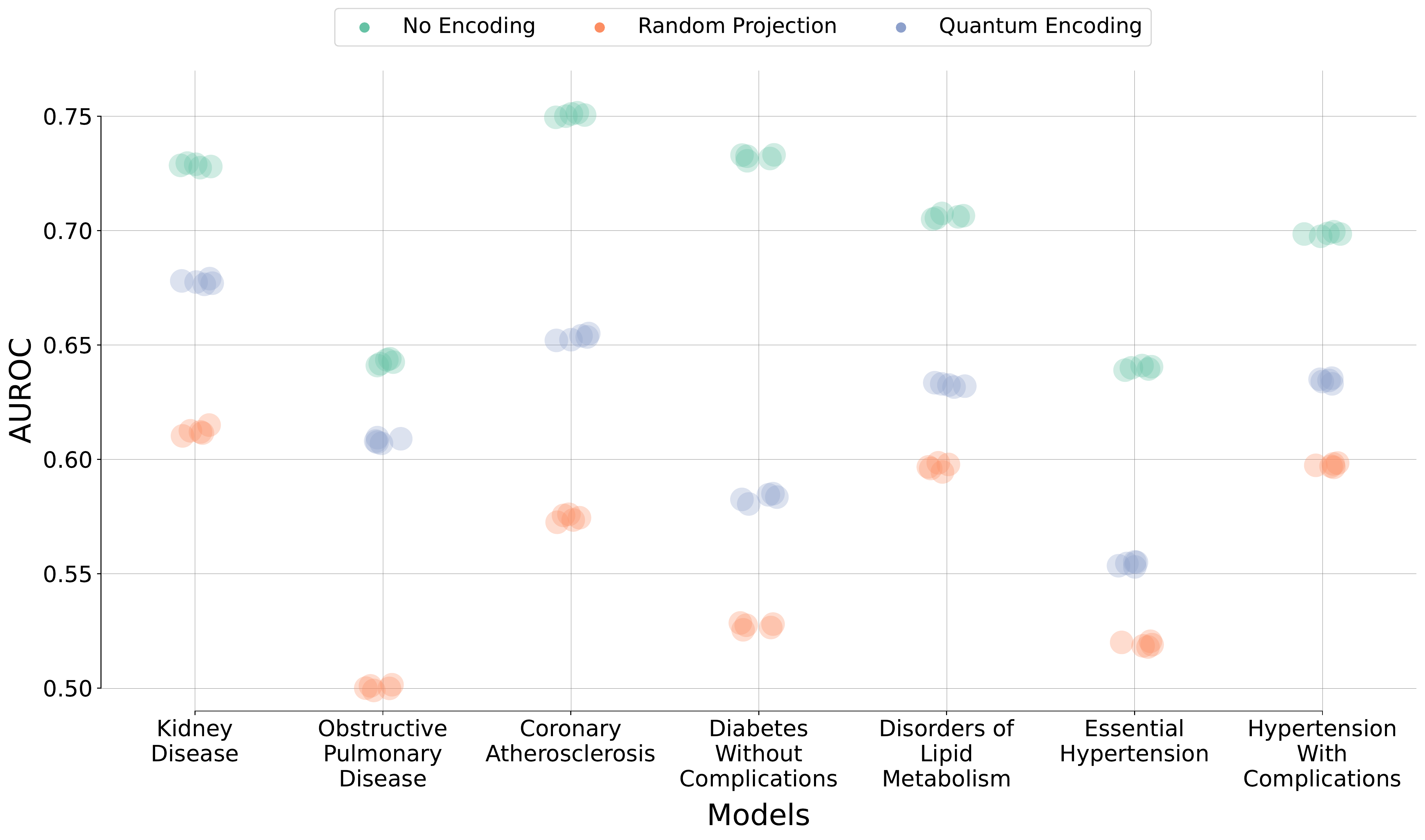}
    \caption{Multi-branch temporal convolutional network.}
\end{subfigure}%

\caption[short]{Latent chronic disorder prediction from different mortality prediction models.}
\label{fig:s3}
\end{figure}

Fig.~5 of the main text shows the accuracy in prediction of different chronic and acute disorders from LSTM mortality prediction models. Similarly, Fig.\ref{fig:s2} and Fig.\ref{fig:s3} documents the performance of predicting chronic and acute disorders from other mortality prediction models.

 \begin{figure}[H]
\centering
\begin{subfigure}{\textwidth}
    \centering
    \includegraphics[scale=0.25]{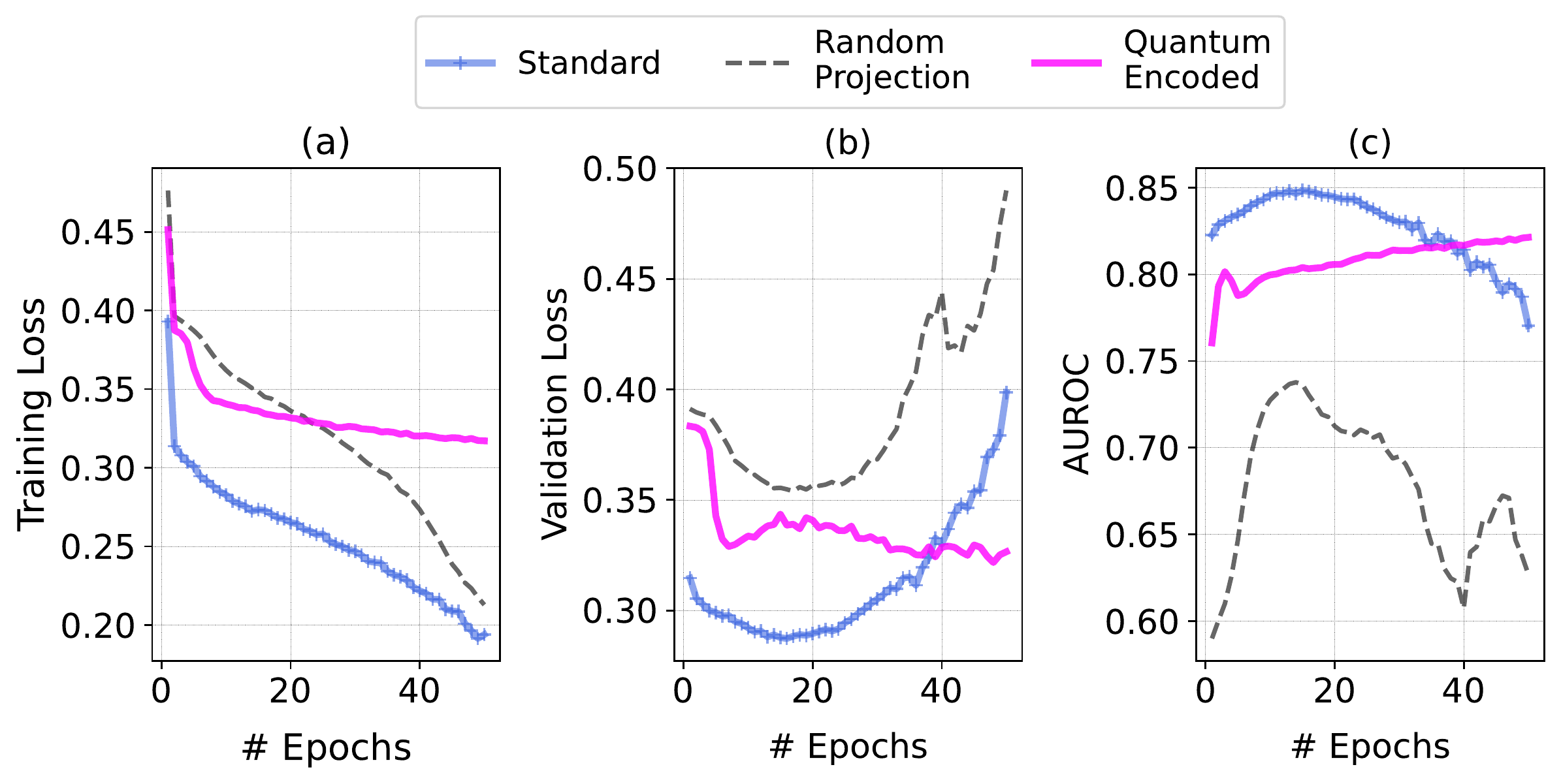}
    \caption{LSTM}
\end{subfigure}%

\begin{subfigure}{\textwidth}
    \centering
    \includegraphics[scale=0.25]{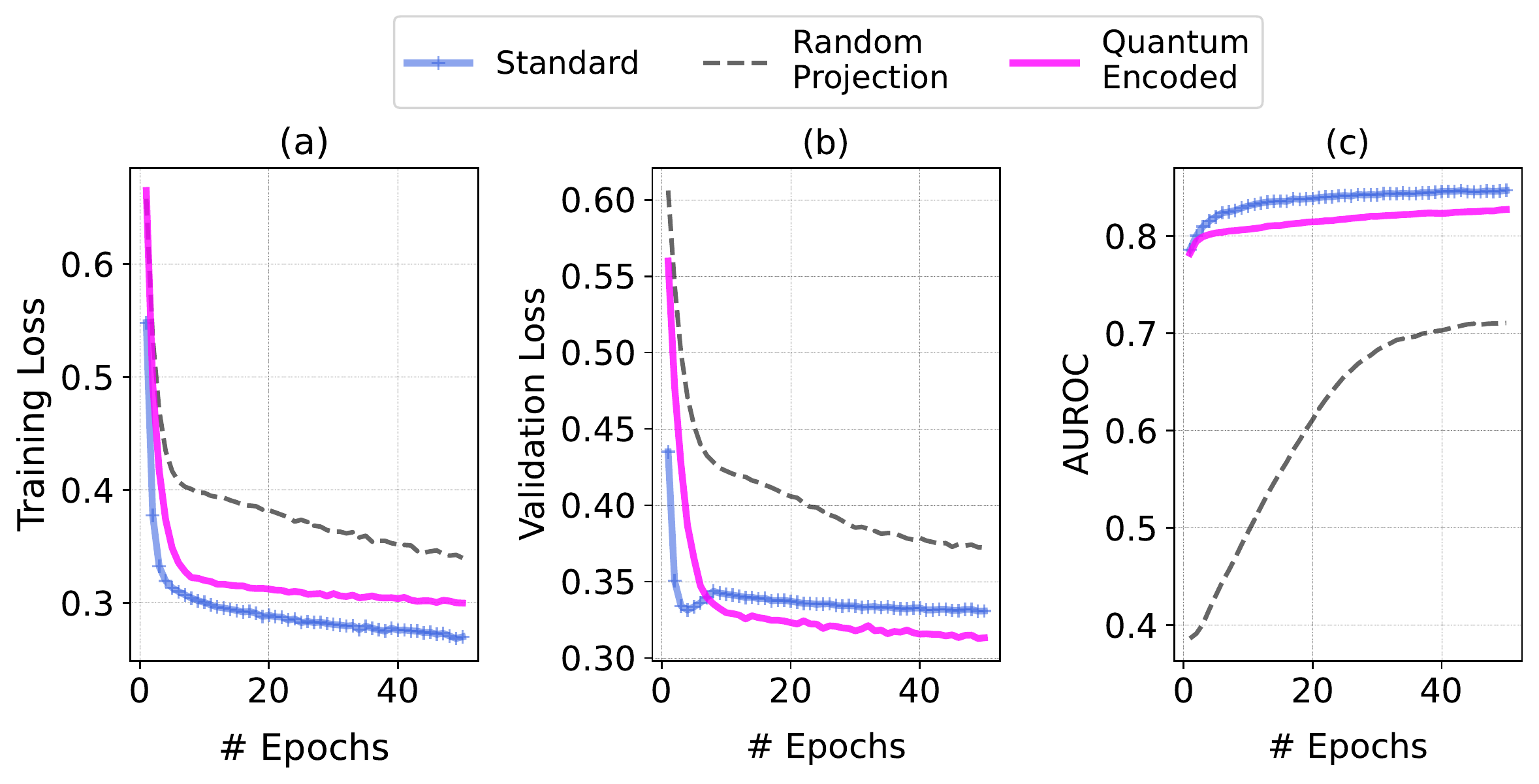}

    \caption{TCN}
\end{subfigure}%

\begin{subfigure}{\textwidth}
    \centering
    \includegraphics[scale=0.25]{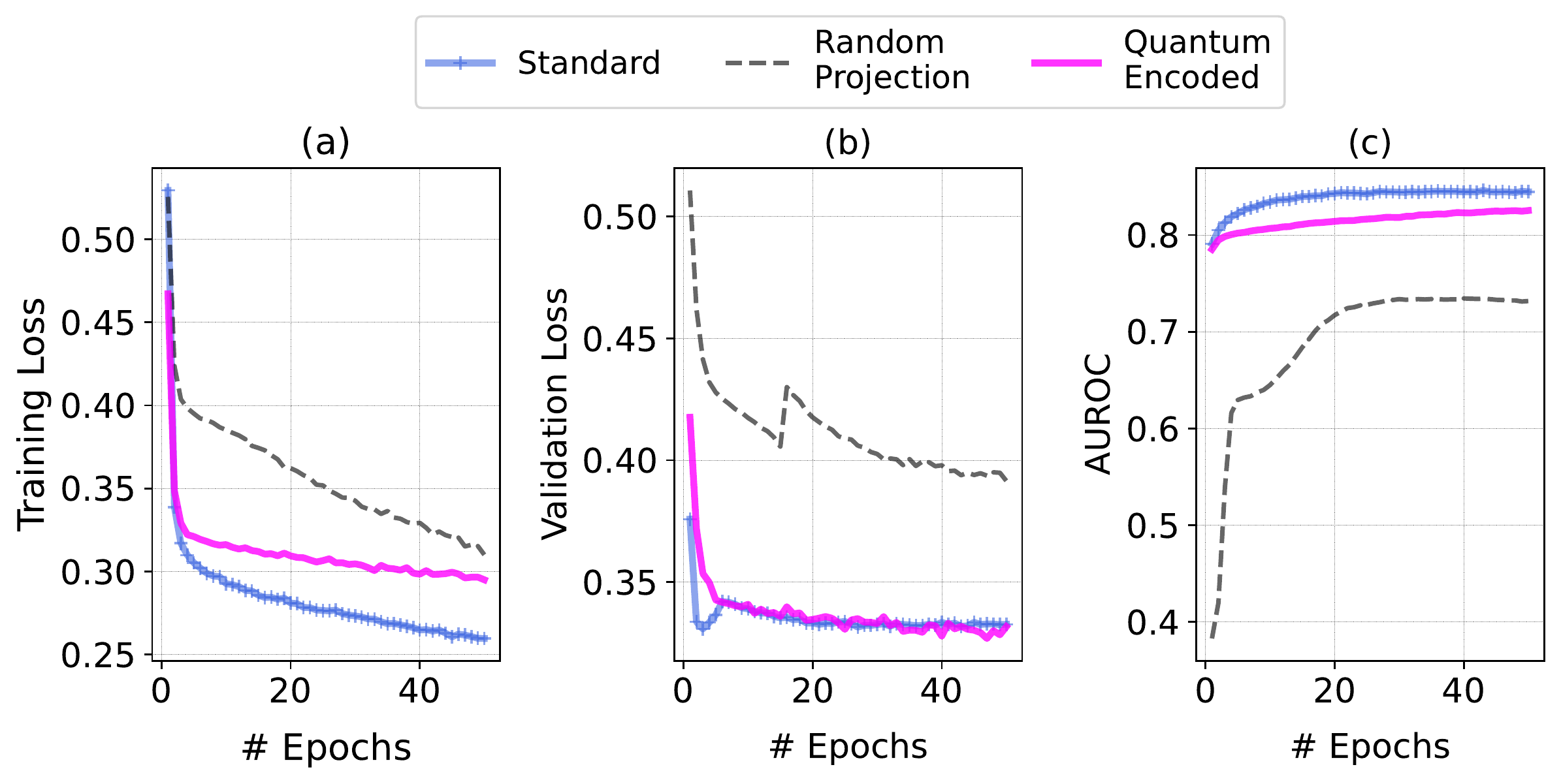}

    \caption{Multi-branch TCN}
\end{subfigure}%

\begin{subfigure}{\textwidth}
    \centering
    \includegraphics[scale=0.25]{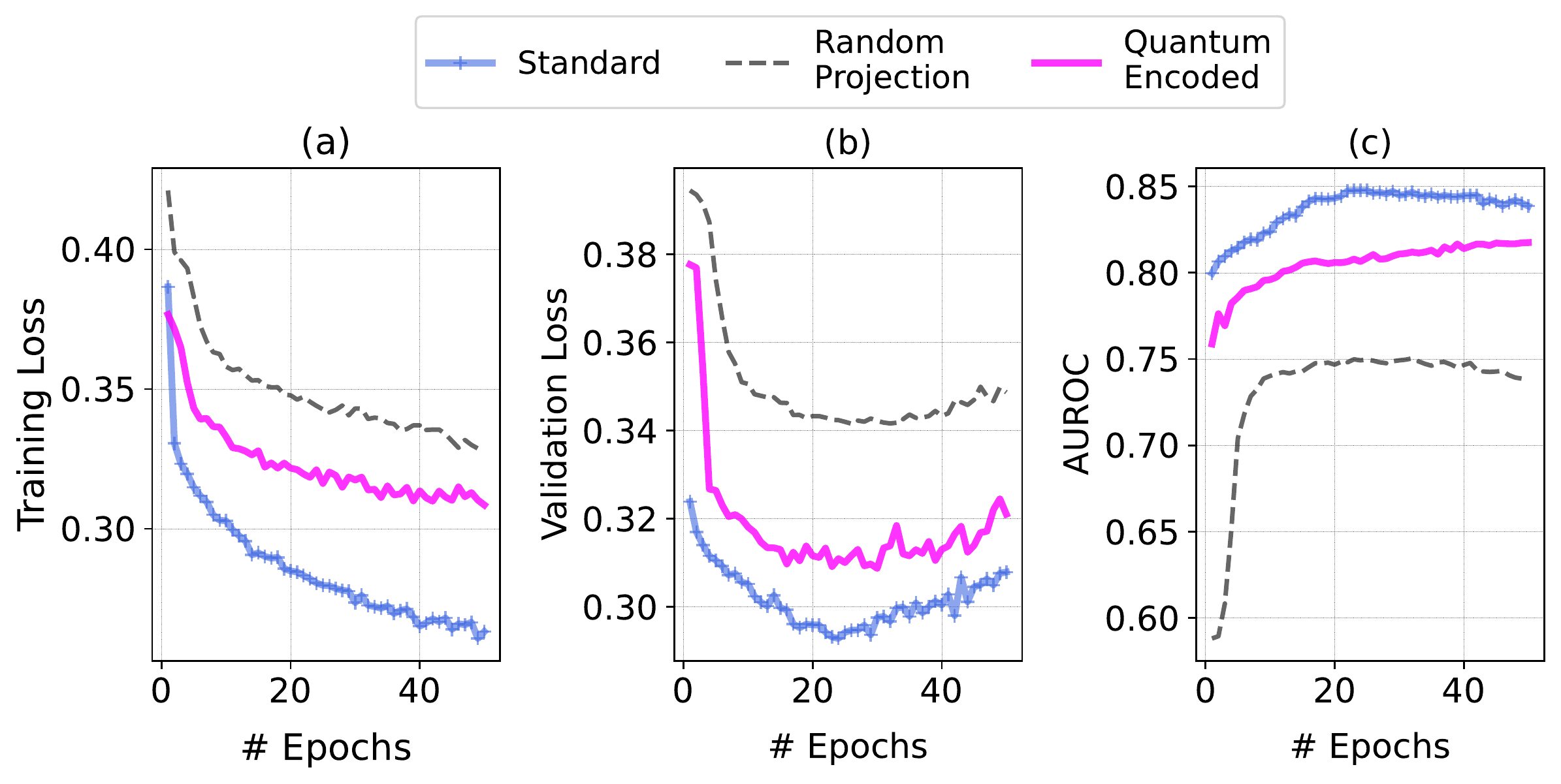}

    \caption{Transformer}
\end{subfigure}%

\begin{subfigure}{\textwidth}
    \centering
    \includegraphics[scale=0.25]{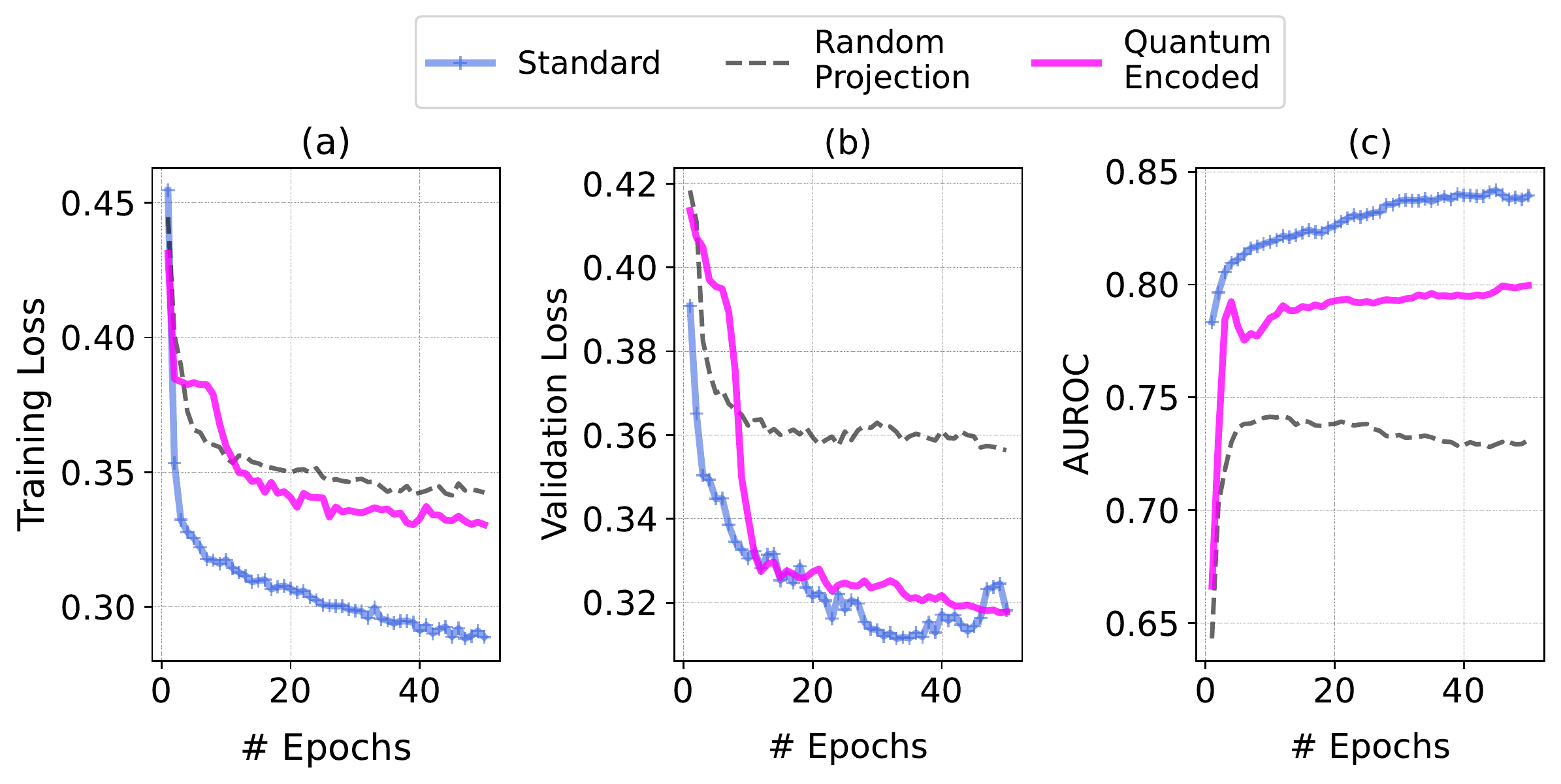}

    \caption{Transformer}
\end{subfigure}%

\caption[short]{Training dynamics of different models being trained for mortality prediction on MIMIC-III dataset.}
\label{fig:s5}
\end{figure}

\section{Dynamics of different models being trained for mortality prediction on MIMIC-III dataset}

Fig.~\ref{fig:s5} illustrates the training loss, validation loss and validation AUROC curves observed during training of LSTM, TCN and multi-branch TCN.

\bibliography{sn-bibliography}